\documentclass[preprint,12pt,authoryear]{elsarticle}

\usepackage{filecontents}
\usepackage{graphicx}
\usepackage{caption}
\usepackage{subcaption}
\usepackage{tabularx}
\usepackage{multirow}%
\usepackage{amsmath,amssymb,amsfonts}%
\usepackage{amsthm}%
\usepackage{mathrsfs}%
\usepackage[title]{appendix}%
\usepackage{xcolor}%
\usepackage{textcomp}%
\usepackage{manyfoot}%
\usepackage{booktabs}%
\usepackage{algorithm}%
\usepackage{algorithmicx}%
\usepackage{algpseudocode}%
\usepackage{listings}%
\usepackage{microtype}
\usepackage{xurl}         
\usepackage{hyperref}
\hypersetup{breaklinks=true}
\usepackage[authoryear]{natbib}

\theoremstyle{thmstyleone}%
\theoremstyle{thmstyletwo}%

\theoremstyle{thmstylethree}%

\journal{Neural Networks}

\begin{document}

\begin{frontmatter}



\title{Unsupervised Physics-Informed Operator Learning through Multi-Stage Curriculum Training} 

\cortext[cor1]{Corresponding author.}
\author[aff1]{Paolo Marcandelli\corref{cor1}}
\ead{paolo.marcandelli@polimi.it}
\author[aff2,aff4]{Natansh Mathur}
\author[aff3]{Stefano Markidis}
\author[aff1]{Martina Siena}
\author[aff1]{Stefano Mariani}

\affiliation[aff1]{organization={Department of Civil and Environmental Engineering, Politecnico di Milano}, city={Milan}, postcode={20133}, country={Italy}}
\affiliation[aff2]{organization={QC Ware Corp.}, city={Paris}, country={France}}
\affiliation[aff3]{organization={Department of Computer Science, KTH Royal Institute of Technology}, city={Stockholm}, country={Sweden}}
\affiliation[aff4]{organization={IRIF, CNRS \& Université Paris Cité}, city={Paris}, country={France}}

\begin{abstract}
Solving partial differential equations remains a central challenge in scientific machine learning. 
Neural operators offer a promising route by learning mappings between function spaces and enabling resolution-independent inference, yet they typically require supervised data. 
Physics-informed neural networks address this limitation through unsupervised training with physical constraints but often suffer from unstable convergence and limited generalization capability. 
To overcome these issues, we introduce a multi-stage physics-informed training strategy that achieves convergence by progressively enforcing boundary conditions in the loss landscape and subsequently incorporating interior residuals.  
At each stage the optimizer is re-initialized, acting as a continuation mechanism that restores stability and prevents gradient stagnation. 
We further propose the Physics-Informed Spline Fourier Neural Operator (PhIS-FNO), combining Fourier layers with Hermite spline kernels for smooth residual evaluation. 
Across canonical benchmarks, PhIS-FNO attains a level of accuracy comparable to that of supervised learning, using labeled information only along a narrow boundary region, establishing staged, spline-based optimization as a robust paradigm for physics-informed operator learning.
\end{abstract}





\end{frontmatter}

\section{Introduction}
Partial Differential Equations (PDEs) form the mathematical foundation of countless phenomena in science and engineering. 
However, numerically solving PDEs remains computationally demanding, as classical discretization schemes scale poorly with resolution and dimensionality. 
Accurate solutions can typically require prohibitively fine meshes, making high-fidelity data generation costly or even infeasible in practice \citep{https://doi.org/10.1002/qj.3803}.

Recent advances in machine learning (ML) have offered new perspectives for accelerating PDE solvers. 
Data-driven methods, for instance, directly learn mappings from inputs to PDE solutions through supervised training on large datasets. 
Despite their empirical success, such models often lack robustness when extrapolating the solution to new domains or parameter regimes~\citep{Li2020FourierNO, KIM2020109216, mohan2020embeddinghardphysicalconstraints, Thuerey_2020, um2021solverinthelooplearningdifferentiablephysics, RAISSI2019686}, and typically rely on vast amounts of high fidelity data that are expensive to generate~\citep{kim19a, pfaff2021learningmeshbasedsimulationgraph}.

Physics–Informed Neural Networks (PINNs)~\citep{raissi2018deephiddenphysics} introduced a major step forward by embedding the governing equations directly into the loss function, enabling training without labeled data and improving generalization to unseen domains~\citep{wandel2021learningincompressiblefluiddynamics, SplineWandel, wandel2025metamizerversatileneuraloptimizer}. 
Yet, despite their promise, PINNs often suffer from unstable convergence and discretization artifacts when trained without ground-truth data. 
They struggle on multi–scale dynamical systems~\citep{wang2020pinnsfailtrainneural, doi:10.1126/science.aaw4741} due to the ill-conditioning of PDE constraints~\citep{wang2020understandingmitigatinggradientpathologies, fuks2020limitations} and the difficulty of propagating boundary information across space and time~\citep{Sun_2020}. 
In addition, their inherent low frequency bias~\citep{rahaman2019spectralbiasneuralnetworks, cao2019spectralbias} hampers the resolution of fine scale features, making pointwise residual enforcement highly sensitive to collocation sampling~\citep{zhu2019physicsconstrained}. 
Fully unsupervised physics–informed learning therefore remains a challenging endeavor.

These limitations have motivated a paradigm shift from local, pointwise discretization toward operator based learning~\citep{Li2020FourierNO,li2020multipolegraphneuraloperator, li2020neuraloperatorgraphkernel, Kovachki}, an approach that approximates the underlying functional mappings governing PDE dynamics rather than their discrete solutions. 
Architectures such as DeepONet~\citep{deeponet}, Neural Operators~\citep{NeuralOperator}, and Fourier Neural Operators (FNOs)~\citep{Li2020FourierNO} have demonstrated that learning the operator itself yields smoother inductive biases and improved generalization across resolutions and geometries. 
In particular, FNOs leverage spectral convolution layers to mitigate the spectral bias of conventional networks~\citep{rahaman2019spectralbiasneuralnetworks, tancik2020fourierfeatures}, enabling global receptive fields and compact representations of both low and high frequency components. 
However, physics-informed extensions such as Physics-Informed Neural Operator (PINO)~\citep{li2021pino} still rely on spectral derivatives, leading to degraded accuracy and instability under non-periodic boundary conditions. 
Conversely, convolutional architectures such as U-Nets ~\citep{ronneberger2015unet} effectively capture local spatial features but remain grid-dependent and require retraining when scaling to different resolutions. 

Several recent studies have proposed extensions of the PINO framework to improve accuracy, stability, and generalization. 
For example the Variational PINO ~\citep{eshaghi2025vino} and the Physics-Informed Geometry-Aware Neural Operator ~\citep{zhong2024pigano} advance the representation and formulation of physics-informed operators, 
yet they remain inherently tied to spectral differentiation and rely on static optimization schemes.
In contrast, our work departs from the conventional formulation-based perspective and focuses instead on the optimizer dynamics that govern unsupervised physics-informed training.

The first key contribution of this work is a theoretically grounded, curriculum–inspired training strategy~\citep{curr1, graves2017automatedcurriculumlearningneural, Khadijeh2025} specifically designed for unsupervised physics–informed learning. 
Other curriculum and continuation schemes~\citep{MATTEY2022114474, duan2025copinn, kharazmi2021hpinns, guo2025longterm} typically adjust loss weights, sampling distributions, or temporal training windows to guide convergence, yet they remain largely empirical and do not explicitly account for optimizer internal dynamics. 
In contrast, we introduce a new training methodology based on the Adam optimizer~\citep{kingma2014adam}, in which each curriculum transition involves a full reinitialization of the optimizer state. 
This stage–wise reset prevents the accumulation of biased momentum statistics and allows the network to adapt as new physics based constraints are progressively introduced. 
The approach is motivated by the well posedness of PDEs: once boundary or initial conditions are satisfied, the interior solution is uniquely determined. 
Drawing inspiration from homotopy continuation~\citep{allgower2003introduction}, by gradually augmenting the loss landscape from boundary consistency to interior residual enforcement, resulting in stable and interpretable unsupervised optimization. This approach redefines how convergence is achieved rather than how the residual is formulated, providing a fundamentally new path toward stable, data-free operator learning.
By explicitly coupling stage–wise optimizer reinitialization with adaptive moment estimation, our framework provides, to our knowledge, the first systematic demonstration that staged Adam resets substantially enhance convergence and stability in unsupervised physics–informed operator learning.

As a second contribution, we introduce the Physics–Informed Spline Fourier Neural Operator (PhIS–FNO), a unified framework that bridges spectral and continuous formulations to enable stable, resolution-invariant, and physics-consistent operator learning. 
PhIS–FNO integrates the global expressiveness of Fourier layers with the continuous differentiation provided by Hermite spline kernels. 
This hybrid representation allows PDE residuals to be computed smoothly and consistently across both periodic and non–periodic domains, avoiding discretization artifacts and ensuring uniform gradient behavior. 
Unlike previous physics-informed operator formulations, PhIS–FNO naturally supports general boundary conditions while preserving the resolution invariance characteristic of operator learning architectures.

We validate the proposed framework across canonical PDE benchmarks, including the one–dimensional Burges' equation, the two–dimensional Poisson problem, and the incompressible Navier Stokes equations under both periodic and non–periodic boundary conditions, as well as the Kolmogorov flow. 
The multi–stage curriculum consistently enables convergence in the fully unsupervised regime, allowing all tested architectures, including PINO, Spline U–Net, and standard FNO variants, to successfully learn the underlying PDE dynamics. 
Among them, PINO achieves the best accuracy on periodic problems, while PhIS–FNO exhibits superior performance and stability in non periodic and large domain scenarios, where the spatial resolution typically involves thousands of grid points rather than tens or hundreds. 
The spline–based continuous differentiation mitigates the spectral leakage and truncation artifacts characteristic of PINO models that compute derivatives in the spectral domain, as discussed in Sec. \ref{sec:burger}.
Together, these findings demonstrate that coupling a continuous spline–based operator representation with a boundary to residual curriculum establishes a robust and generalizable foundation for unsupervised physics–informed learning.
Beyond improving training stability, this paradigm outlines a path toward scalable, data–free operator learning architectures applicable across diverse physical systems. 
The following sections detail the theoretical formulation, architectural design, and experimental validation.

Partial Differential Equations (PDEs) form the mathematical foundation of countless phenomena in science and engineering, spanning micromechanics, medical imaging, and turbulent fluid dynamics. Despite their central role, numerically solving PDEs remains computationally demanding, as classical discretization methods such as finite differences or finite elements scale poorly with resolution and dimensionality. Achieving accurate solutions typically requires prohibitively fine meshes, making high-fidelity data generation costly or even infeasible in practice~\citep{https://doi.org/10.1002/qj.3803}.\\
Recent advances in machine learning (ML) have offered new perspectives for accelerating PDE solvers through data-driven surrogate models. 
Physics-Informed Neural Networks (PINNs)~\citep{raissi2018deephiddenphysics} introduced a major step forward by embedding the governing equations directly into the loss function, enabling training without labeled data. 
Yet, despite their promise, PINNs face intrinsic difficulties: their optimization landscape is ill-conditioned, with unbalanced gradients between boundary and residual terms leading to unstable convergence~\citep{wang2020understandingmitigatinggradientpathologies, fuks2020limitations}. 
Moreover, standard neural architectures exhibit a strong \emph{spectral bias}~\citep{rahaman2019spectralbiasneuralnetworks, cao2019spectralbias}, learning smooth, low-frequency components more easily than high-frequency or multiscale dynamics. 
Consequently, PINNs struggle to capture stiff or turbulent regimes, and their pointwise residual enforcement remains highly sensitive to the sampling of collocation points~\citep{zhu2019physicsconstrained}. 
Fully unsupervised physics-informed learning therefore remains a challenging and unstable endeavor.\\
These challenges have motivated a paradigm shift from local, pointwise discretization toward \emph{operator-based learning}—an approach that seeks to approximate the underlying functional mappings governing PDE dynamics rather than their discrete solutions.
Operator-learning frameworks have recently emerged as a compelling alternative. 
Rather than approximating discrete field values, these models aim to learn mappings between infinite-dimensional function spaces, allowing resolution-independent inference and improved generalization.
Architectures such as DeepONet~\citep{deeponet}, Neural Operators~\citep{NeuralOperator}, and Fourier Neural Operators (FNOs)~\citep{Li2020FourierNO} have demonstrated that learning the underlying operator, rather than its pointwise realization, yields smoother inductive biases and better transfer across resolutions and geometries.
In particular, FNOs leverage spectral convolution layers to mitigate the \emph{spectral bias} of conventional networks~\citep{rahaman2019spectralbiasneuralnetworks, tancik2020fourierfeatures}, enabling global receptive fields and compact representations of both low- and high-frequency components. 
However, physics-informed extensions such as PINO~\citep{li2023physicsinformedneuraloperatorlearning} still rely on spectral derivatives that assume periodic domains, which leads to degraded accuracy and instability under non-periodic boundary conditions. 
Conversely, convolutional architectures such as U-Nets~\citep{ronneberger2015unet} effectively capture local spatial features but remain grid-dependent and are not inherently designed to model continuous PDE dynamics. 
As a result, current operator-learning approaches remain constrained by boundary assumptions and by the difficulty of achieving stable optimization in fully unsupervised regimes.\\
In this work, we propose a unified framework that bridges spectral and continuous formulations to enable stable, resolution-invariant, and physics-consistent operator learning. 
The proposed \emph{Physics-Informed Spline Fourier Neural Operator} (PhIS-FNO) integrates the global expressiveness of Fourier layers with the continuous differentiation provided by Hermite spline kernels. 
This hybrid representation allows PDE residuals to be computed in a smooth and continuous manner, avoiding discretization artifacts and ensuring consistent gradients across both periodic and non-periodic boundaries. 
Unlike previous physics-informed operator formulations, PhIS-FNO naturally supports general boundary conditions while preserving the resolution invariance characteristic of operator-learning architectures.\\
A second contribution of this work is a theoretically motivated, curriculum-inspired training strategy~\citep{curr1, graves2017automatedcurriculumlearningneural, Khadijeh2025}, specifically designed for unsupervised physics-informed learning. 
The method is grounded in the well-posedness of PDEs: once boundary and initial conditions are satisfied, the interior solution is uniquely determined. 
Drawing inspiration from homotopy continuation ~\citep{allgower2003introduction}, i.e. gradual deformation of the loss landscape toward the target objective, we first optimize the network solely on boundary constraints and then progressively introduce the physics residual loss. 
At each transition, the optimizer is fully reinitialized to prevent the accumulation of biased gradient information and to escape local minima inherited from earlier stages. 
This staged optimization acts as a form of self-transfer learning and provides, to our knowledge, the first systematic demonstration that optimizer reinitialization across curriculum phases can significantly improve convergence in unsupervised physics-informed operator learning.\\
We validate the proposed framework across canonical benchmarks, including the one-dimensional Burgers’ equation, the two-dimensional Poisson problem, and the incompressible Navier–Stokes equations under both periodic and non-periodic boundary conditions, as well as the Kolmogorov flow. 
The multi-stage curriculum consistently enables convergence in the fully unsupervised regime, allowing all tested architectures—including PINO, Spline-U-Net, and standard FNO variants—to successfully learn the underlying PDE dynamics. 
Among them, PINO achieves the best accuracy on periodic problems, while the proposed PhIS-FNO exhibits superior performance and stability in non-periodic and large-domain scenarios. 
The spline-based continuous differentiation mitigates spectral leakage and truncation artifacts, leading to smooth residual gradients and scale-consistent optimization. 
These findings demonstrate that coupling a continuous spline-based operator representation with a boundary-to-residual curriculum yields a robust and generalizable foundation for unsupervised physics-informed learning.\\
In summary, this work bridges the gap between physics-informed and operator-based paradigms, showing that unsupervised operator learning can be both theoretically grounded and practically stable across domains, resolutions, and boundary conditions.\\

\textbf{Our Contributions.}
This work introduces a unified framework for stable, resolution-invariant, and fully unsupervised physics-informed operator learning. 
The main contributions are summarized as follows:

\begin{itemize}
  \item \textbf{Hybrid Operator Representation.} 
  We propose the \emph{Physics-Informed Spline Fourier Neural Operator}, which combines the global spectral representation of FNOs with the continuous differentiation of Hermite spline kernels. 
  This hybrid design ensures smooth residual evaluation and consistent gradients across both periodic and non-periodic boundaries, overcoming the limitations of spectral derivatives and grid-dependent convolutional models.

  \item \textbf{Curriculum-Based Optimization.} 
  We introduce a multi-stage training strategy in which the network is first optimized on boundary constraints and then progressively incorporates the physics residual loss within the domain. 
  At each transition, the optimizer is reinitialized—acting as a form of self-transfer learning—that stabilizes convergence and prevents gradient bias accumulation. 
  Across all tested PDEs, this mechanism proved essential for convergence in fully unsupervised settings.

  \item \textbf{Generalized unsupervised learning across architectures.} 
  We show that physics-informed operator networks can be trained entirely without labeled data in the interior domain, achieving accuracy comparable to supervised counterparts. 
  PhIS-FNO achieves consistently strong performance across all settings, outperforming PINO and U-Net baselines in non-periodic and large-domain problems, while multi-stage PINO surpasses even the supervised FNO in periodic regimes. 
  These results highlight the universality and robustness of the proposed training paradigm.

  \item \textbf{Comprehensive Validation.} 
  Extensive experiments on Burgers’, Poisson, and incompressible Navier–Stokes equations (including Kolmogorov flow) demonstrate that the proposed framework achieves stable and accurate predictions across resolutions, reducing residual errors by up to an order of magnitude compared to baseline models.
\end{itemize}

By coupling continuous spline differentiation with staged physics-informed optimization grounded in PDE well-posedness, PhIS-FNO establishes a general and practical framework for unsupervised operator learning. 
The following sections detail the theoretical formulation, architectural design, and numerical validation.

\section{Methods} 
In this section, we present the conceptual and methodological components underlying the proposed framework.
The central idea is to decompose the learning process into successive stages, each focusing on increasingly complex physical objectives, from boundary consistency to full PDE residual minimization. 
This strategy is theoretically grounded in the optimizer dynamics discussed in Sec.~\ref{curriculum}, and is shown to stabilize convergence in regimes where direct residual-based training fails.

Furthermore, we couple the proposed training scheme with a novel operator-learning architecture, the PhIS-FNO, which extends the standard FNO by introducing Hermite spline kernels that provide a continuous, resolution-independent representation across spatial domains. 

We begin by introducing the class of partial differential equations considered in this work which serves to establish the notation and
the operator-theoretic perspective underlying physics-informed learning. On this basis, we then describe
the principles of Fourier-based operator learning, the spline-based representation adopted in PhIS--FNO,
and finally the integration of the multi-stage curriculum training strategy within this framework.
\subsection{Partial Differential Equations}\label{sec:pdes}
We begin by considering two standard categories of PDE problems. 
In the first, we consider three Banach spaces: the solution space $\mathcal{U}$, the parameter space $\mathcal{A}$, and the residual space $\mathcal{F}$. 
On these, we define the (possibly nonlinear) operator $\mathcal{L}: \mathcal{U} \times \mathcal{A} \to \mathcal{F}$.
 Then, we consider the system of equations:
\begin{align} \label{elliptic}
    \begin{split}
    \mathcal{L}(u, \alpha) = 0, \quad & \text{in } \Omega \subset \mathbb{R}^d, \\
    u = h, \quad & \text{on } \partial\Omega,
    \end{split}
\end{align}
where $\Omega$ is a bounded spatial domain, $\alpha \in \mathcal{A}$ denotes some PDE coefficient or parameter, 
and $u \in \mathcal{U}$ is the unknown solution. 
The boundary function $h$ specifies the imposed Dirichlet condition on $\partial\Omega$, although in certain formulations it can also be treated as part of the parameter set. 
This formulation defines the solution operator $\mathcal{G}^\dagger : \mathcal{A} \to \mathcal{U}$, mapping parameters $\alpha$ to solutions $u$, 
where the superscript $\dagger$ denotes the exact (ground-truth) operator.

For instance, in the Poisson equation
\begin{equation}
    \Delta\psi = \varphi,
\end{equation}
where $\Delta$ denotes the Laplace operator, the source term $\varphi \in \mathcal{A}$ is mapped into the potential $\psi \in \mathcal{U}$. \\[6pt]

In the second category we deal with time-dependent systems of PDEs and we consider $\mathcal{U}$ as the Banach space of the evolving solutions $u(t)$ for $t>0$ subjected to the dynamical law
\begin{align} \label{time-dependent}
\frac{\mathrm{d}u}{\mathrm{d}t} &= \mathcal{R}(u, \alpha)
\quad \text{in } \Omega \times (0,\infty), \\
u &= g
\quad \text{on } \partial\Omega \times (0,\infty), \\
u(\cdot,0) &= a
\quad \text{in } \overline{\Omega}\times\{0\},
\end{align}
where $a = u(0) \in \mathcal{A}$ is the initial condition, $g$ is a fixed boundary condition and $\mathcal{R}$ denotes a (possibly nonlinear) differential operator. 
This formulation naturally induces a solution operator
\begin{equation}
    \mathcal{G}^\dagger : \mathcal{A} \to C([0,T]; \mathcal{U}), \qquad a \mapsto u,
\end{equation}
which associates each initial condition with its temporal trajectory in $\mathcal{U}$. 
Classical examples of this formulation include the viscous Burgers' equation and the Navier Stokes equations. \\[6pt]

From a theoretical standpoint, the solution of a PDE is well-defined only once appropriate initial and boundary conditions are specified. 
In the absence of such constraints, the problem becomes ill-posed, admitting infinitely many possible solutions that differ by an additive constant or by functions in the null space of $\mathcal{L}$. 
This observation motivates our multi-stage training strategy: in the first stage, the network learns to satisfy the boundary conditions, effectively fixing the gauge of the solution and selecting one physically consistent member of the solution manifold. 
Subsequent stages progressively introduce the PDE residual term, following an homotopy-inspired curriculum training which allows the model to transition from a simple, well-conditioned optimization problem to the full physics-informed objective without destabilizing the training dynamics~\citep{allgower2003introduction}. 
The homotopy-inspired formulation is discussed in Sec. \ref{homotopy} and the detailed treatment of optimizer adaptation across stages in Sec.~\ref{curriculum}.\\

\subsection{Homotopy formulation of the curriculum loss}\label{homotopy}
Let $\mathcal{L}_\lambda$ denote the total training loss parametrized by 
$\lambda \in [0,1]$, $\mathcal{L}_{\mathrm{bc}}$ the loss enforcing boundary conditions 
and $\mathcal{L}_{\mathrm{res}}$ the physics-based residual over the interior of the domain.
The multistage training strategy adopted in this work is conceptually related to 
homotopy continuation methods~\citep{allgower2003introduction}, but does not enforce 
the normalized one-parameter loss interpolation
\[
\mathcal{L}_\lambda = (1-\lambda)\mathcal{L}_{\mathrm{bc}} + \lambda\mathcal{L}_{\mathrm{res}},
\]
which is commonly assumed in classical homotopy formulations.

Instead, each training stage is associated with an independent pair of loss weights: $\lambda_{\mathrm{bc}}$ for the boundary and $\lambda_{\mathrm{res}}$ for the interior domain. In this way, we have
\begin{equation}\label{lambda_bdlambda_res}
\mathcal{L}_{(\lambda_{\mathrm{bc}},\lambda_{\mathrm{res}})} 
= \lambda_{\mathrm{bc}}\,\mathcal{L}_{\mathrm{bc}} 
+ \lambda_{\mathrm{res}}\,\mathcal{L}_{\mathrm{res}},
\end{equation}
modulating the relative influence of boundary supervision and physics-based residual enforcement.
In the early stages, the weights are chosen to prioritize boundary consistency, thereby fixing the gauge of the solution through supervised boundary constraints.
Subsequently, the weights are progressively adjusted to incorporate residual information from the interior of the domain, 
shifting the optimization from boundary-dominated to residual-dominated regimes, where the interior is learned in a fully data-free (unsupervised) manner.

This discrete, stagewise progression can be interpreted as a generalized form of homotopy continuation,
in which the loss landscape is augmented incrementally rather than via a continuously varying parameter.
In practice, this formulation offers greater flexibility in shaping the training dynamics,
while retaining the stabilizing effects typically associated with homotopy-based continuation methods.

\subsection{Training with Physics-Informed Neural Networks}
Building on the problem formulation introduced above, which frames partial differential equations
as solution operators acting between function spaces, we now describe how this perspective
is leveraged within a physics-informed learning framework.

The objective of the physics-informed approach is to approximate the unknown solution operator 
$\mathcal{G}^\dagger$, which maps the input space $\mathcal{A}$ to the solution space $\mathcal{U}$ 
in the stationary setting, or to the space of time-dependent trajectories 
$C([0,T];\mathcal{U})$ for evolutionary partial differential equations.

By parameterizing the operator with a neural network with parameters $\theta$, denoted by 
$\mathcal{G}_\theta$, the training problem is cast as the minimization of a physics-informed loss 
such that $\mathcal{G}_\theta \approx \mathcal{G}^\dagger$. The corresponding network prediction 
is denoted by $u_\theta$. Rather than relying on supervised data, the optimization is driven by 
enforcing the governing PDE and the associated boundary and initial conditions directly.

For elliptic problems, the loss function is defined as a normalized residual in $L^2(\Omega)$ 
combined with boundary penalties, ensuring that the predicted solution satisfies the PDE in the 
interior and matches the prescribed conditions on $\partial\Omega$:
\begin{equation}\label{pde1}
    \mathcal{L}_{\mathrm{pde}}(\theta) = 
    \lambda_{\mathrm{res}}\left(\frac{1}{|\Omega|}
    \int_\Omega \big|\mathcal{L}(u_{\theta}(x),\alpha(x))\big|^2\,\mathrm{d}x\right)
    + \lambda_{\mathrm{bd}}\left(\frac{1}{|\partial\Omega|}
    \int_{\partial\Omega}\big|u_{\theta}(x)-h(x)\big|^2\,\mathrm{d}x\right),
\end{equation}
with notation defined in Eq.~\eqref{elliptic}, and $\mathrm{d}x$ denoting the Lebesgue measure in 
$\mathbb{R}^d$.

For time-dependent problems, such as the Burgers’ or Navier Stokes equations, the loss naturally 
extends to the space--time domain $L^2((0,T]\times\Omega)$, where the residual involves both temporal 
and spatial derivatives:
\begin{align}\label{pde2}
\begin{split}
\mathcal{L}_{\mathrm{pde}}(\theta) =\;
&\lambda_{\mathrm{res}}\left(\frac{1}{T|\Omega|}
\int_{0}^{T}\!\!\int_{\Omega}
\big|\partial_t u_\theta(t,x)-\mathcal{R}(u_\theta,\alpha)(t,x)\big|^{2}
\,\mathrm{d}x\,\mathrm{d}t\right) \\
&+ \lambda_{\mathrm{bc}}\left(\frac{1}{T|\partial \Omega|}
\int_{0}^{T}\!\!\int_{\partial \Omega}
\big|u_\theta(t,x)-g(t,x)\big|^{2}\,\mathrm{d}x\,\mathrm{d}t\right),
\end{split}
\end{align}
where the notation follows Eq.~\eqref{time-dependent}. Initial conditions may be enforced either 
through additional penalty terms or implicitly through the network input representation.
This formulation enables training without access to ground-truth solutions, 
while still converging to an accurate approximation of the underlying solution operator.

\subsection{Physics-Informed Spline Fourier Neural Operator}
In this work, we present the Physics-Informed Spline Fourier Neural Operator, 
which combines the ability of FNO~(\citep{NeuralOperator,Li2020FourierNO}) to learn mappings between functions 
and to propagate boundary information through the Fourier transform, with Hermite spline basis functions~(\citep{SplineWandel}) 
to efficiently solve PDEs within a physics-informed framework.\\
We start by introducing the two fundamental building blocks on which PhIS--FNO is based:
the Hermite spline basis, used to construct a continuous representation of the solution,
and the Fourier Neural Operator framework underlying the operator-learning component.

\subsubsection{Hermite Splines Basis kernel}
Hermite splines are piecewise-defined polynomials characterized by their values and the first $n$ derivatives 
at prescribed support points (see~\citep{SplineWandel,splinedeep} for details). 
In contrast to B-splines, Hermite splines employ local polynomial representations defined on individual grid cells.
Specifically, each physical grid cell is affinely mapped to a local, normalized coordinate $z \in [-1,1]$, 
within which the spline basis has compact support at the points $z=-1,0,1$, corresponding to the cell interfaces 
and midpoint.

The space of Hermite polynomials of degree $L$ on $z \in [-1,1]$ is spanned by the basis
\[
\mathcal{H}^L = \{ h_0^L(z), h_1^L(z), \dots, h_L^L(z) \},
\]
where the lower index $i$ denotes the basis function associated with the $i$-th derivative.
By construction, these basis functions vanish at the cell boundaries,
\[
h_i^L(\pm 1) = 0, \qquad h_i^L(0) \in [-1,1], \quad \forall i \in [0,L].
\]

With this basis, any function defined over a one-dimensional grid 
$\hat{X} = \{\hat{x}_1,\dots,\hat{x}_N\}$ on the interval $[x_1,x_N]$ can be represented as a continuous 
Hermite spline
\begin{equation}
    f(x) = \sum_{\substack{l \in [0,L],\\ \hat{x} \in \hat{X}:\, |x-\hat{x}| \le 1}}
    c_{\hat{x}}^{\,l}\, h_l^L\!\left(x-\hat{x}\right),
    \qquad x \in [x_1,x_N],
\end{equation}
where $x$ is a continuous spatial variable and the summation is restricted to grid points $\hat{x}$ in the local
neighborhood of $x$, due to the compact support of the basis functions.

On a two-dimensional grid $\hat{X} \times \hat{Y}$, the spline basis is constructed via tensor products of the
one-dimensional kernels, yielding the continuous representation
\begin{equation}
    f(x,y) =
    \sum_{\substack{l,m \in [0,L]\times[0,M],\\
    (\hat{x},\hat{y}) \in \hat{X}\times\hat{Y}:\,
    |x-\hat{x}| \le 1,\ |y-\hat{y}| \le 1}}
    c_{\hat{x},\hat{y}}^{\,l,m}\,
    h_l^L\!\left(x-\hat{x}\right)\,
    h_m^M\!\left(y-\hat{y}\right),
\end{equation}
where, again, the summation involves only grid points in the immediate neighborhood of $(x,y)$.

This local tensor-product construction facilitates the analytical evaluation of spatial derivatives.
Within our framework, the neural network predicts the spline coefficients $c$, from which a continuous
representation of the solution over the entire domain is reconstructed through localized convolutions.
When periodic boundary conditions are required, circular padding is employed.

\subsubsection{Learning Operator and Fourier Neural Operators}
Learning operators aim to approximate mappings between function spaces from a finite collection of input--output pairs.
Neural Operators (NOs), such as FNOs, provide a multilayer neural implementation of this idea;
see~\citep{Li2020FourierNO,NeuralOperator} for details.

Let $D \subset \mathbb{R}^d$ denote the spatial domain and $D_j \subset D$ its discretized grid.
We consider two separable Banach spaces:
$\mathcal{A}=\mathcal{A}(D_j;\mathbb{R}^{d_a})$, containing input functions with values in $\mathbb{R}^{d_a}$,
and $\mathcal{U}=\mathcal{U}(D_j;\mathbb{R}^{d_u})$, consisting of output functions valued in $\mathbb{R}^{d_u}$.
Here, $d_a$ denotes the dimensionality of the input features evaluated at each spatial location,
which in our setting consist of the (scalar) PDE solution augmented with the spatial coordinates.
For example, $d_a=2$ for the one-dimensional Burgers’ equation, corresponding to $(u,x)$,
whereas $d_a=3$ for the two-dimensional Navier--Stokes equations in vorticity form, corresponding to $(\omega,x,y)$.
The output dimension $d_u$ represents the dimensionality of the learned solution representation,
which in our framework corresponds to a parametric expansion of the solution with spline coefficients.

Our objective is to approximate a generally nonlinear solution operator $\mathcal{G}^\dagger:\mathcal{A}\to\mathcal{U}$.

The neural operator introduced in ~\citep{NeuralOperator} works by applying a pointwise lifting map $P$, 
which transforms the input function $\{a:D_j\to\mathbb{R}^{d_a}\}$ into a higher-dimensional representation 
$\{v_0:D_j\to\mathbb{R}^{d_{v}}\}$, i.e. $v_0(x) = P(a(x))$ with $x\in D_j$. 
The operator is then defined as an iterative update sequence 
$v_0 \mapsto v_1 \mapsto \cdots \mapsto v_T$, 
where each $v_j: D_j \to \mathbb{R}^{d_{v}}$ for $j=0,\ldots,T-1$ is defined as
\begin{equation}\label{kernel}
    v_{t+1}(x) = \sigma \!\left(Wv_t(x) + (\mathcal{K}(a;\phi)v_t)(x)\right), 
    \qquad \forall\, x \in D_j,
\end{equation}
with $\sigma:\mathbb{R}\to\mathbb{R}$ a nonlinear activation, 
$W:\mathbb{R}^{d_v}\to\mathbb{R}^{d_v}$ a learnable linear transformation, 
and $\mathcal{K}(a;\phi)$ a kernel integral operator parameterized by a neural network with trainable weights $\phi$.  

Finally, a projection map $Q$ transforms the last latent state 
$\{v_T:D_j\to\mathbb{R}^{d_v}\}$ into the output function 
$\{u:D_j\to\mathbb{R}^{d_u}\}$.  
In our framework, PDE solutions are learned from boundary information and physics-informed losses in the interior domain. 
To this end, we replace the kernel integral operator in~\eqref{kernel} with a convolution operator defined in Fourier space~\citep{Li2020FourierNO}, 
where each Fourier mode encodes information aggregated over the entire spatial domain. 
Formally, for a latent state $v_t:D_j \to \mathbb{R}^{d_v}$, a spectral layer acts as
\[
v_{t+1}(x_n) \;=\; \mathcal{F}^{-1}\!\big( R(k)\,\widehat{v_t}(k)\big)(x_n) 
= \sum_{k\in\Lambda} R(k)\,\widehat{v_t}(k)\, e^{i k\cdot x_n},
\qquad 
\widehat{v_t}(k) = \sum_{y\in D_j} v_t(y)\,e^{-ik\cdot y},
\]
where $\Lambda \subset \mathbb{Z}^d$ denotes the set of Fourier modes retained in the truncation, 
$k\in\Lambda$ is the corresponding frequency vector, and 
$R(k)\in \mathbb{C}^{d_v\times d_v}$ is a learnable complex-valued multiplier that acts as a spectral filter and the dot product accounts for the multidimensional spatial coordinates.  
Equivalently, in the physical domain this operation corresponds to a global convolution
\[
v_{t+1}(x_n) \;=\; \sum_{y\in D_j} K(x_n-y)\,v_t(y),
\]
with kernel $K$ supported on the whole domain $D_j$. 
Therefore, the loss applied at the boundary $\partial D_j$ propagates through $K$ to all spatial locations. 
In other words, although supervision is imposed only on the boundary, the corresponding gradients update the Fourier coefficients $R(k)\widehat{v_t}(k)$, 
each of which contributes to the reconstruction of the solution at every spatial location. 
This mechanism ensures that boundary information propagates globally across the domain.\\

\subsubsection{PhIS-FNO} 
The novelty of our architecture lies in the integration of the spline formulation with the FNO, resulting in a framework particularly well suited for unsupervised, physics-informed operator learning. 
Instead of directly predicting the PDE solution, the network outputs a set of spline coefficients that parameterize a continuous Hermite spline representation of the target field. 
From these coefficients, the solution of the PDE is reconstructed by convolution with the spline basis. 
This design enables efficient computation of differential operators by directly acting on the kernel basis functions, 
and through convolution these operators can be applied to the same spline coefficients to reconstruct the differentials of the solution. 
As a result, the spline representation provides a natural interface for physics-informed training, where PDE residuals can be directly evaluated and used to guide the training process.

\subsection{Multi-stage Curriculum Training}\label{curriculum}

In this work, we propose a multi-stage curriculum training strategy that mirrors the mathematical well-posedness of PDEs:
the network, given an initial condition, first learns to satisfy the boundary conditions, thereby fixing the solution gauge,
and only in subsequent stages are the PDE residuals enforced in the interior domain.
Each transition corresponds to a new stage in the curriculum, where the relative weighting of boundary and residual losses
evolves according to the homotopy-inspired formulation introduced in Sec.~\ref{sec:pdes}.

At every stage, training is carried out using the Adam optimizer~\citep{kingma2014adam}, which updates the network parameters
$\theta$ by combining gradient information with adaptive, moment-based estimates of first- and second-order statistics.
Specifically, Adam maintains exponential moving averages of the gradients and their squared values,
denoted by $m_t$ and $v_t$, respectively, and updates the parameters according to
\begin{equation}
\begin{aligned}
\begin{cases}
m_{t} &= (1 - \beta_1) g_{t} + \beta_1 m_{t-1} \\[4pt]
v_{t} &= (1 - \beta_2) g_{t}^{2} + \beta_2 v_{t-1} \\[6pt]
\theta_{t} &= \theta_{t-1} - \eta \,
\frac{m_{t}}{\sqrt{v_{t}} + \epsilon},
\end{cases}
\end{aligned}
\end{equation}
where $g_t = \nabla_\theta \mathcal{L}_t(\theta_t)$ denotes the gradient of the loss function with respect to the network parameters.
Here, $m_t$ can be interpreted as a smoothed estimate of the descent direction, obtained by averaging recent gradients,
while $v_t$ measures their typical magnitude and is used to adaptively rescale the update. They can be written as:
\begin{equation}
\begin{aligned}
\begin{cases}
m_{t-1} &= \dfrac{(1 - \beta_1)}{1 - \beta_1^t}
\sum_{k=0}^{t-1} \beta_1^k g_{t-k} \\[6pt]
v_{t-1} &= \dfrac{(1 - \beta_2)}{1 - \beta_2^t}
\sum_{k=0}^{t-1} \beta_2^k g_{t-k}^{2}.
\end{cases}
\end{aligned}
\end{equation}

Together, these quantities stabilize the optimization by reducing the effect of noisy gradients
and by adjusting the effective learning rate during training.
The parameters $\beta_1$ and $\beta_2$ control the exponential decay rates of these moving averages and are set to their
standard values $\beta_1=0.9$ and $\beta_2=0.999$, which are known to provide stable convergence across a wide range of problems ~\citep{kingma2014adam}.
The constant $\epsilon \approx 10^{-8}$ is introduced for numerical stability and $t$ denotes the iteration index within the current training stage.

\vspace{6pt}
\noindent
When the optimizer is reset at the beginning of a new stage, all internal statistics are set to zero, 
i.e. $m_{t} = v_{t} = 0$. 
In this case (denoted by the subscript $r$, for \emph{reset}), the parameter update simplifies to:
\begin{equation}
\Delta\theta_{r} = \theta_t-\theta_{t-1}= -\text{sign}(g_t)\,\eta\,\frac{1-\beta_1}{\sqrt{1-\beta_2}}.
\end{equation}

In contrast, when no reset is applied (denoted by the subscript $nr$, for \emph{no-reset}), 
the parameter update depends on the accumulated first and second moments from the previous stage:
\begin{equation}
\Delta\theta_{nr} = -\eta\,\frac{(1-\beta_1)g_t + \beta_1 m_{t-1}}{\sqrt{(1-\beta_2)g_t^2 + \beta_2 v_{t-1}} + \epsilon}.
\end{equation}

After convergence, the optimizer settles into a local minimum. 
When the loss is subsequently updated with new weighting factors ~$(\lambda_{bd},\lambda_{res})$, 
the optimizer often fails to adapt effectively: 
the gradient term $(1-\beta_2)g_t^2$ becomes negligible compared to the accumulated variance term~$\beta_2 v_{t-1}$, 
while $\epsilon$ remains vanishingly small. 
As a result, the optimizer remains trapped in the previous local basin of attraction. 
This phenomenon is empirically confirmed in Appendix (Sec. ~\ref{sup:multi-stage}, Fig.~\ref{check3}), 
which reports results from the one-dimensional Burgers experiment, showing that 
$(1-\beta_2)\lVert g_{t}\rVert^2 \ll \beta_2 v_{t-1}$ 
across all layers and stage transitions.

Then, under this assumption, the non-reset update simplifies to
\begin{equation}
\Delta\theta_{nr} \approx 
-\eta\,\frac{(1-\beta_1)g_t + \beta_1 m_{t-1}}{\sqrt{\beta_2 v_{t-1}} + \epsilon}.
\end{equation}
We define the ratio between the magnitude of the reset and no-reset updates as
\begin{equation}
R = \frac{|\Delta\theta_r|}{|\Delta\theta_{nr}|}
  = \frac{(1-\beta_1)\sqrt{\beta_2\,v_{t-1}}}
         {\sqrt{(1-\beta_2)}\lvert(1-\beta_1)g_t + \beta_1 m_{t-1}\rvert}.
\end{equation}
We observe that the ratio $R$ remains consistently greater than one across all layers and stage transitions 
(see Appendix, Sec.~\ref{sup:multi-stage}, Fig.~\ref{Rbl}), 
confirming that optimizer reinitialization systematically increases the effective update magnitude 
$|\Delta\theta_r|$ relative to the no-reset case. 
This mechanism enables the optimizer to escape shallow minima 
and avoid stagnation in flat regions of the loss landscape where accumulated Adam moments would otherwise induce inertia.

A similar behavior is observed in the evolution of the effective learning rate 
$\bar{\eta}_{\text{eff}} = \eta m_t / (\sqrt{v_t}+\epsilon)$ (Appendix, Fig.~\ref{etaeff}). 
Without reset, $\bar{\eta}_{\text{eff}}$ decays monotonically as $v_t$ accumulates, 
leading to optimizer freezing, whereas reinitialization restores large effective steps 
and reactivates convergence at each stage. 

As detailed in Sec.~\ref{results}, this reset-based multi-stage scheme consistently yields faster and more stable convergence 
than both no-reset and single stage baselines, validating the continuation based interpretation introduced above.

\section{Results}\label{results}
In this section, we present a comprehensive set of experiments designed to evaluate the generality and effectiveness of the proposed physics-informed operator learning framework and its multi-stage training strategy. 
Our objective is to demonstrate that the method achieves high-fidelity solutions to nonlinear partial differential equations without access to labeled data in the interior of the domain. 
To this end, we reproduce and extend canonical benchmarks originally introduced in~\citep{Li2020FourierNO, li2023physicsinformedneuraloperatorlearning, RAISSI2019686}, 
thereby enabling a direct comparison with supervised and semi-supervised operator learning approaches.

We evaluate performance across a hierarchy of problems of increasing complexity and spatio-temporal richness:
(i) the two-dimensional Poisson equation with Dirichlet boundary conditions, 
(ii) the one-dimensional viscous Burgers’ equation, 
(iii) the two-dimensional incompressible Navier Stokes equations in vorticity form, 
(iv) the Kolmogorov flow, and 
(v) the cylinder wake dynamics. 

\subsection{Experimental Setup}
All models were implemented in \texttt{PyTorch} and trained on a single NVIDIA GeForce RTX~4070 GPU (8\,GB VRAM) using CUDA~12.6. 
Each Fourier-based model (FNO, PINO, and PhIS-FNO) employs four Fourier layers as the architectural backbone, 
with PhIS-FNO extending the output to nine spline channels corresponding to second-order two-dimensional basis functions.
Unless otherwise specified, all experiments were performed under identical training conditions and optimizer settings to ensure fairness. To have more information about model specifics and loss weights, see Appendix Sec. \ref{sup:multi-stage}. 
All reported results are fully reproducible, and the complete implementation is available on \href{https://github.com/PaoloMarcandelli/Unsupervised-Physics-Informed-Operator-Learning-through-Multi-Stage-Curriculum-Training}{GitHub} under request.
Furthermore, to evaluate the robustness of the proposed multi-stage curriculum, we repeated some experiments over four independent random initialization of the network parameters for each benchmark problem: Burgers’ equation, incompressible Navier Stokes (viscosity $10^{-3}$), and the Cylinder Wake flow (viscosity $10^{-2}$), see Appendix Sec. \ref{sup:seed} for more details. Furthermore, we introduced in Sec. \ref{lambda_config}, an ablation study regarding different configurations of $\lambda$ parameters, to ensure the generalizability of our method over different multi-stage loss weights configurations. The results clearly indicate that the multi-stage (MS) strategy yields the most stable and consistent performance among all configurations.

\subsection{Poisson Equation}\label{sec:poisson}

As a first benchmark, we consider the two-dimensional Poisson equation on the unit square domain $\Omega = [0,1]^2$,
\begin{equation}
\label{eq:poisson}
\begin{cases}
-\Delta \psi(x,y) = f(x,y), & (x,y) \in \Omega, \\[4pt]
\psi(x,y) = \psi^*(x,y), & (x,y) \in \partial\Omega,
\end{cases}
\end{equation}
with the source term
\[
f(x,y) = \sin(2\pi x)\sin(2\pi y),
\]
for which the analytic solution is
\[
\psi^*(x,y) = \frac{1}{8\pi^2}\sin(2\pi x)\sin(2\pi y).
\]
The Dirichlet boundary condition $\psi = \psi^*$ ensures that the problem is well-posed and uniquely defined, effectively fixing the additive gauge inherent to Laplacian operators. 
This configuration provides an ideal controlled setting to evaluate the method, as both the source and the solution are smooth and periodic while maintaining an exact boundary reference for training.

In our implementation, the network input consists of the concatenation of the source field $f(x,y)$ with the spatial coordinates $(x,y)$ defined on a uniform grid $128\times128$. 
The model outputs a continuous spline-based representation $\psi(x,y)$ of the potential field, 
from which both the potential and its Laplacian can be computed analytically via spline convolution operators. 
Hence, the network implicitly learns the operator
\begin{equation}
\mathcal{G}^\dagger : f(\cdot)\ \mapsto\ \psi(\cdot),
\end{equation}
mapping the source term to the corresponding potential field.

The training objective combines two components:  
(i) a PDE residual loss, enforced as a mean-squared error (MSE) on $\Delta\psi + f = 0$ in the interior of the domain, and  
(ii) a boundary loss, which constrains $\psi$ to match the analytic solution on a narrow band along the boundary.  
Optimization is performed in three stages according to the proposed curriculum, 
progressively shifting the relative weighting as presented in Tab. \ref{tab:loss-weights} in Appendix.

\begin{figure*}[htbp]
    \centering   
    \includegraphics[width=\linewidth]{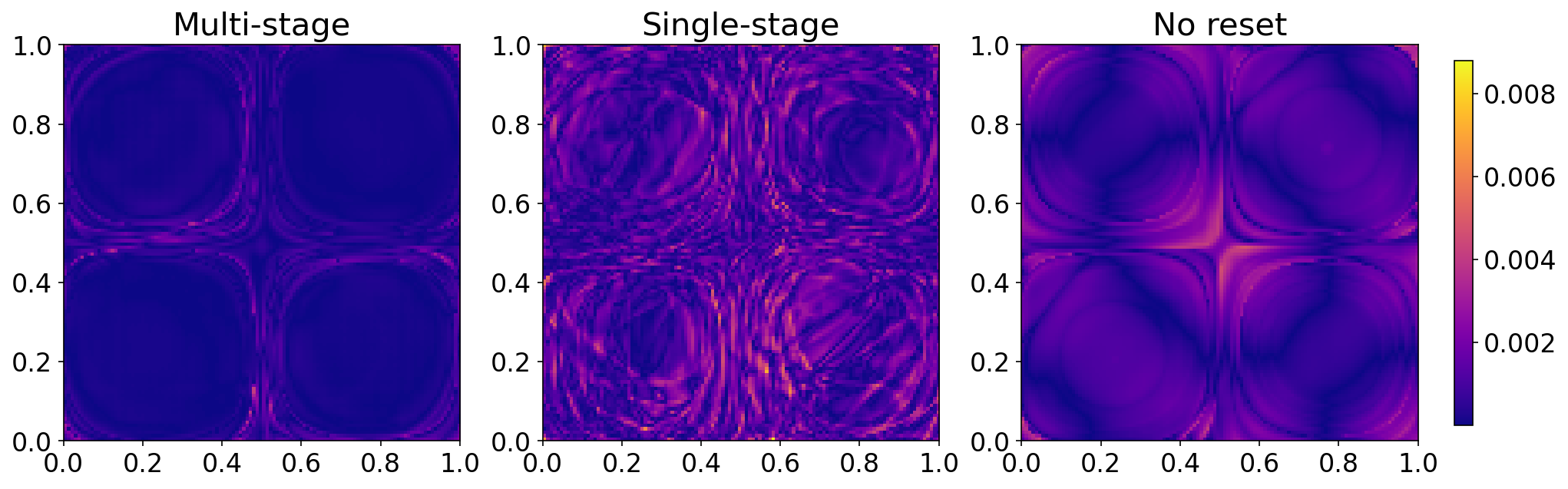}
    \caption{
    Comparison among different training strategies on the Poisson benchmark. 
    Each panel shows the magnitude of the PDE residual $|\Delta\psi + f|$ within the interior domain.
    The multi-stage approach (left) achieves the lowest residual and the most uniform convergence,
    whereas both the single-stage and the no-reset variants exhibit higher residuals and spatial oscillations,
    confirming the benefit of staged training and optimizer reinitialization.}
    \label{Poisson}
\end{figure*}

To evaluate the effect of the proposed multi-stage scheme, we compare three training strategies:  
(i) multi-stage training with optimizer reset (our proposed configuration),  
(ii) multi-stage training without reset, and  
(iii) a single-stage baseline, matching the standard PINN formulation, where $\lambda_{\mathrm{bd}}=\lambda_{\mathrm{res}}=1$ throughout training.  
Figure~\ref{Poisson} reports the spatial distribution of the internal PDE residual $|\Delta\psi + f|$ for the three cases.

The results show that the multi-stage curriculum achieves the most accurate solution, 
with an interior mean residual of $2.64\times10^{-4} \pm 3.66\times10^{-4}$, 
significantly lower than both the single-stage ($1.23\times10^{-3} \pm 9.27\times10^{-4}$) 
and no-reset ($1.03\times10^{-3} \pm 7.28\times10^{-4}$) configurations. 
These results confirm that the proposed multi-stage optimization consistently reduces PDE residuals and enforces both boundary consistency and interior physical constraints, yielding smoother and more physically coherent solutions in the fully unsupervised regime.

\subsection{One dimensional Burgers' Equation}\label{sec:burger}
The one-dimensional Burgers’ equation is a nonlinear PDE for viscous transport and serves as a standard benchmark for physics-informed models. It is written as
{\small
\begin{align*}
    \partial_t u(x,t) + \partial_x\!\left(\tfrac{1}{2}u(x,t)^2\right) &= \nu\,\partial_{xx}u(x,t),
    \qquad x \in (0,L),\ t>0,\\
    u(x,0) &= u_0(x),\qquad u(\cdot,t)\ \text{periodic in } x,
\end{align*}}
with viscosity $\nu>0$. 
In our experiment, we consider a periodic domain of length $L=2\pi$ and set $\nu=0.1$, see Appendix Sec. \ref{sup:dataset} for more details.
The learning objective is to approximate the solution operator that maps the initial condition to a single short-time forecast,
\[
\mathcal{G}^\dagger:\ u_0(\cdot)\ \mapsto\ u(\cdot,\Delta t),
\qquad \Delta t=0.1.
\]

The PhIS-FNO takes as input the concatenation of the initial state and the spatial coordinate field, 
$[\,u_0(x),\,x\,]$, and outputs spline coefficients that parameterize a continuous Hermite-spline representation of $u$.
From these coefficients we obtain $u$, its first and second derivatives $u_x$ and $u_{xx}$ via spline-based convolutions.
We enforce a physics-informed residual in semi-discrete form,
\[
r(x)\;=\;\frac{u(x,\Delta t)-u_0(x)}{\Delta t}\;+\;u(x,\Delta t)\,\partial_x u(x,\Delta t)\;-\;\nu\,\partial_{xx}u(x,\Delta t),
\]
penalized in the interior of the domain, together with a boundary regularization on a narrow periodic band to promote consistency at the domain edges.
Optimization is carried out in three stages following the multi-stage training strategy described in Sec.~\ref{curriculum}, 
progressively decreasing the weight of the boundary loss while increasing that of the residual loss (see Table \ref{tab:loss-weights} for more information).

Evaluation is performed against held-out targets at $t=\Delta t$ using $L^2$ metrics:
\begin{equation}
\mathcal{L}_{L^2}(u,\hat u)
= \frac{1}{N} \sum_{i=1}^{N}
\lvert u(x_i) - \hat u(x_i) \rvert^2 .
\end{equation}
Here, $u$ denotes the ground-truth solution, $\hat u$ the model prediction,
$\{x_i\}_{i=1}^N$ are the evaluation points on the spatial grid,
and $N$ is the total number of points used to approximate the $L^2$ norm.
\begin{figure*}[htbp]
    \centering

    \includegraphics[width=0.8\linewidth]{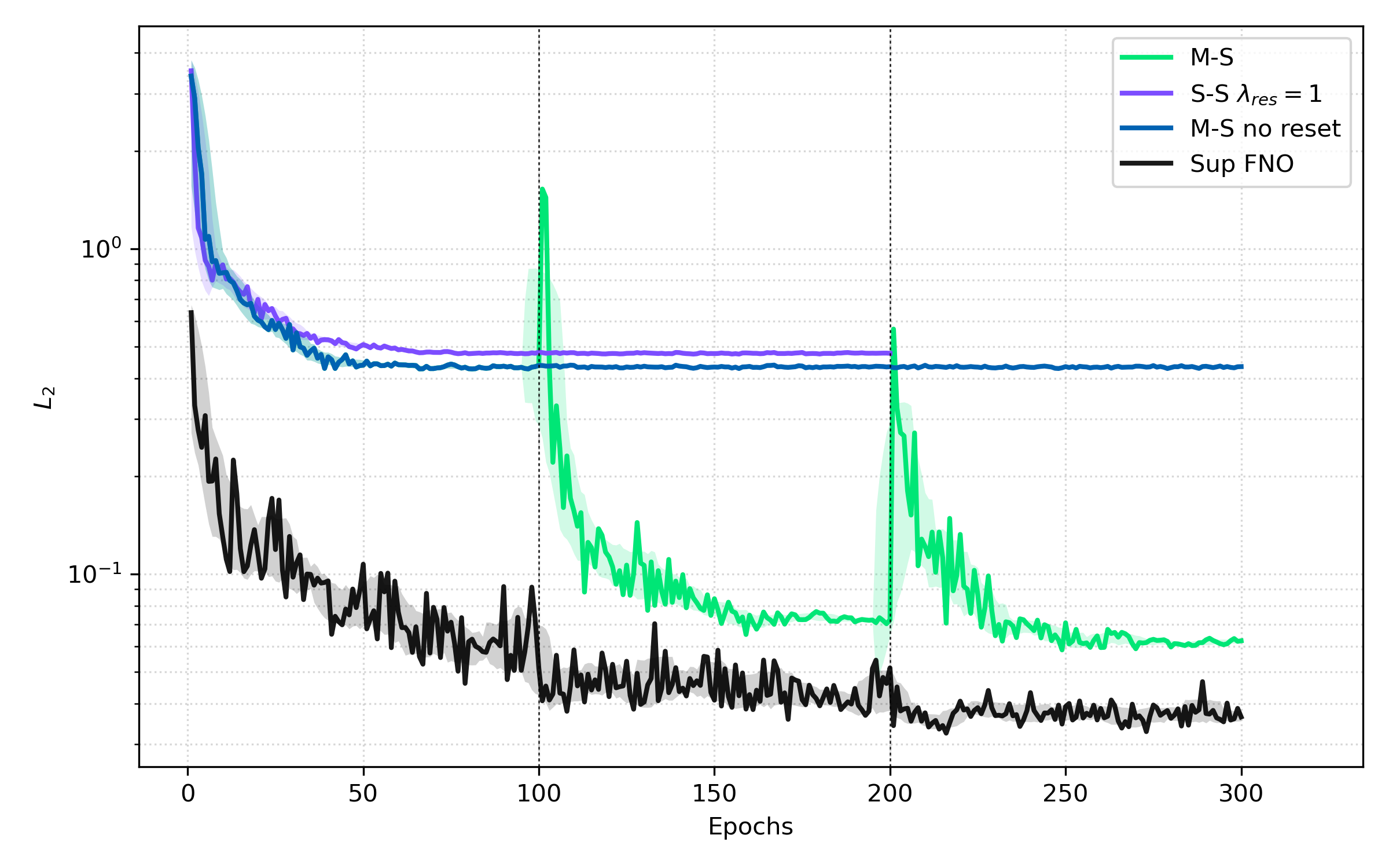}
    \caption{Validation loss comparison across epochs for the PhIS-FNO under different training strategies: multi-stage  with reset (green), multi-stage without reset (purple), single-stage (blue) and supervised FNO (black). 
    Sharp loss drops correspond to stage transitions in the multi-stage scheme.}
    \label{LossBurger}
\end{figure*}

The results of our experiments are reported in Fig.~\ref{LossBurger},  Table~\ref{tab:ns_kolm_all} and further analysis in Appendix (Sec. \ref{sup:burger}). 
The proposed PhIS-FNO achieves accuracy comparable to that of the standard supervised FNO, 
confirming the effectiveness of the approach in solving complex time-dependent PDEs. 
We do not include the Physics-Informed Neural Operator ~\citep{li2021pino} as a baseline, because its Fourier domain differentiation to compute residuals becomes numerically unstable on the fine spatial grids required in this problem (thousands of points). In the spectral domain, differential operators scale with powers of the wave number (e.g., $k^2$ for the Laplacian). On large grids, the high-frequency modes acquire very large coefficients, which amplify any truncation or approximation error introduced by the neural operator. As a consequence, the residual blows up and training rapidly diverges. Our spline–based formulation, operating entirely in the physical domain to compute loss residuals, does not suffer from this issue.


The multi-stage configuration with optimizer reset consistently outperforms both the no-reset and single-stage variants, 
achieving nearly an order of magnitude lower $L_2$ loss and variance at convergence. 
This experiment provides a first quantitative validation of the framework, paving the way for its application to higher-dimensional and turbulent regimes.



  

  

\subsection{Navier Stokes and Kolmogorov Flow}

The incompressible Navier Stokes equations in vorticity form for a viscous fluid are given by
{\small
\begin{align*}
    \partial_{t} \omega(x, t) + u(x, t)\cdot \nabla \omega(x,t) &= \nu \Delta \omega(x,t) + f(x), \quad
    x \in (0,1)^2,\; t \in (0,T], \\
    \nabla \cdot u(x,t) &= 0, \quad x \in (0,1)^2,\; t \in [0,T], \\
    \omega(x,0) &= \omega_0(x), \quad x \in (0,1)^2,
\end{align*}}
defined on the unit torus, then subjected to periodic boundary conditions. Here, $\nabla = (\partial_{x_1},\partial_{x_2})$ denotes the spatial gradient,
$\nabla\cdot$ the divergence operator, and $\Delta = \nabla\cdot\nabla$ the Laplacian. 
The velocity field is denoted by $u:(0,1)^2\times \mathbb{R}_+ \to \mathbb{R}^2$, 
$\omega$ represents the out-of-plane component $(0,0,\omega)$ of the vorticity $\nabla\times u:(0,1)^2\times \mathbb{R}\to\mathbb{R}^3$, 
$f$ the external forcing term, and $\nu$ the viscosity. 
The initial vorticity $\omega_0(x)$ is sampled from the Gaussian measure 
$\mu = \mathcal{N}\!\left(0,\;7^{3/2}(-\Delta + 49I)^{-2.5}\right)$ under periodic boundary conditions, see Ref.~\citep{Li2020FourierNO}.  

In the Navier Stokes experiments we employ a trigonometric forcing term
\[
f(x,y) \;=\; 0.1 \,\bigl(\sin(2\pi(x+y)) + \cos(2\pi(x+y))\bigr),
\]
whereas for the Kolmogorov flow we adopt the classical sinusoidal form
\[
f(x,y) \;=\; -n \cos(n y),
\]
which drives the long-time dynamics.  

Training is performed in time windows of length 10, starting at $T_{\text{in}}=40, 20$ for Navier Stokes with viscosity $\nu=10^{-3}, 10^{-4}$ respectively and $T_{\text{in}}=30$ for Kolmogorov flow, where the dynamics is already in a statistically stationary regime driven by the external forcing, see Appendix Sec. \ref{sup:dataset} for more information. From these windows we aim to learn the solution operator
\[
\mathcal{G}^\dagger : C([T_{\text{in}},T_{\text{in}}+10];\,\mathcal{H}^s_{\text{per}}((0,1)^2;\mathbb{R}))
\;\to\; C((T_{\text{in}}+10,T];\,\mathcal{H}^s_{\text{per}}((0,1)^2;\mathbb{R})),
\]
for any $s>0$, which maps the vorticity trajectory on the interval $[T_{\text{in}},T_{\text{in}}+10]$ to its subsequent evolution. Here, $\mathcal{H}^s_{\mathrm{per}}$ denotes the periodic Sobolev space of order $s>0$,
with $s$ specifying the spatial regularity of the vorticity field. 
The PhIS-FNO predicts spline coefficients from which we reconstruct the stream function $\psi$, and subsequently obtain the velocity field $u = \nabla^\perp \psi = (\partial_y \psi,\,-\partial_x \psi)$ and the vorticity $\omega=\nabla\times u$ by spline-based convolutions.  \\
\begin{figure*}[htbp]
    \centering
    \begin{subfigure}[t]{0.32\textwidth}
        \centering
        \includegraphics[width=\linewidth]{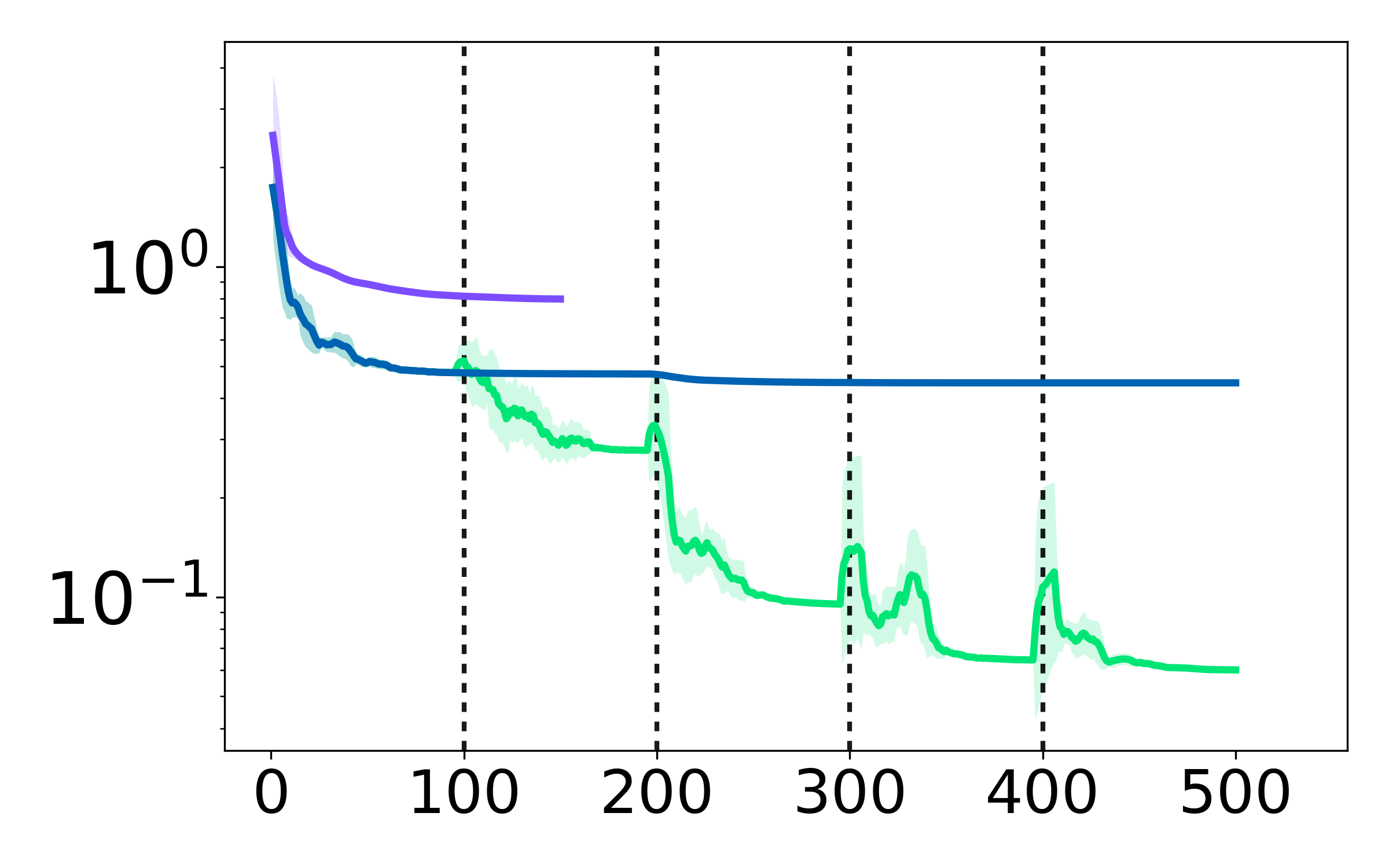}
        \caption{\textbf{PhIS-FNO}: $\nu=10^{-3}$}
    \end{subfigure}\hfill
    \begin{subfigure}[t]{0.32\textwidth}
        \centering
        \includegraphics[width=\linewidth]{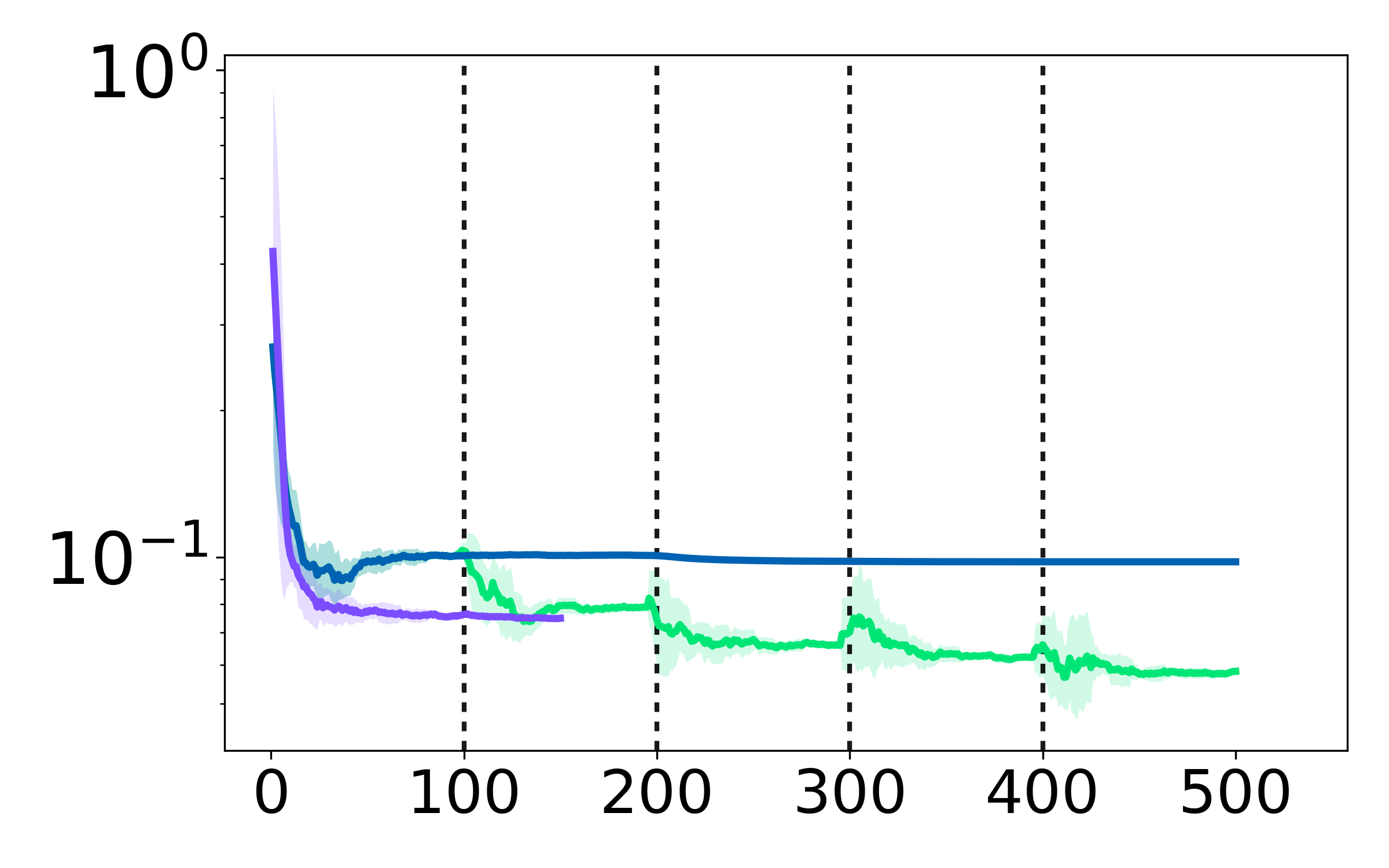}
        \caption{\textbf{PINO}: $\nu=10^{-3}$}
    \end{subfigure}\hfill
    \begin{subfigure}[t]{0.32\textwidth}
        \centering
        \includegraphics[width=\linewidth]{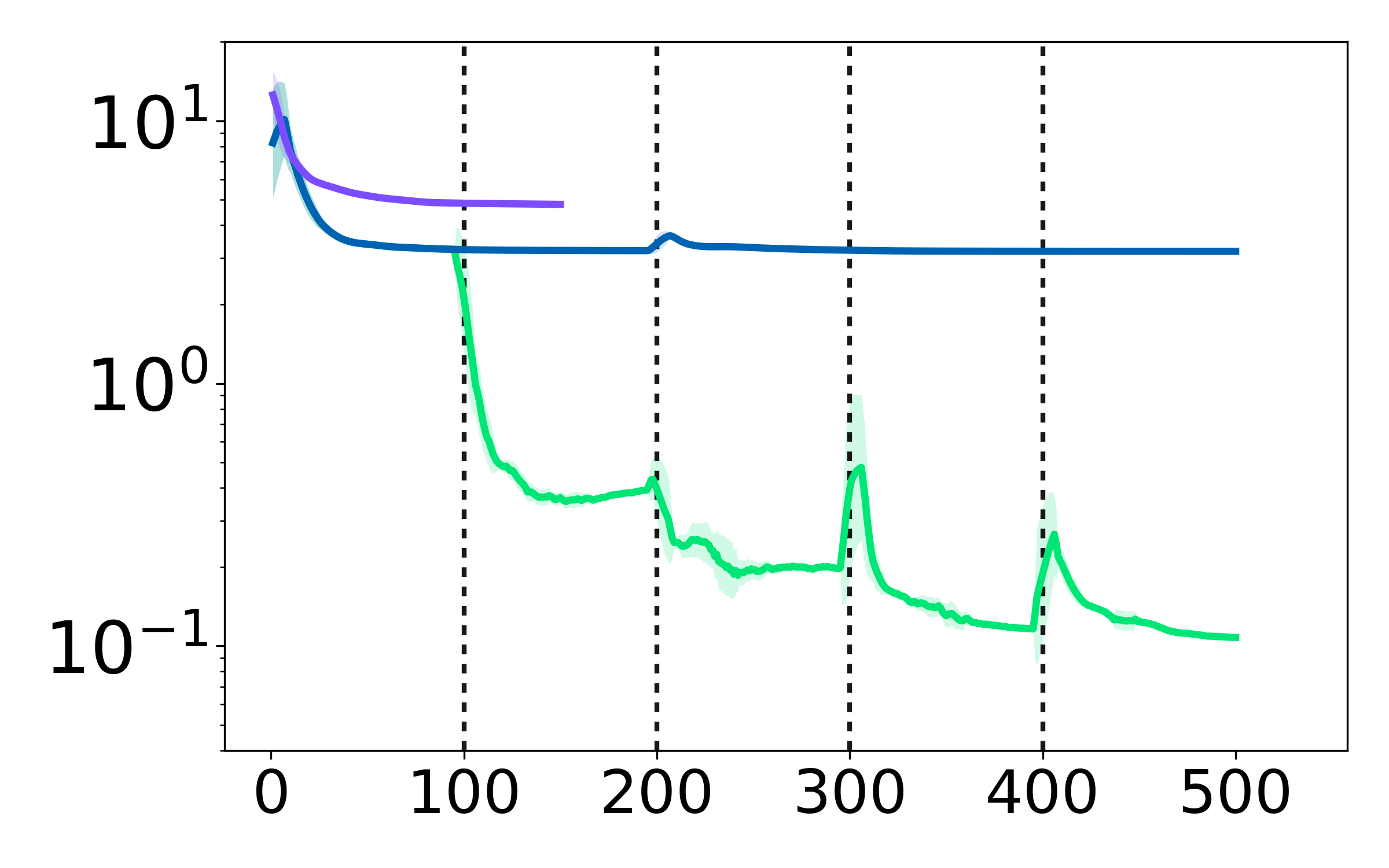}
        \caption{\textbf{UNet}: $\nu=10^{-3}$}
    \end{subfigure}
    \begin{subfigure}[t]{0.32\textwidth}
        \centering
        \includegraphics[width=\linewidth]{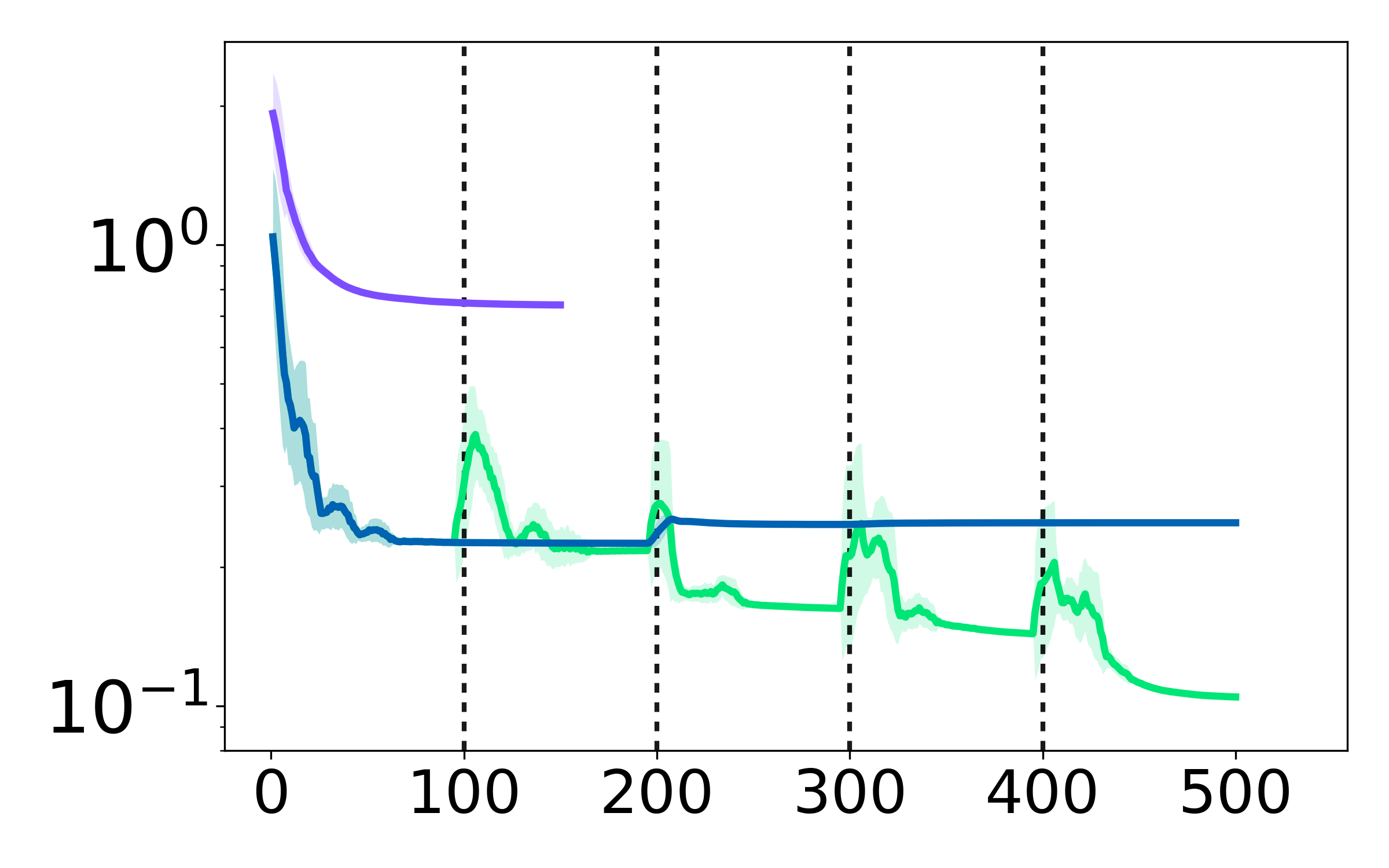}
        \caption{\textbf{PhIS-FNO}:$\nu=10^{-4}$}
    \end{subfigure}\hfill
    \begin{subfigure}[t]{0.32\textwidth}
        \centering
        \includegraphics[width=\linewidth]{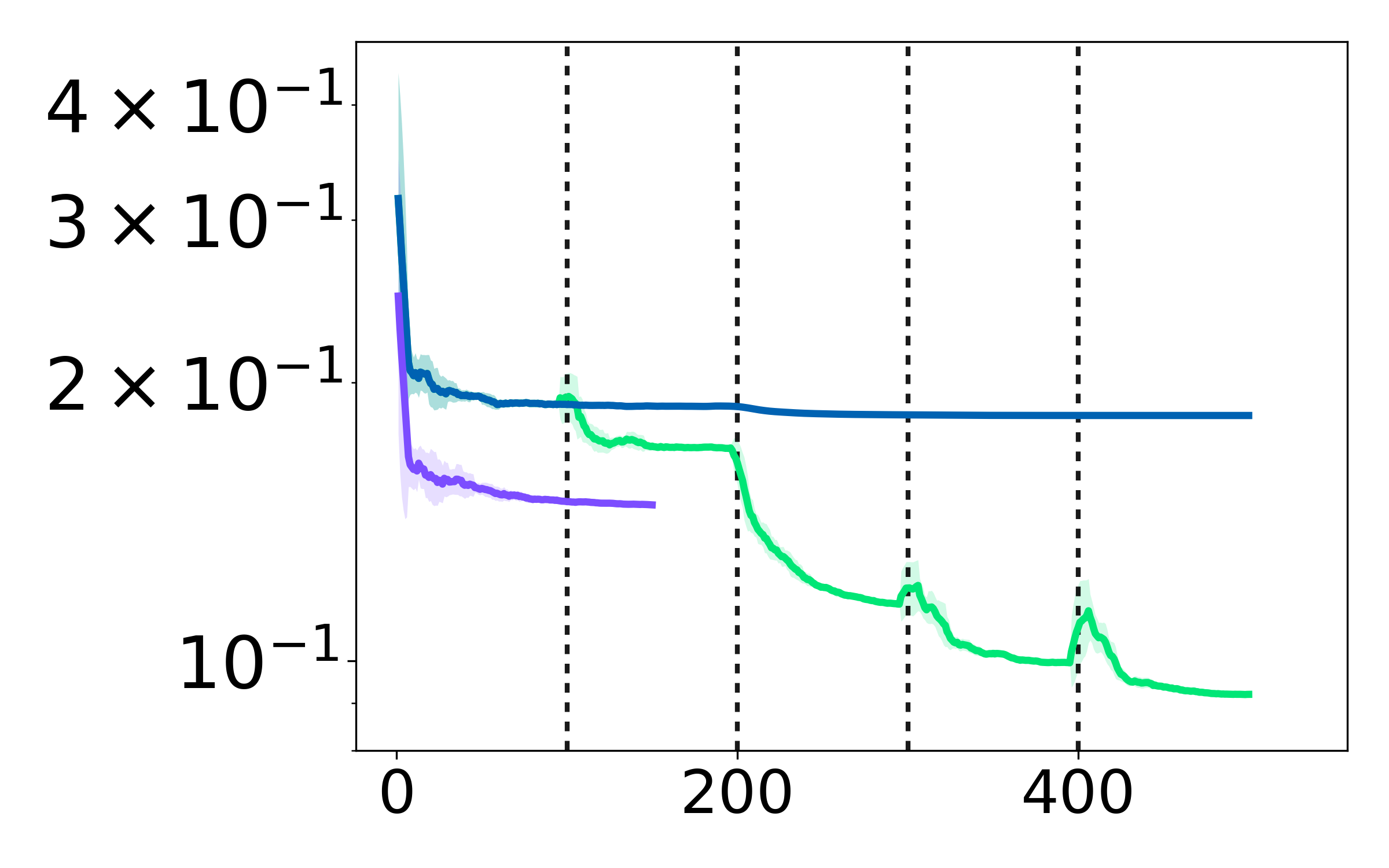}
        \caption{\textbf{PINO}: $\nu=10^{-4}$}
    \end{subfigure}\hfill
    \begin{subfigure}[t]{0.32\textwidth}
        \centering
        \includegraphics[width=\linewidth]{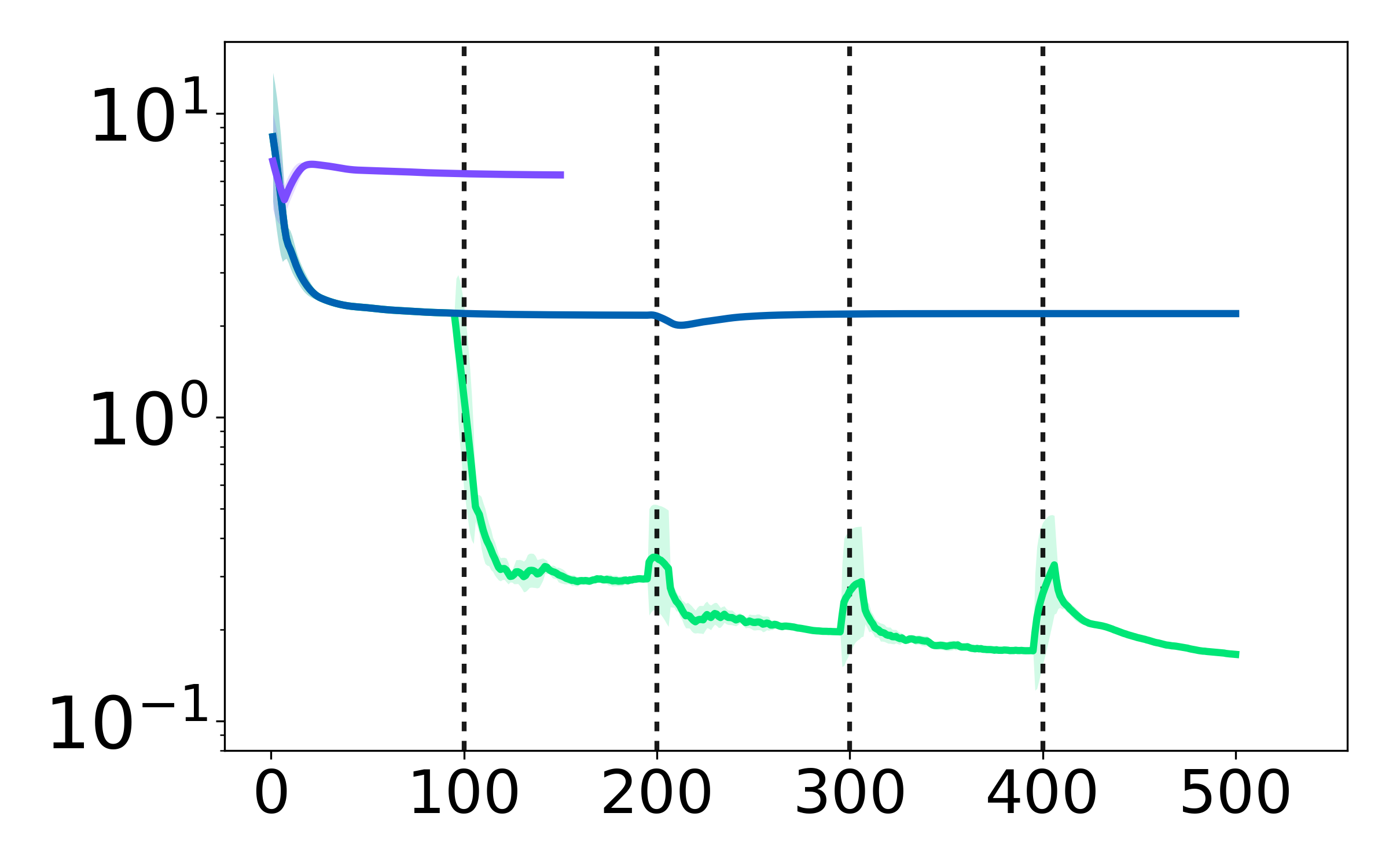}
        \caption{\textbf{UNet}: $\nu=10^{-4}$}
    \end{subfigure}
    \begin{subfigure}[t]{0.32\textwidth}
        \centering
        \includegraphics[width=\linewidth]{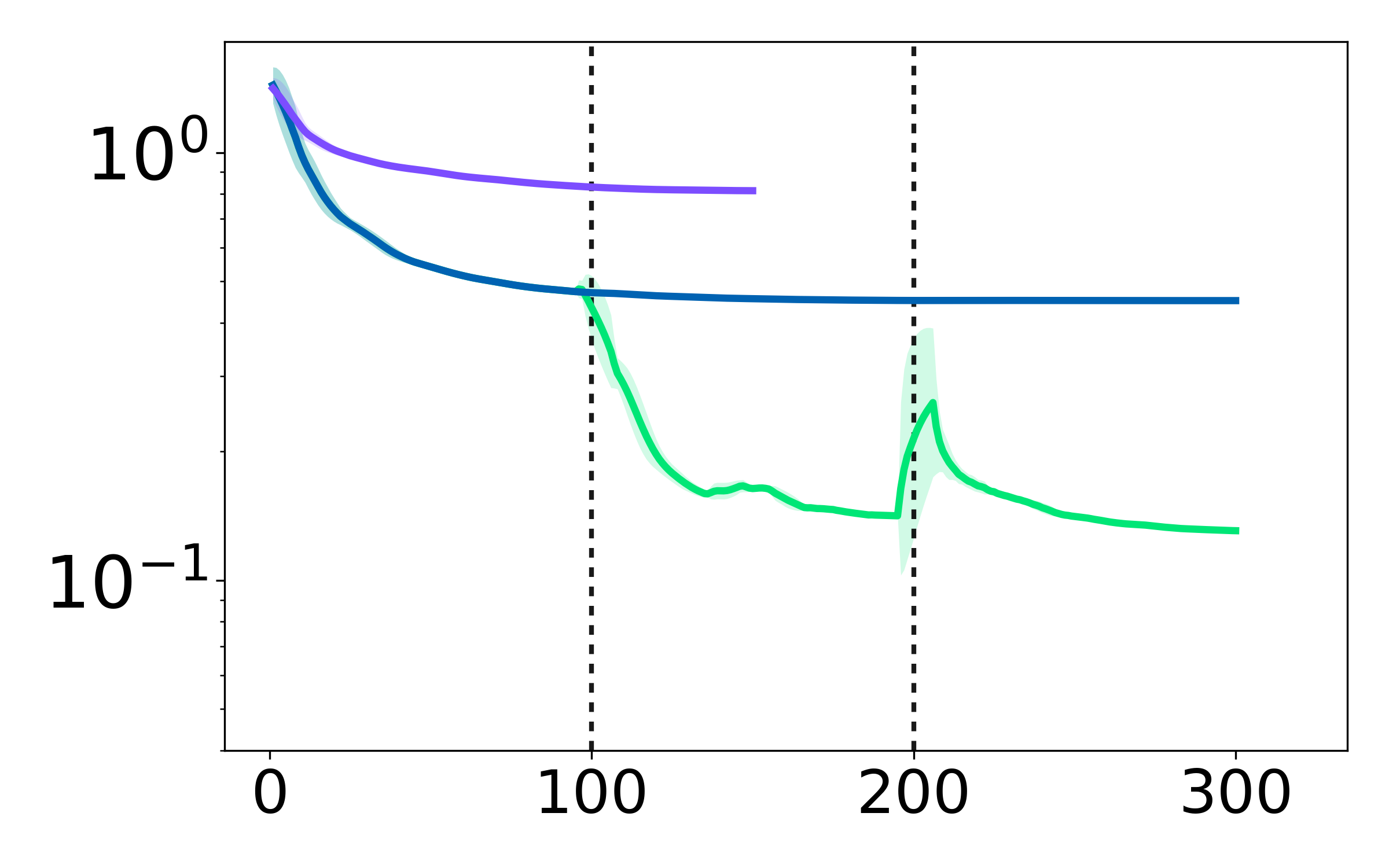}
        \caption{\textbf{PhIS-FNO}: Kolmogorov}
    \end{subfigure}\hfill
    \begin{subfigure}[t]{0.32\textwidth}
        \centering
        \includegraphics[width=\linewidth]{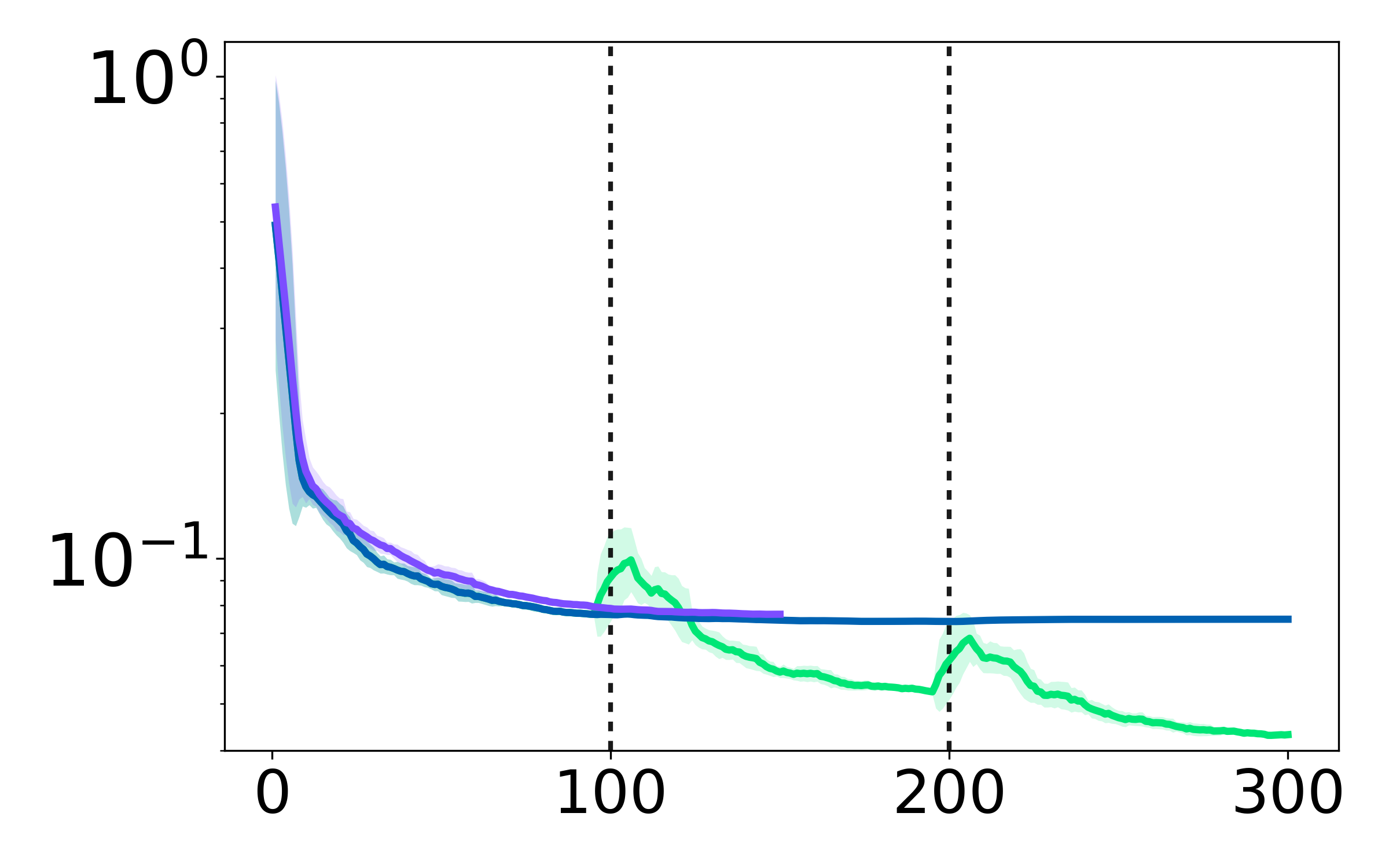}
        \caption{\textbf{PINO}: Kolmogorov}
    \end{subfigure}
    \begin{subfigure}[t]{0.32\textwidth}
        \centering
        \includegraphics[width=\linewidth]{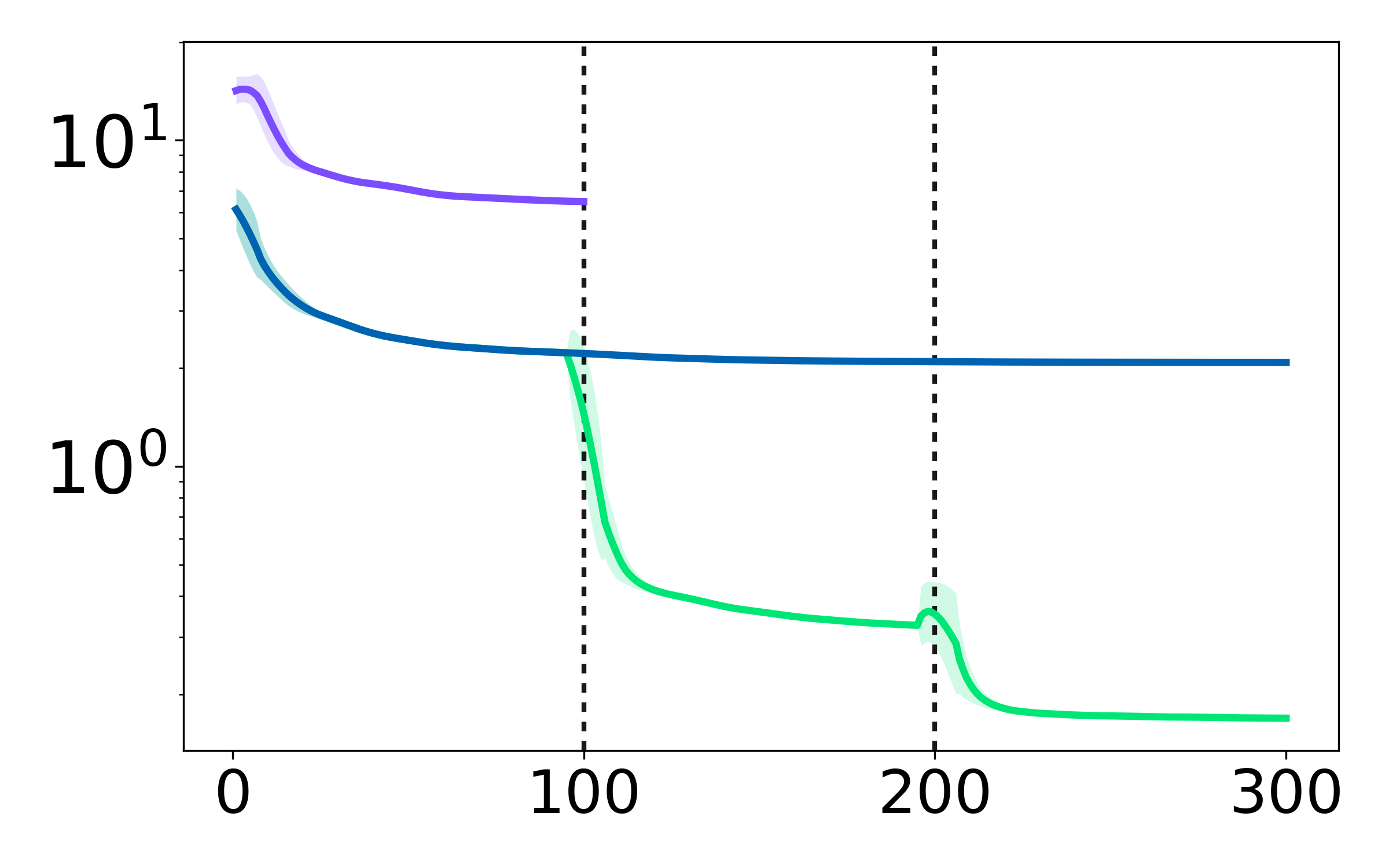}
        \caption{\textbf{UNET}: Kolmogorov}
    \end{subfigure}
    \caption{\textbf{Comparison of training strategies for 2D Navier--Stokes and Kolmogorov flow.} 
    Each panel shows the validation $L_2$ loss versus training epochs across the full forecasting horizon $T$ 
    for three architectures (PhIS-FNO, PINO, and UNet), two viscosity regimes 
    $\nu=10^{-3}$ (top row) and $\nu=10^{-4}$ (middle row) and Kolmogorov flow (bottom row).
    The green curves correspond to the multi-stage (M-S) configuration with optimizer reset, 
    the blue curves to multi-stage without reset (M-S nr), 
    and the purple curves to the single-stage (S-S) setup with $\lambda_{\mathrm{res}}{=}1$. For numerical values of convergence see Table \ref{tab:ns_kolm_all}.
    Across all models and viscosities, the reset-based multi-stage strategy achieves the most stable convergence and lowest validation error, 
    confirming its effectiveness in enforcing PDE consistency during training.}
    \label{fig:NS_viscosity_comparison}
\end{figure*}

\begin{table}[htbp]
\captionsetup{font=footnotesize, skip=4pt}
\centering
\renewcommand{\arraystretch}{1.15}
\setlength{\tabcolsep}{8pt}

\begin{minipage}{\linewidth}
\centering
\text{Burgers 1D}\\[2pt]
\resizebox{\linewidth}{!}{%
\begin{tabular}{lcccc}
\toprule
$\nu=10^{-1}$ & \textbf{MS} & \textbf{MS (n-r)} & \textbf{SS} & \textbf{Sup} \\
\midrule
\textbf{PhIS-FNO} 
& $0.0616 \pm 9.62\times10^{-4}$ 
& $0.478 \pm 8.62\times10^{-4}$ 
& $0.433 \pm 1.89\times10^{-3}$ 
& $0.0376 \pm 2.66\times10^{-3}$ \\
\bottomrule
\end{tabular}%
}
\end{minipage}

\vspace{6pt}

\begin{minipage}{\linewidth}
\centering
\text{Navier--Stokes}\\[2pt]
\resizebox{\linewidth}{!}{%
\begin{tabular}{lcccc}
\toprule
$\nu=10^{-3}$ & \textbf{MS} & \textbf{MS (n-r)} & \textbf{SS} & \textbf{Sup} \\
\midrule
\textbf{PhIS-FNO} 
& $0.0602 \pm 8.183\times10^{-5}$ 
& $0.4460 \pm 2.109\times10^{-3}$ 
& $0.8103 \pm 9.886\times10^{-3}$ 
& $0.0589 \pm 2.299\times10^{-4}$ \\
\textbf{PINO} 
& $0.0578 \pm 1.149\times10^{-3}$ 
& $0.0983 \pm 8.835\times10^{-4}$ 
& $0.0749 \pm 4.251\times10^{-4}$ 
& $0.0290 \pm 2.80\times10^{-4}$ \\
\textbf{UNet} 
& $0.1086 \pm 6.798\times10^{-4}$ 
& $3.212 \pm 4.142\times10^{-2}$ 
& $4.846 \pm 2.574\times10^{-2}$ 
& $0.0690 \pm 1.455\times10^{-3}$ \\
\bottomrule
\end{tabular}%
}
\end{minipage}

\vspace{6pt}

\begin{minipage}{\linewidth}
\centering
\resizebox{\linewidth}{!}{%
\begin{tabular}{lcccc}
\toprule
 $\nu=10^{-4}$ & \textbf{MS} & \textbf{MS (n-r)} & \textbf{SS} & \textbf{Sup} \\
\midrule
\textbf{PhIS-FNO} 
& $0.1053 \pm 6.208\times10^{-4}$ 
& $0.2488 \pm 7.947\times10^{-4}$ 
& $0.7456 \pm 4.409\times10^{-3}$ 
& $0.1231 \pm 4.739\times10^{-4}$ \\
\textbf{PINO} 
& $0.0920 \pm 1.702\times10^{-4}$ 
& $0.1853 \pm 1.552\times10^{-3}$ 
& $0.1477 \pm 2.100\times10^{-4}$ 
& $0.0948 \pm 2.649\times10^{-4}$ \\
\textbf{UNet} 
& $0.1680 \pm 1.490\times10^{-3}$ 
& $2.188 \pm 1.264\times10^{-2}$ 
& $6.371 \pm 7.735\times10^{-2}$ 
& $0.1493 \pm 1.332\times10^{-3}$ \\
\bottomrule
\end{tabular}%
}
\end{minipage}

\vspace{6pt}

\begin{minipage}{\linewidth}
\centering
\text{Kolmogorov flow }\\[2pt]
\resizebox{\linewidth}{!}{%
\begin{tabular}{lcccc}
\toprule
 $\nu=2\times10^{-3}$ & \textbf{MS} & \textbf{MS (n-r)} & \textbf{SS} & \textbf{Sup} \\
\midrule
\textbf{PhIS-FNO} 
& $0.1313 \pm 5.640\times10^{-4}$ 
& $0.4561 \pm 8.362\times10^{-3}$ 
& $0.8253 \pm 1.023\times10^{-2}$ 
& $0.0297 \pm 7.397\times10^{-5}$ \\
\textbf{PINO} 
& $0.0433 \pm 7.011\times10^{-4}$ 
& $0.0746 \pm 3.213\times10^{-4}$ 
& $0.0768 \pm 3.137\times10^{-4}$ 
& $0.0504 \pm 1.621\times10^{-4}$ \\
\textbf{UNet} 
& $0.1698 \pm 5.695\times10^{-4}$ 
& $2.136 \pm 7.011\times10^{-2}$ 
& $6.549 \pm 4.893\times10^{-2}$ 
& $0.4058 \pm 4.826\times10^{-3}$ \\
\bottomrule
\end{tabular}%
}
\end{minipage}

\vspace{6pt}

\begin{minipage}{\linewidth}
\centering
\text{Cylinder Wake flow}\\[2pt]
\resizebox{\linewidth}{!}{%
\begin{tabular}{lcccc}
\toprule
 $\nu = 10^{-2}$& \textbf{MS} & \textbf{MS (n-r)} & \textbf{SS} & \textbf{Sup} \\
\midrule
\textbf{PhIS-FNO} 
& $0.0512 \pm 3.312\times10^{-5}$ 
& $0.6521 \pm 1.258\times10^{-2}$ 
& $0.1560 \pm 2.028\times10^{-4}$ 
& $0.0093 \pm 2.088\times10^{-4}$ \\
\textbf{PINO} 
& $0.1222 \pm 1.469\times10^{-2}$ 
& $0.1601 \pm 3.658\times10^{-2}$ 
& $0.3597 \pm 2.169\times10^{-2}$ 
& $0.0068 \pm 9.672\times10^{-5}$ \\
\bottomrule
\end{tabular}%
}
\end{minipage}

\caption{\textbf{Ablation study on the multi-stage training strategy across PDE benchmarks.}
Validation $L_2$ loss (mean $\pm$ s.d.) for different PDEs and training strategies: 
values are computed on the final plateau (last 100 epochs).
MS \(=\) multi-stage with optimizer reset; MS (n-r) keeps Adam states across stages;
SS \(=\) single-stage with $\lambda_{\mathrm{res}}{=}1$; Sup \(=\) supervised training.}
\label{tab:ns_kolm_all}
\end{table}

To maintain consistency with the numerical solver used to generate the dataset, the PDE residual is computed in Fourier space with the same dealiasing rule ($2/3$ truncation), ensuring that the physics-informed loss enforces the equations under identical numerical assumptions. 
As in previous experiments, training follows a MS strategy: initial stages emphasize boundary supervision, while later ones progressively increase the weight of the residual term, leading to stable convergence and accurate reproduction of the nonlinear dynamics.  \\
The results in Fig.~\ref{fig:NS_viscosity_comparison} and Table \ref{tab:ns_kolm_all} clearly demonstrate the effectiveness of the proposed MS training strategy with optimizer reset across all tested architectures and viscosity regimes. Reported values correspond to the mean and standard deviation computed over the final training epochs, once the loss has reached a stable regime.
In every case, the reset-based MS configuration yields a markedly faster and more stable convergence, achieving significantly lower validation losses and smaller variance bands compared to both the S-S and the nr variants. 
Remarkably, in several settings the unsupervised models trained with the proposed curriculum reach the accuracy of their fully supervised counterparts. We chose not to include these curves in the plots to keep the focus on the effectiveness of the multi-stage unsupervised training strategy relative to standard training regimes, and to make the plots easier to interpret.

Among the compared architectures, PhIS-FNO and PINO achieve the highest overall accuracy.
At higher turbulence levels, smaller viscosity values, PINO performs slightly better on periodic Navier Stokes benchmarks, where Fourier continuation naturally aligns with the boundary conditions.
All models, including UNet, exhibit consistent improvements when trained with the proposed multi-stage scheme. 
This highlights the generality of the proposed training paradigm, which proves effective regardless of architectural design or parameterization of the underlying differential operators as shown in Sec. \ref{lambda_config}. 
Furthermore, experiments on the supervised UNet, to get the supervised baseline, confirm that the multi-stage approach remains beneficial even in the presence of labeled data, suggesting that the curriculum enhances optimization stability and physical consistency beyond the unsupervised regime. 
To the best of our knowledge, this is the first demonstration of multiple neural operator architectures achieving stable convergence in a fully unsupervised setting, thus establishing a new foundation for generalizable physics-informed operator learning.\\

\begin{figure*}[htbp]
    \centering
    \begin{subfigure}[t]{0.32\textwidth}
        \centering
        \includegraphics[width=\linewidth]{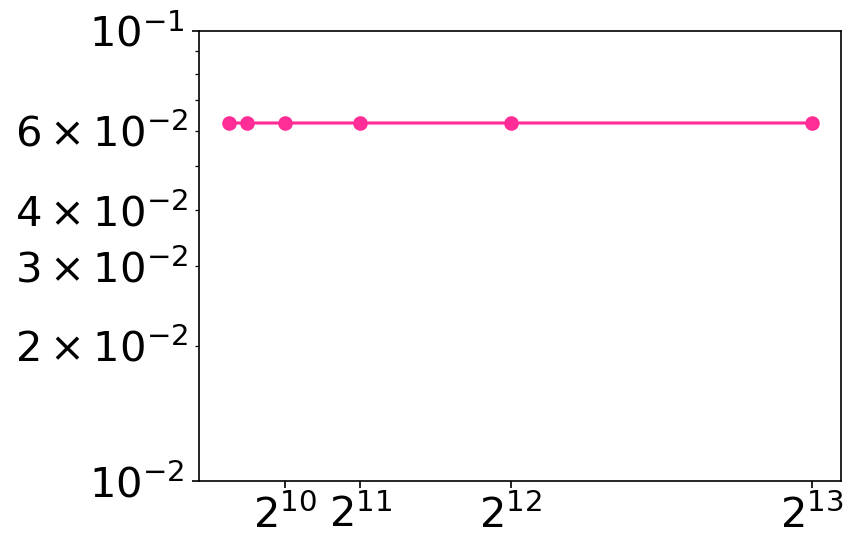}
        \caption{Burgers' equation}
    \end{subfigure}\hfill
    \begin{subfigure}[t]{0.32\textwidth}
        \centering
        \includegraphics[width=\linewidth]{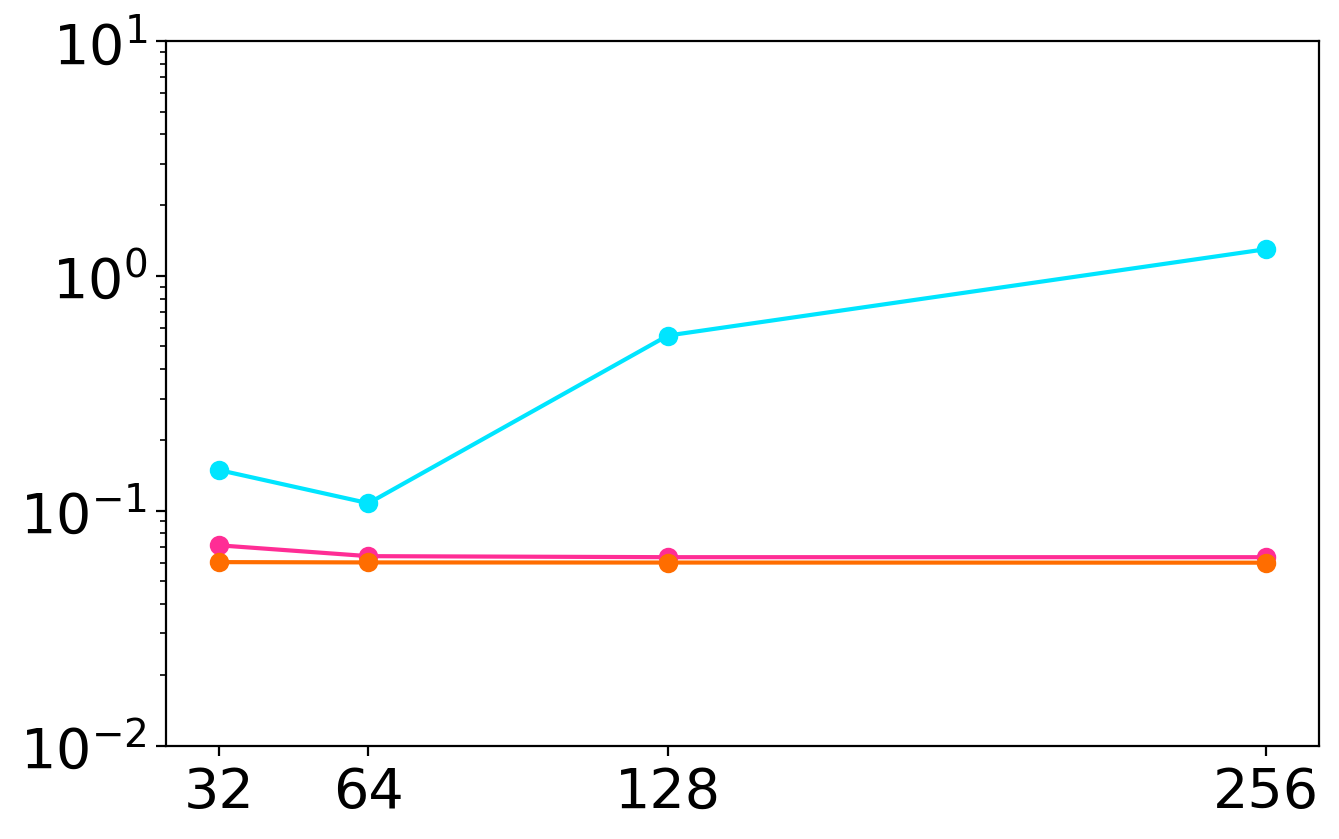}
        \caption{Navier--Stokes, $\nu=10^{-3}$}
    \end{subfigure}\hfill
    \begin{subfigure}[t]{0.32\textwidth}
        \centering
        \includegraphics[width=\linewidth]{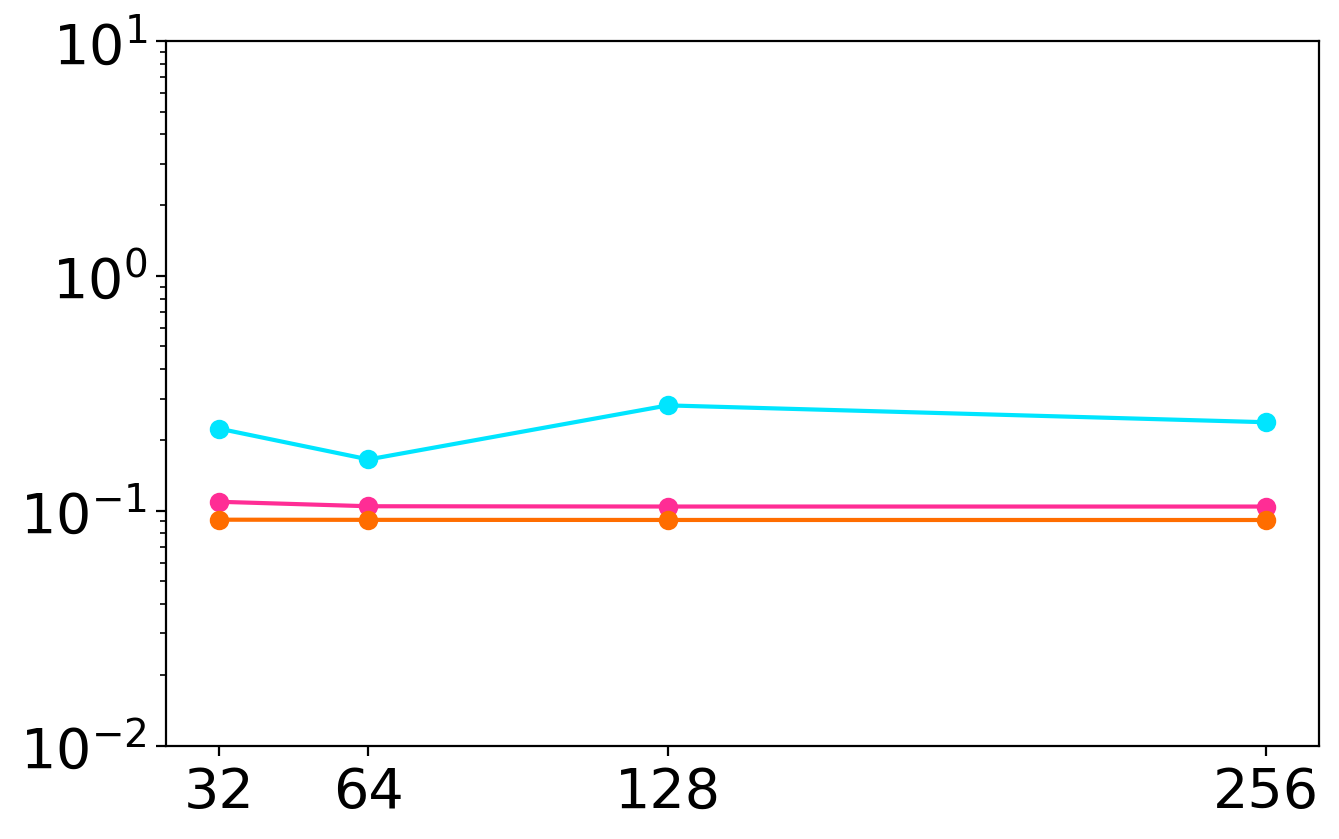}
        \caption{Navier--Stokes, $\nu=10^{-4}$}
    \end{subfigure}
    \caption{\textbf{Resolution-invariance of neural operator architectures.}
    Relative error across spatial resolutions for different neural operator models:
    PINO (orange), PhIS--FNO (magenta), and UNet (cyan).
    Panels show results for (\textbf{a}) the 1D Burgers’ equation and
    (\textbf{b,c}) the 2D Navier--Stokes equations with viscosities $\nu=10^{-3}$ and $\nu=10^{-4}$.
    PINO and PhIS--FNO maintain stable accuracy across all grid sizes, indicating strong mesh invariance,
    whereas UNet exhibits a sharp degradation when evaluated away from the training resolution.
    These results highlight the ability of Fourier-based operator networks to generalize in a zero-shot manner to unseen spatial scales.}
    \label{resolution}

\end{figure*}

\noindent
\textbf{Resolution Invariance.} 
A key property of neural operators is their intrinsic mesh invariance, enabling them to generalize seamlessly across spatial resolutions. 
We exploit this property to assess the super-resolution capability of PhIS-FNO, PINO, and UNet, training all models on coarse grids and evaluating them on increasingly finer meshes without retraining. 
Specifically, PhIS-FNO is trained on the one-dimensional Burgers’ equation with $2^{11}$ points and evaluated up to $2^{13}$ points, while for the two-dimensional Navier Stokes equations the training grid is $64\times64$ with testing performed up to $256\times256$. 
As shown in Fig.~\ref{resolution}, both the PINO and the PhIS-FNO exhibit nearly constant relative errors across all resolutions, confirming their strong resolution invariance. 
In contrast, the UNet displays a pronounced degradation when evaluated outside its training resolution, indicating that its convolutional structure lacks the spectral continuity required for true operator generalization. 
While Fourier-based models are also time-resolution independent, this aspect cannot be fully leveraged here because the PDE residual involves a time derivative estimated from discrete snapshots, making it sensitive to large temporal spacing. 
Overall, PhIS-FNO retains accurate, resolution-invariant predictions even in the unsupervised regime, highlighting its potential for zero-shot super-resolution in scientific operator learning.

\subsection{Cylinder Wake Flow (Non-Periodic Setting)}

We further evaluate the proposed framework on the incompressible flow past a circular cylinder, 
a well-known benchmark characterized by rich nonlinear dynamics and periodic vortex shedding in the wake region.  
The setup follows~\citep{RAISSI2019686}, with kinematic viscosity $\nu = 0.01$, 
cylinder diameter $D = 1$, and free-stream velocity $u_\infty = 1$, 
corresponding to a Reynolds number $Re = u_\infty D / \nu = 100$.
Here, $Re$ denotes the (dimensionless) Reynolds number obtained after non-dimensionalization of the governing equations. 
For simplicity, we restrict the computational domain to a rectangular region downstream of the cylinder, 
where the flow develops complex unsteady patterns due to the asymmetric Kármán vortex street.\\
Unlike the periodic Navier Stokes tests, this configuration involves non-periodic boundary conditions and strong spatial gradients near solid walls.  
To ensure training stability and isolate the effects of non-periodicity, the pressure field $p(x,y,t)$ is supervised across the entire domain to provide a physically consistent reference signal, 
while the velocity components $u(x,y,t)$ and $v(x,y,t)$ are constrained only along the boundary.  
This semi-supervised setup is intentionally designed to focus the analysis on how different differentiation schemes, spectral versus spline-based, handle non-periodic boundaries.  
The model learns the temporal evolution operator mapping the flow state at $t$ to $t+\Delta t$, with $\Delta t = 0.1$.\\
While PINO computes spatial derivatives in Fourier space through Fourier continuation,
PhIS-FNO performs differentiation via spline-based convolutions, which are inherently local and naturally compatible with arbitrary geometries and boundary conditions.  
This design eliminates the need for periodic extension and allows accurate gradient estimation up to the physical boundary.

\begin{figure*}[htbp]
    \centering
    \begin{subfigure}[t]{0.32\textwidth}
        \centering
        \includegraphics[width=\linewidth]{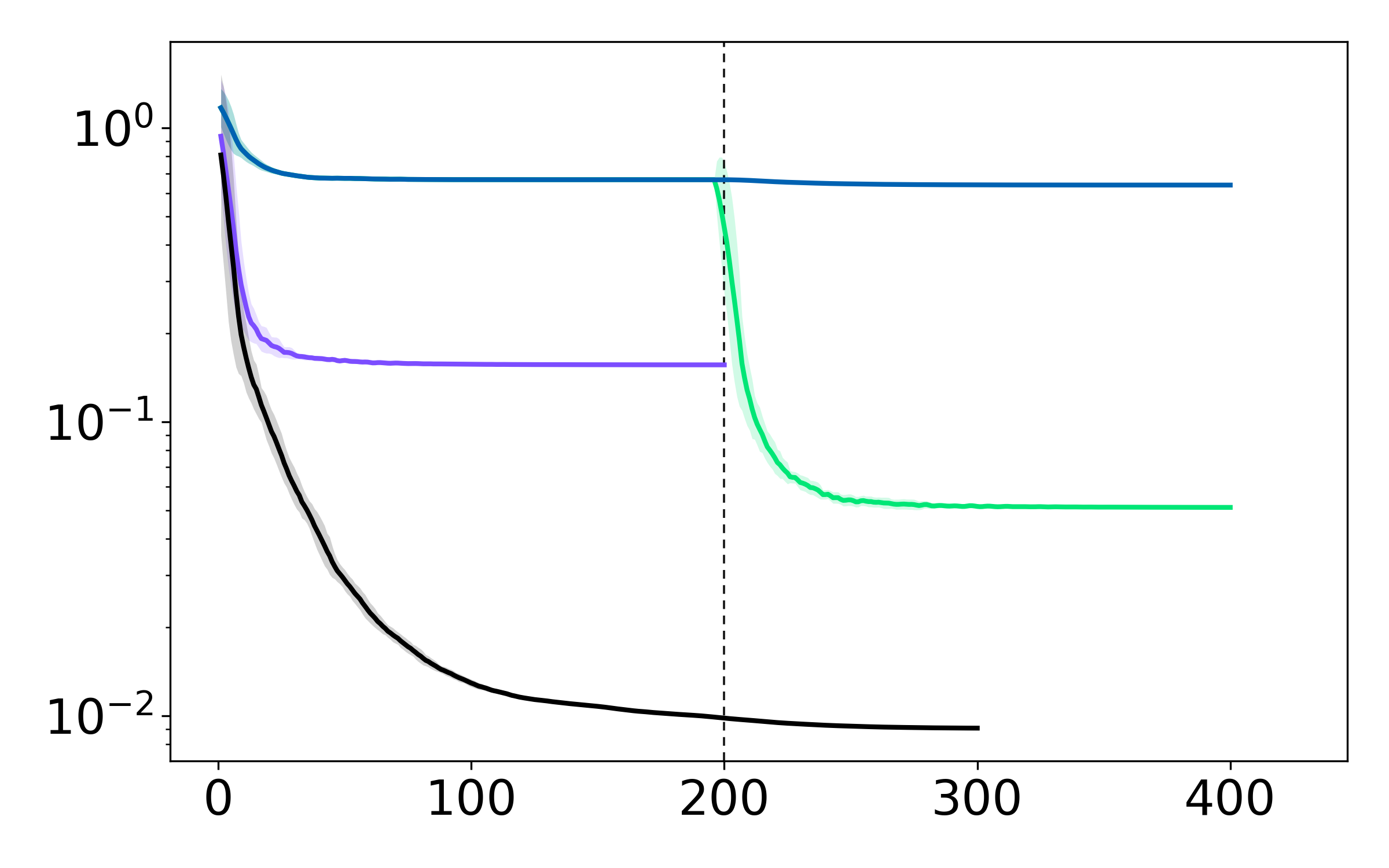}
        \caption{\textbf{PhIS-FNO}}
    \end{subfigure}\hfill
    \begin{subfigure}[t]{0.32\textwidth}
        \centering
        \includegraphics[width=\linewidth]{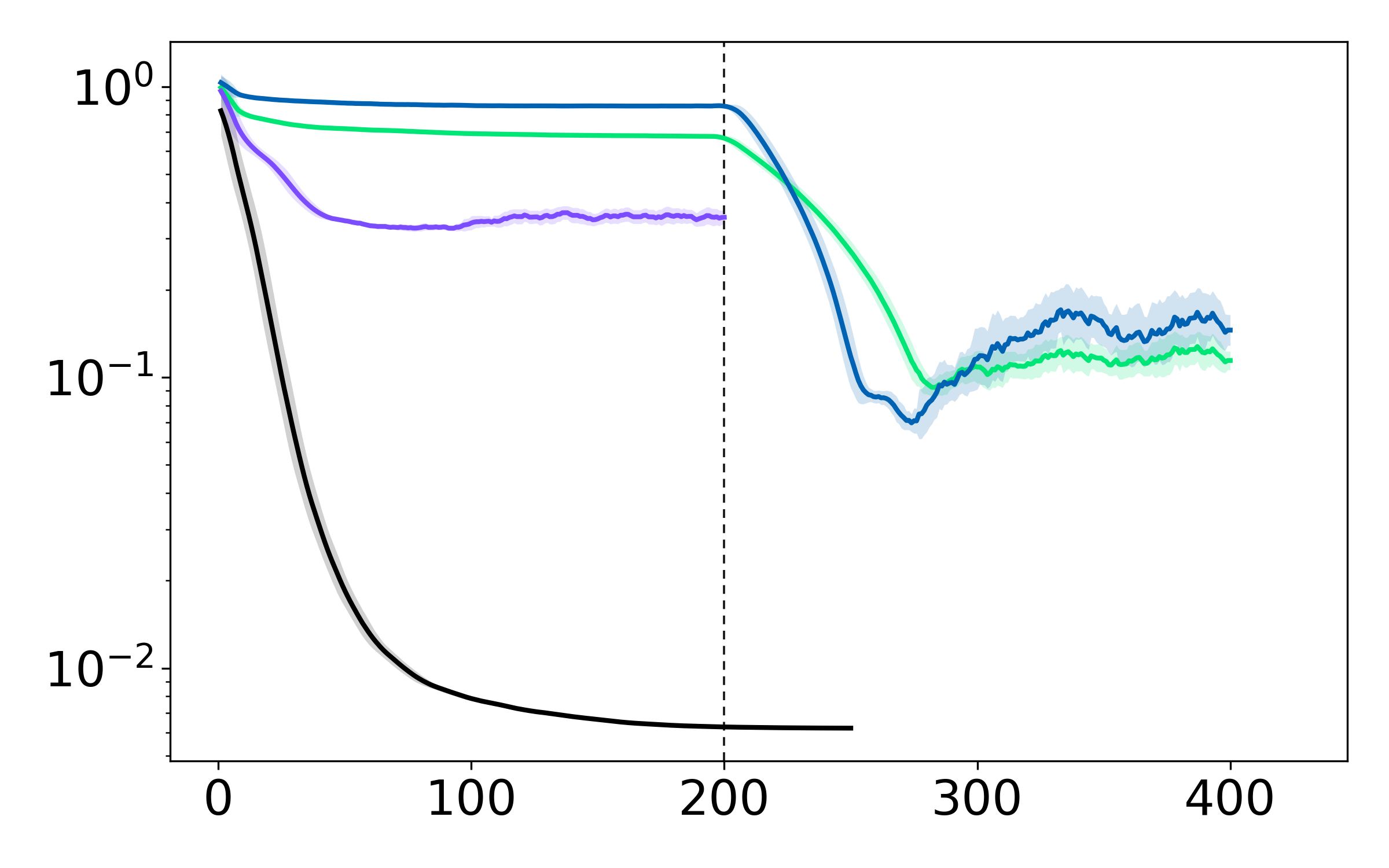}
        \caption{\textbf{PINO}}
    \end{subfigure}\hfill
    \begin{subfigure}[t]{0.32\textwidth}
        \centering
        \includegraphics[width=\linewidth]{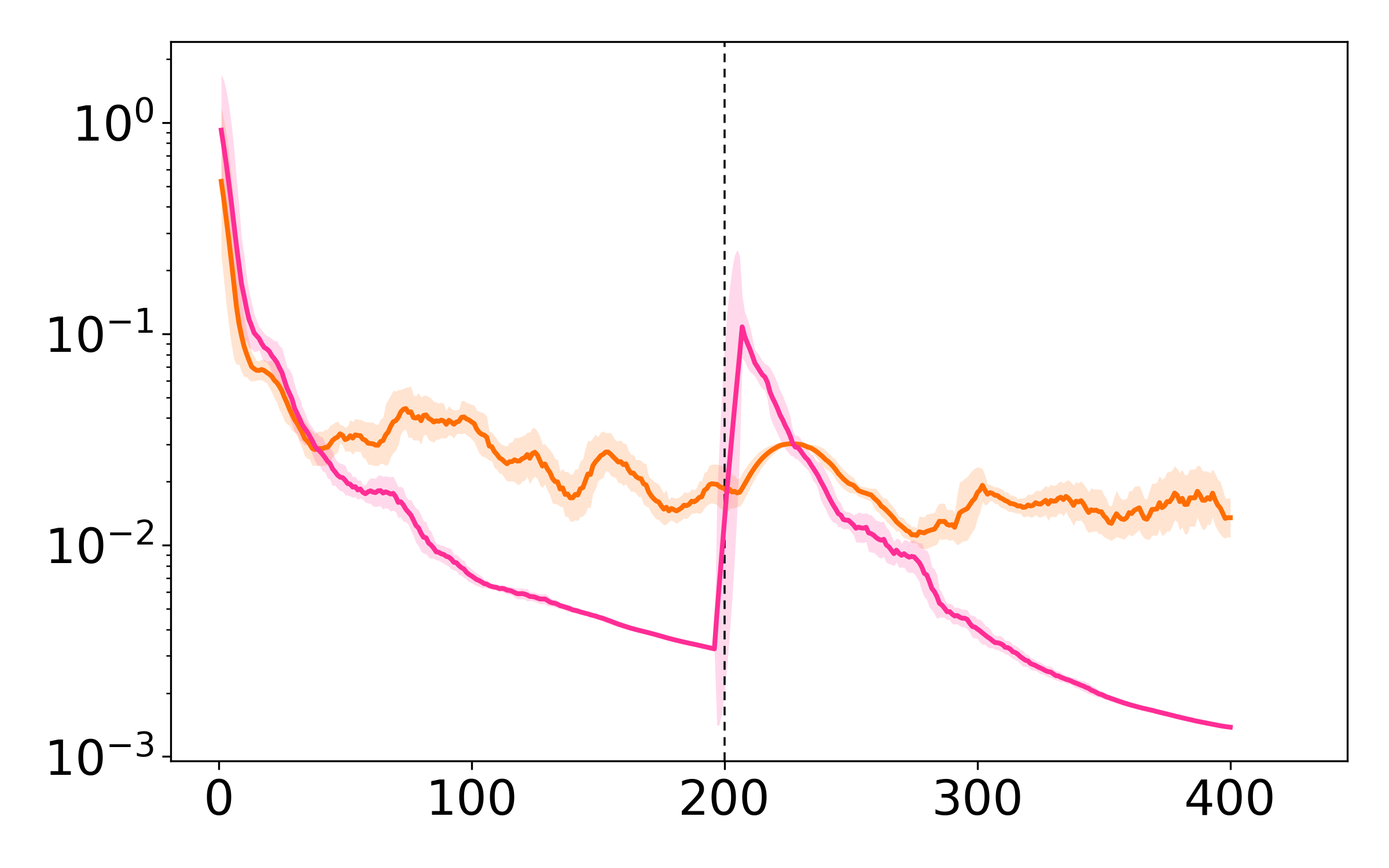}
        \caption{\textbf{Boundary loss evolution during multi-stage training for PINO (orange curve), and PhIS-FNO (magenta curve)}}
    \end{subfigure}
    
    \caption{\textbf{Cylinder wake flow.} 
    Comparison between PhIS-FNO and PINO on the non-periodic cylinder wake benchmark. 
    Panels~(a,b) show the validation $L_2$ loss versus training epochs for both models under different training strategies: 
    green = MS with optimizer reset, 
    blue = MS nr, 
    purple = SS, and 
    black = fully supervised training. 
    Panel~(c) reports the boundary loss evolution during multi-stage training 
    magenta: PhIS-FNO, orange: PINO). 
    The reset-based multi-stage strategy yields faster and more stable convergence across both models, 
    with PhIS-FNO achieving a markedly lower boundary loss 
    ($1.44\times10^{-3}\,\pm\,4.59\times10^{-5}$) compared to PINO 
    ($1.66\times10^{-2}\,\pm\,4.79\times10^{-3}$).}
    
    \label{fig:cylinder}
\end{figure*}

\medskip

The results in Fig.~\ref{fig:cylinder} and Tab.~\ref{tab:ns_kolm_all} show that 
the MS strategy significantly accelerates convergence and reduces variance relative to both the MS nr and SS setups, 
achieving performance comparable to the fully supervised reference.  
In the non-periodic setting, PhIS-FNO exhibits higher boundary consistency and a lower final loss plateau compared to PINO.  
As evident from Fig.~\ref{fig:cylinder}(c), the boundary loss of PINO remains substantially higher throughout training, 
indicating that the spectral differentiation based on Fourier continuation struggles to represent local gradient information near the boundaries.  
This limitation propagates inward, leading to poorer PDE residual consistency within the flow domain.  
Conversely, the spline formulation of PhIS-FNO preserves smooth derivative continuity up to the boundary, 
allowing the model to accurately recover both velocity and pressure fields in complex, non-periodic geometries.

\section{Discussion}

The proposed multi-stage curriculum introduces a robust and general framework for physics-informed optimization, 
independent of the underlying neural architecture. 
By progressively transferring the loss weighting from boundary consistency to interior residual minimization, 
the training dynamics follows a smooth continuation path that mirrors the mathematical well-posedness of PDEs. 
At each transition, reinitializing the Adam optimizer resets its internal moment statistics, 
restoring a high effective learning rate and preventing stagnation that typically arises 
when accumulated momentum misaligns with the new gradient directions. 
This mechanism provides a principled interpretation of optimizer resets as a continuation process in the loss landscape, 
allowing the network to escape local minima and maintain convergence stability across stages and random parameters initialization.

Importantly, the multi-stage training scheme is independent on the architecture and thus applicable to a wide variety of operator-learning frameworks. 
Rather than being tied to a specific representation of the solution operator, 
it defines a general optimization protocol for enforcing physical consistency 
in both unsupervised and supervised regimes. 
As a future perspective, it will be of particular interest to investigate how this staged optimization 
interacts with different architectural families, such as convolutional, graph-based, or transformer-based operators,
and whether similar continuation effects emerge in their optimizer dynamics. Moreover, extending the curriculum principle to the temporal dimension, by training over progressively longer prediction horizons, may further improve stability and accuracy in fully autoregressive, unsupervised simulations.

The introduced Physics-Informed Spline Fourier Neural Operator 
unifies the spectral expressiveness of Fourier layers with the continuous differentiability of spline kernels, 
enabling smooth residual computation on both periodic and non-periodic domains. 
This hybrid formulation overcomes a key limitation of the original PINO, 
which performs optimally only under periodic boundary conditions. 
When trained under the multi-stage curriculum, PhIS-FNO achieves stable convergence, 
reduced variance across random seeds, and strong cross-resolution generalization, an essential property 
for scalable and adaptive physical simulations.

While the framework has been validated on various PDEs including Burgers’, Poisson, and Navier Stokes equations at multiple viscosities, 
its modular design readily extends to three-dimensional PDEs, adaptive mesh discretizations, 
and hybrid neural–numerical solvers. 
Given its compatibility with diverse operator learning architectures, 
the proposed approach opens the path toward unified, architecture-independent training strategies 
for physics-constrained learning in complex scientific domains.

Overall, these findings establish multi-stage, physics-informed optimization 
as a general bridge between continuation methods and adaptive gradient dynamics. 
By coupling this training paradigm with spline-based or other differentiable operator architectures, 
this work provides a scalable foundation for stable, unsupervised, and resolution-independent PDE learning.

\section*{Acknowledgements}

\noindent Financial support from ICSC – “National Research Centre in High Performance Computing, Big Data and Quantum Computing”, funded by the European Union – NextGenerationEU, is gratefully acknowledged.

\smallskip
\noindent The authors also acknowledge the use of large language model (LLM) tools, in particular OpenAI's ChatGPT, which was employed to assist in language refinement, editing, and clarity improvements of the manuscript text. 
All conceptual content, analyses, and interpretations were produced entirely by the authors.

\section*{Author Contributions}

\noindent P.M. conceived the research idea, developed the methodology, implemented the models,
performed all experiments, and wrote the manuscript.
Martina Siena and Stefano Mariani provided financial support for the research.
All co-authors contributed to the scientific discussion and provided guidance and feedback
throughout the research process and the preparation of the manuscript.

\appendix
\section{Appendix}

\subsection{Multi-Stage Training Diagnostics}\label{sup:multi-stage}

To provide empirical support for the theoretical analysis introduced in Sec.~\ref{curriculum}, 
we conducted a detailed diagnostic study on the one-dimensional Burgers equation.
The network was trained in three consecutive stages (\textit{Stage 1}~$\to$~\textit{Stage 2}~$\to$~\textit{Stage 3}), 
corresponding to progressive enforcement of the PDE residual loss and two optimizer reinitializations:
\begin{align*}
    \mathcal{L}_1&=0.8\mathcal{L}_{bd}+0.5\mathcal{L}_{res}\quad \text{re-initialization}\\
    \mathcal{L}_2&=0.5\mathcal{L}_{bd}+1.0\mathcal{L}_{res}\quad \text{re-initialization}\\
    \mathcal{L}_2&=0.5\mathcal{L}_{bd}+1.5\mathcal{L}_{res}.
\end{align*}
During training, the internal state of the Adam optimizer was recorded after each phase, specifically, we logged:
\begin{itemize}
    \item the bias-corrected first and second moment estimates $(m_t, v_t)$ for each layer at the end of every stage;
    \item the corresponding gradients $g_{\text{new}}$ computed at the first iteration of the subsequent stage (subscript new to make the new stage explicit);
    \item the effective learning rate term $\bar{\eta}_{\mathrm{eff}} = \eta \, m_t / (\sqrt{v_t} + \epsilon)$.
\end{itemize}
These quantities were used to verify the analytical assumptions introduced in the multi-stage optimizer analysis.

\begin{figure*}[ht!]
    \centering
    \includegraphics[width=0.8\linewidth]{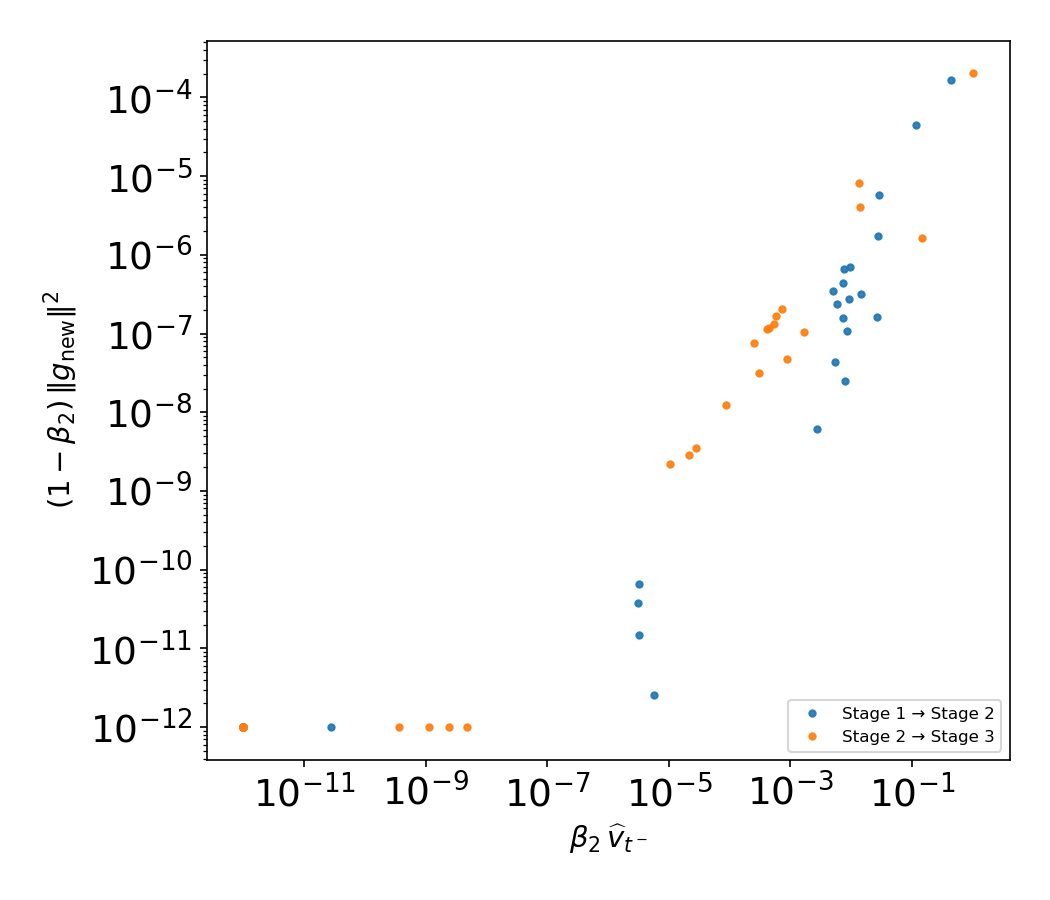}
    \caption{\textbf{Gradient and variance dominance.} 
    Comparison between $(1-\beta_2)\lVert g_{\text{new}}\rVert^2$ and $\beta_2 v_{t^-}$ across different layers and stage transitions 
    for the Burgers~1D experiment. 
    The difference by orders of magnitude between the values of the two terms confirms that $(1-\beta_2)\lVert g_{\text{new}}\rVert^2 \ll \beta_2 v_{t^-}$, 
    justifying the approximation made in Sec.~\ref{curriculum} for the non-reset case.}
    \label{check3}
\end{figure*}
\begin{figure*}[htbp]
    \centering
    \includegraphics[width=0.8\linewidth]{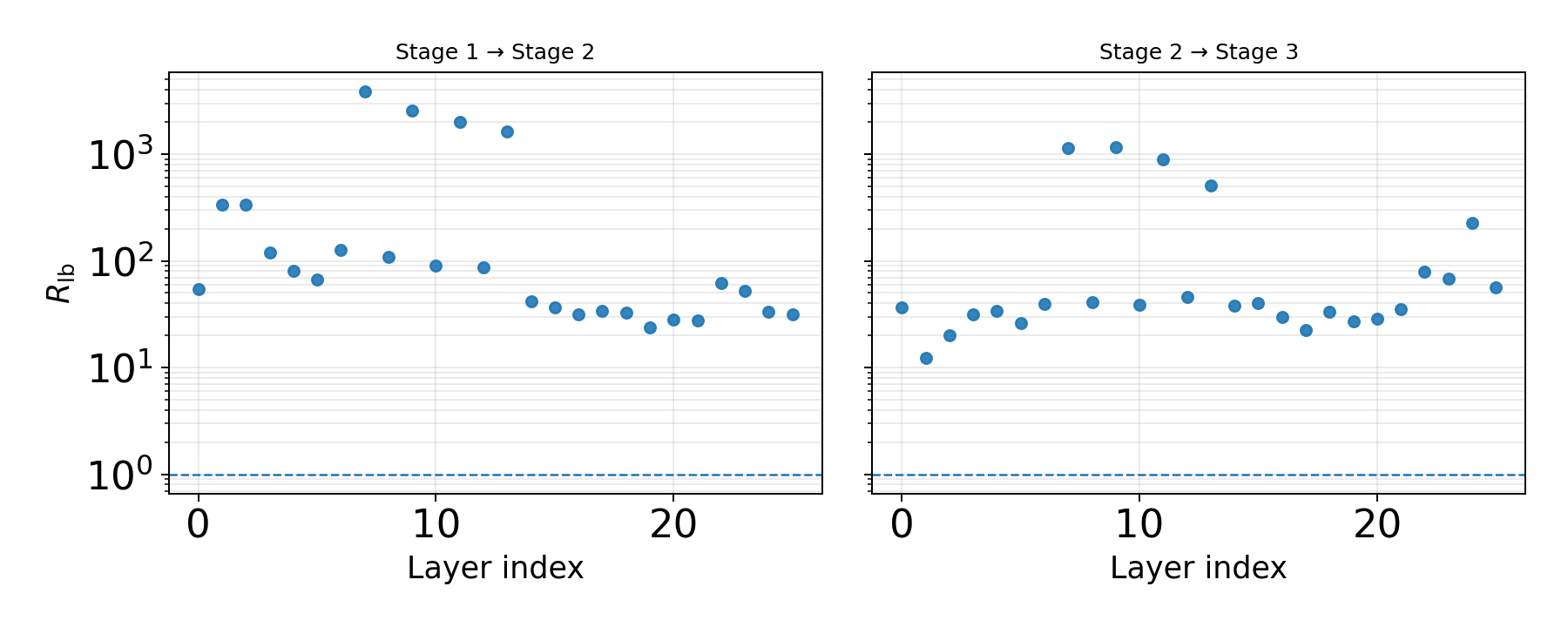}
    \caption{\textbf{Layer-wise update amplification ratio.} 
    Layer-wise values of $R_{\mathrm{lb}} = |\Delta\theta_r| / |\Delta\theta_{nr}|$ 
    for the Burgers~1D case. Each dot corresponds to one network layer, 
    computed at the transition between consecutive stages. 
    In all layers and across both transitions, $R_{\mathrm{lb}} \gg 1$, 
    indicating that optimizer reinitialization consistently amplifies 
    the effective parameter update magnitude and prevents stagnation in flat loss regions.}
    \label{Rbl}
\end{figure*}
\begin{figure*}[htbp]
    \centering
    \includegraphics[width=0.8\linewidth]{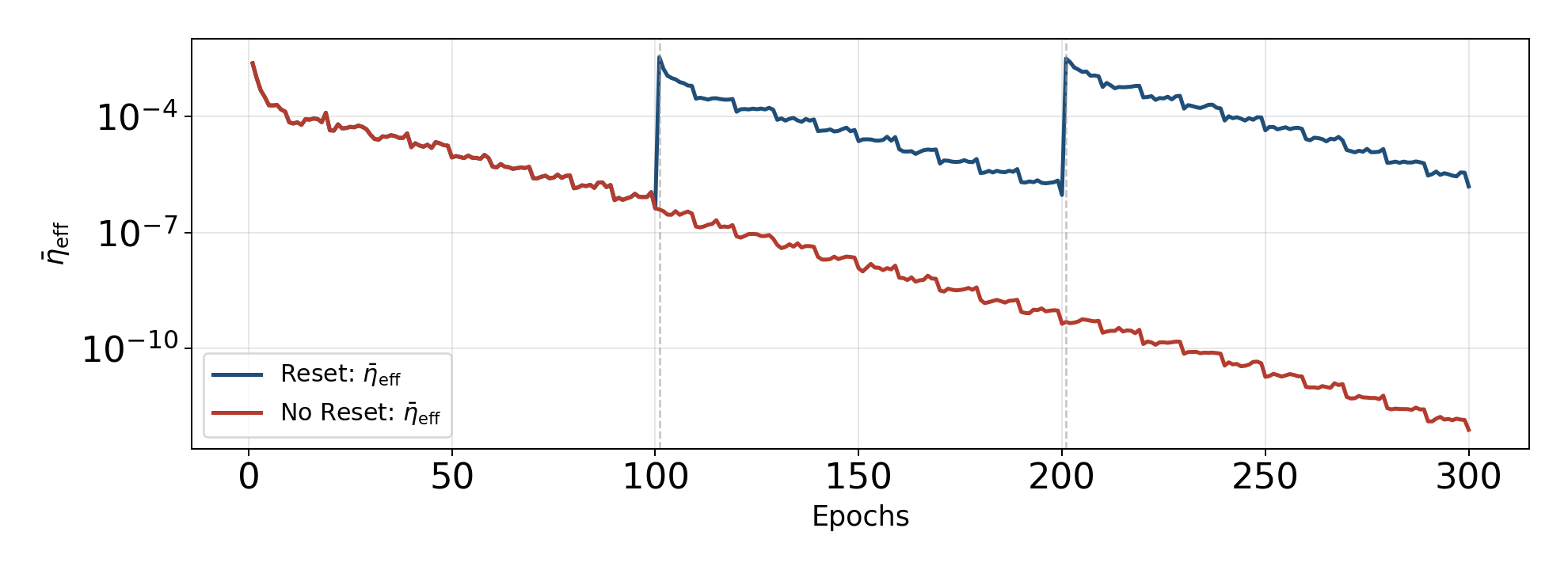}
    \caption{\textbf{Effective learning rate evolution.} 
    Time evolution of the mean effective learning rate $\bar{\eta}_{\text{eff}} = \eta \, m_t / (\sqrt{v_t}+\epsilon)$ during multi-stage training. 
    The no-reset configuration (red curve) exhibits a monotonic decay across phases, 
    leading to optimizer stagnation, whereas the reset configuration (blue curve) restores 
    high learning rates at each transition, reactivating convergence dynamics.}
    \label{etaeff}
\end{figure*}

\subsection{Training and Architectural Details}\label{sup:training}

All experiments were conducted in \texttt{PyTorch} (float64 precision) using the Adam optimizer
with $\beta_1=0.9$, $\beta_2=0.999$, and $\epsilon=10^{-8}$.  
Each curriculum stage consists of 100~epochs, with a step-wise learning-rate decay ($\gamma=0.5$ every 20~epochs).  
Models were trained in a physics-informed regime, where only boundary conditions are supervised while the interior domain is optimized via the PDE residual.  
The complete implementation, configuration files, and pre-trained checkpoints are available at  
\href{https://github.com/PaoloMarcandelli/Unsupervised-Physics-Informed-Operator-Learning-through-Multi-Stage-Curriculum-Training}{\texttt{GitHub}}.

\medskip
\noindent
\textbf{Curriculum schedule.}  
The relative weights of the boundary and residual terms $\lambda_{\mathrm{bd}}$ and $\lambda_{\mathrm{res}}$ of Eq. \ref{lambda_bdlambda_res}, were empirically tuned to balance constraint magnitudes 
and to enable a smooth transition from boundary-driven to physics-driven supervision.  
The adopted configuration, summarized in Table~\ref{tab:loss-weights}, yielded stable convergence across all PDE benchmarks.

\begin{table}[htbp]
    \centering
    \caption{\textbf{Loss weights used per stage and task.} 
    $\lambda_{\mathrm{bd}}$: boundary term; $\lambda_{\mathrm{res}}$: residual term. 
    Each phase corresponds to one curriculum stage (100 epochs).}
    \label{tab:loss-weights}
    \setlength{\tabcolsep}{4pt}
    \renewcommand{\arraystretch}{1.15}
    \begin{tabular}{ccccccccccc}
        \toprule
        & \multicolumn{2}{c}{\textbf{Poisson}} 
        & \multicolumn{2}{c}{\textbf{Burgers}}
        & \multicolumn{2}{c}{\textbf{Navier--Stokes}}
        & \multicolumn{2}{c}{\textbf{Kolmogorov}}
        & \multicolumn{2}{c}{\textbf{Cylinder Wake}}\\
        \cmidrule(lr){2-3}\cmidrule(lr){4-5}\cmidrule(lr){6-7}\cmidrule(lr){8-9}\cmidrule(lr){10-11}
        \textbf{Phase} & $\lambda_{bd}$ & $\lambda_{res}$ 
                       & $\lambda_{bd}$ & $\lambda_{res}$ 
                       & $\lambda_{bd}$ & $\lambda_{res}$ 
                       & $\lambda_{bd}$ & $\lambda_{res}$ 
                       & $\lambda_{bd}$ & $\lambda_{res}$ \\
        \midrule
        Phase 1 & 1.0 & 0.0 & 0.8 & 0.5 & 1.0 & 0.0 & 1.0 & 0.2 & 1.0 & 0.0 \\
        Phase 2 & 0.8 & 0.5 & 0.5 & 1.0 & 1.0 & 0.0 & 0.8 & 0.5 & 1.0 & 0.5 \\
        Phase 3 & 0.5 & 1.0 & 0.5 & 1.5 & 0.8 & 0.5 & 0.5 & 1.0 & -- & -- \\
        Phase 4 & -- & -- & -- & -- & 0.5 & 1.0 & -- & -- & -- & -- \\
        Phase 5 & -- & -- & -- & -- & 0.2 & 1.5 & -- & -- & -- & -- \\
        \bottomrule
    \end{tabular}
\end{table}

\medskip
\noindent
\textbf{Architectural setup.}  
All PhIS--FNO architectures comprise $L=4$ Fourier--spline operator blocks, 
combining spectral convolution and Hermite spline differentiation.  
Hyperparameters are summarized in Table~\ref{tab:arch-details}.  
The same configuration is adopted for Navier Stokes, Kolmogorov, and Cylinder wake benchmarks.

\begin{table}[htbp]
  \centering
  \caption{\textbf{Architectural and training hyperparameters for all PDE benchmarks.}
  $m$: Fourier modes; $d$: hidden width; $N_p$: trainable parameters; $\eta$: learning rate; $\gamma$: scheduler learning rate.}
  \label{tab:arch-details}
  \setlength{\tabcolsep}{6pt}
  \renewcommand{\arraystretch}{1.15}
  \begin{tabular}{lcccccccc}
  \toprule
  \textbf{Task} & $L$ & $m$ & $d$ & Batch & $\eta$ & $\gamma$ & Step & $N_p$ \\
  \midrule
  Poisson 2D          & 4 & 36 & 20 & 2  & $2\times10^{-3}$ & 0.5 & 20 & 8,302,233 \\
  Burgers 1D          & 4 & 32 & 64 & 10 & $1\times10^{-2}$ & 0.5 & 20 & 550,339 \\
  Navier--Stokes (2D) & 4 & 12 & 20 & 2  & $2\times10^{-3}$ & 0.5 & 20 & 927,549 \\
  Kolmogorov flow     & 4 & 12 & 20 & 2  & $2\times10^{-3}$ & 0.5 & 20 & 927,549 \\
  Cylinder wake flow  & 4 & 12 & 20 & 2  & $2\times10^{-3}$ & 0.5 & 20 & 927,549 \\
  \bottomrule
  \end{tabular}
\end{table}

\begin{quote}
\footnotesize
\end{quote}

\subsection{Multi-Seed analysis}\label{sup:seed}

Table~\ref{tab:multi_seed} reports the mean and standard deviation of the test errors across four different random seeds
$(0,11,22,33)$ for each training regime, where each seed corresponds to a different random initialization
of the network parameters and training procedure. 
Across all PDEs, the variability across seeds remains minimal, typically an order of magnitude smaller than the mean, corresponding to a coefficient of variation $\mathrm{CoV} = \sigma / \mu < 0.15$. 
This confirms that the staged optimization process acts as an implicit regularizer, mitigating the dependence on random initialization and guiding the optimizer toward reproducible convergence trajectories.

In contrast, both the single-stage (SS) and the no-reset variant (MS n-r) exhibit higher fluctuations, particularly in more complex or non-periodic regimes such as the cylinder wake flow, where the absence of optimizer reinitialization leads to stagnation in suboptimal minima.
Overall, these findings demonstrate that the proposed multi-stage training not only improves accuracy but also enhances robustness, ensuring statistically consistent outcomes across random seeds.

\begin{table}[htbp]
\captionsetup{font=footnotesize, skip=4pt}
\centering
\renewcommand{\arraystretch}{1.15}
\setlength{\tabcolsep}{8pt}
\begin{minipage}{\linewidth}
\centering
\text{Burger}\\[2pt]
\resizebox{\linewidth}{!}{%
\begin{tabular}{lcccc}
\toprule
  & \textbf{MS} & \textbf{MS (n-r)} & \textbf{SS} & \textbf{Sup} \\
\midrule
\textbf{PhIS-FNO} 
& $6.68\times10^{-2} \pm 8.26\times10^{-3}$
& $5.78\times10^{-1} \pm 2.50\times10^{-1}$
& $4.37\times10^{-1} \pm 1.30\times10^{-1}$
& $3.87\times10^{-2} \pm 4.38\times10^{-3}$\\
\bottomrule
\end{tabular}%
}
\end{minipage}

\vspace{6pt}

\begin{minipage}{\linewidth}
\centering
\text{Navier--Stokes}\\[2pt]
\resizebox{\linewidth}{!}{%
\begin{tabular}{lcccc}
\toprule
$\nu=10^{-3}$ & \textbf{MS} & \textbf{MS (n-r)} & \textbf{SS} & \textbf{Sup} \\
\midrule
\textbf{PhIS-FNO} 
& $6.32\times10^{-2} \pm 5.41\times10^{-3}$ 
& $7.44\times10^{-1} \pm 3.31\times10^{-1}$ 
& $6.05\times10^{-1} \pm 1.97\times10^{-1}$
& $6.47\times10^{-2} \pm 1.25\times10^{-2}$ \\
\textbf{PINO} 
& $5.51\times10^{-2} \pm 5.75\times10^{-3}$
& $1.17\times10^{-1} \pm 2.10\times10^{-2}$
& $8.01\times10^{-2} \pm 3.78\times10^{-3}$
& $3.00\times10^{-2} \pm 1.51\times10^{-3}$ \\
\bottomrule
\end{tabular}%
}
\end{minipage}

\vspace{6pt}

\begin{minipage}{\linewidth}
\centering
\text{Cylinder Wake flow}\\[2pt]
\resizebox{\linewidth}{!}{%
\begin{tabular}{lcccc}
\toprule
 $\nu = 10^{-2}$& \textbf{MS} & \textbf{MS (n-r)} & \textbf{SS} & \textbf{Sup} \\
\midrule
\textbf{PhIS-FNO} 
& $5.357\times10^{-2} \pm 6.892\times10^{-3}$
& $6.643\times10^{-1} \pm 5.245\times10^{-2}$
& $1.455\times10^{-1} \pm 2.331\times10^{-2}$
& $1.029\times10^{-2} \pm 1.213\times10^{-3}$\\
\textbf{PINO} 
& $1.210\times10^{-1} \pm 1.144\times10^{-2}$
& $1.528\times10^{-1} \pm 1.021\times10^{-2}$
& $3.543\times10^{-1} \pm 8.510\times10^{-3}$
& $6.556\times10^{-3} \pm 2.635\times10^{-4}$\\
\bottomrule
\end{tabular}%
}
\end{minipage}
\caption{Mean $\pm$ standard deviation of the test error computed across four random seeds (0, 11, 22, 33) for all training strategies: multi-stage with optimizer reset (MS), multi-stage without reset (MS n-r), single-stage training (SS), and supervised training (Sup). Results are reported for the Burgers’ equation, 2D incompressible Navier--Stokes equations ($\nu = 10^{-3}$), and the Cylinder Wake flow ($\nu = 10^{-2}$).}
\label{tab:multi_seed}
\end{table}

\subsection{Multi-Stage Ablation Study} \label{lambda_config}

We report an ablation study on the choice of the parameters $(\lambda_{\mathrm{bd}}, \lambda_{\mathrm{res}})$ used in the multistage curriculum training. To keep the analysis general, the study was performed on two representative PDEs: the one-dimensional Burgers equation and the two-dimensional Navier Stokes equations with viscosity $\nu = 10^{-3}$.

\begin{figure*}[htbp]
    \centering
    \begin{subfigure}[t]{0.5\textwidth}
        \centering
        \includegraphics[width=\linewidth]{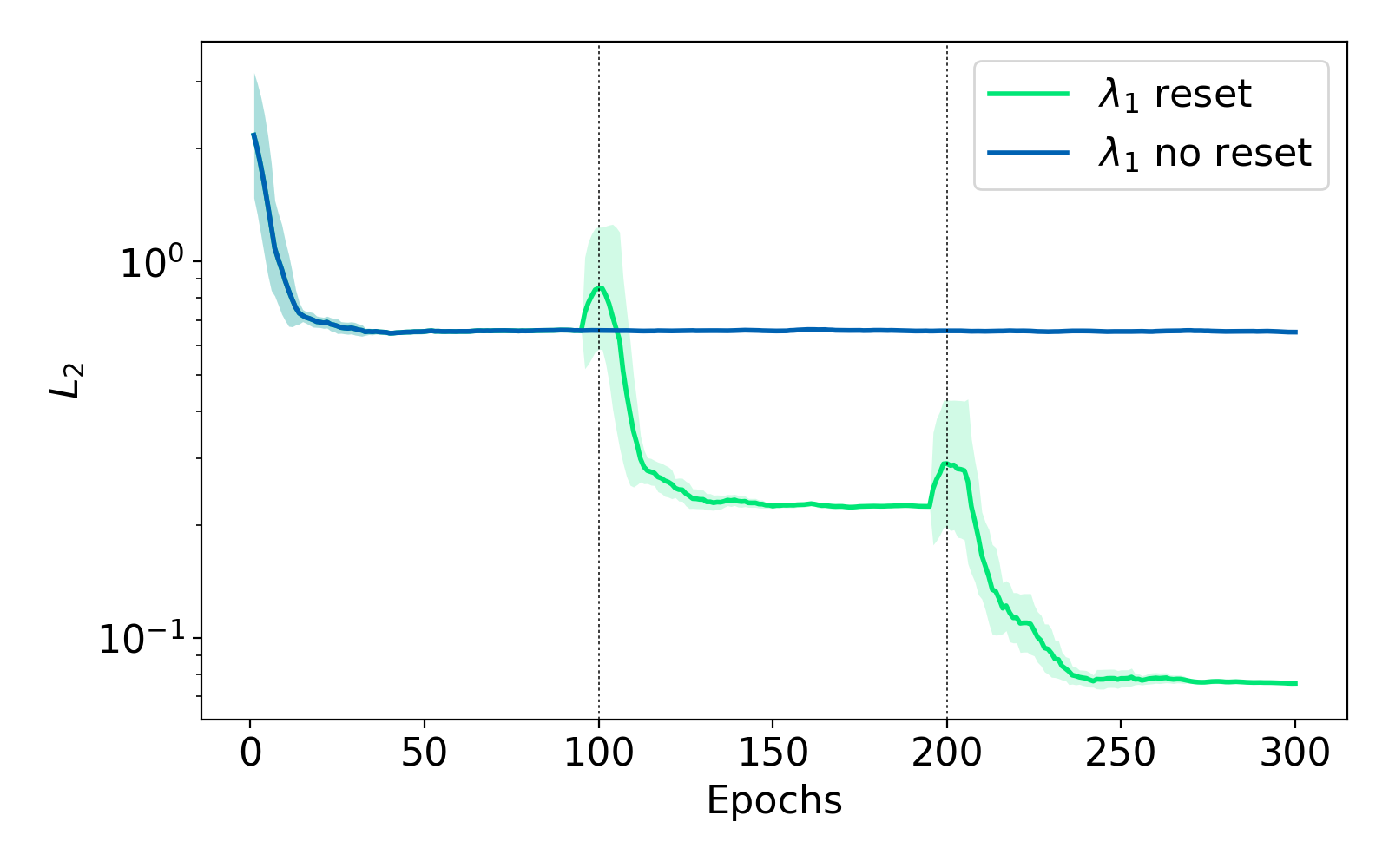}
    \end{subfigure}\hfill
    \begin{subfigure}[t]{0.5\textwidth}
        \centering
        \includegraphics[width=\linewidth]{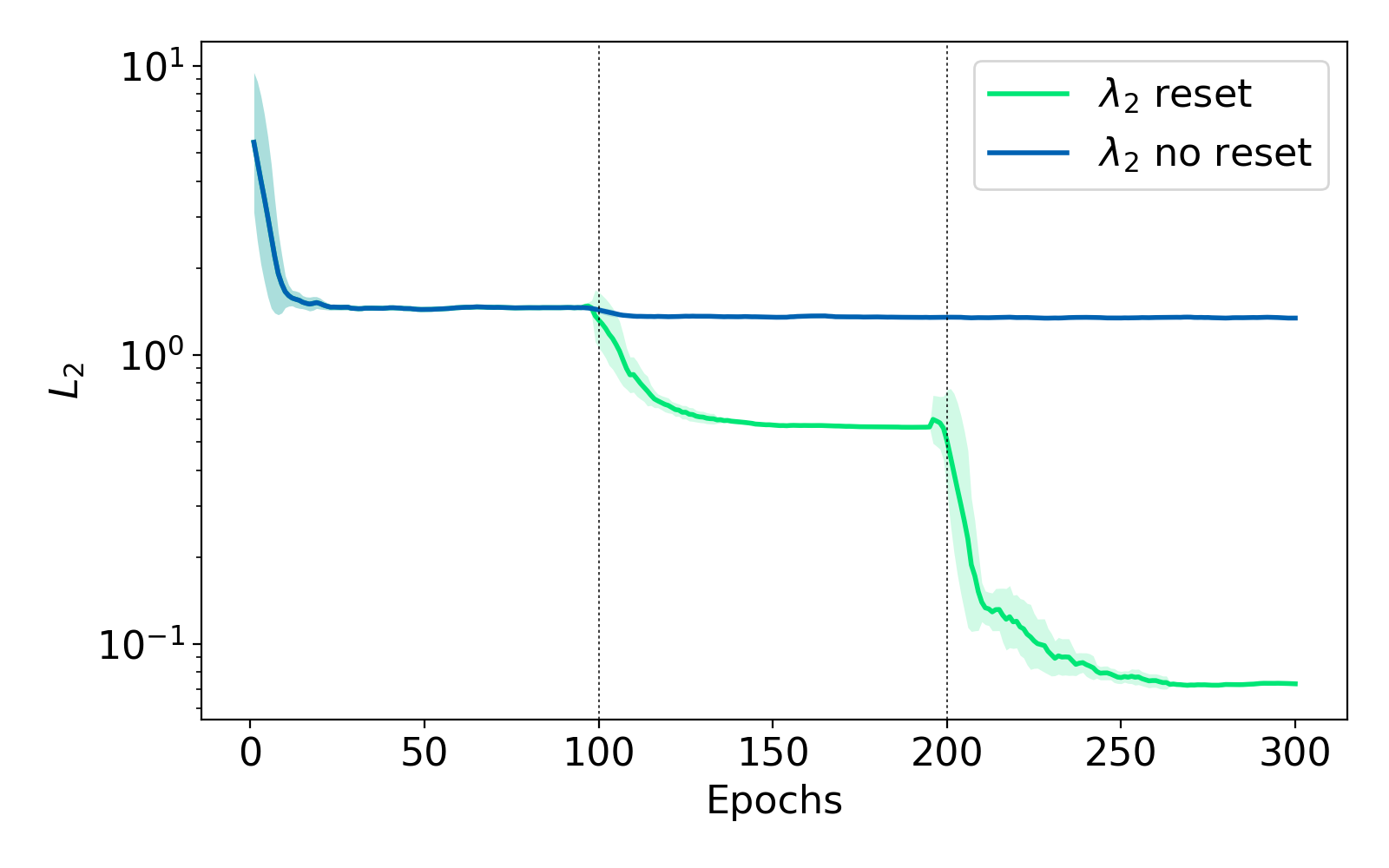}
    \end{subfigure}\hfill
    \begin{subfigure}[t]{0.5\textwidth}
        \centering
        \includegraphics[width=\linewidth]{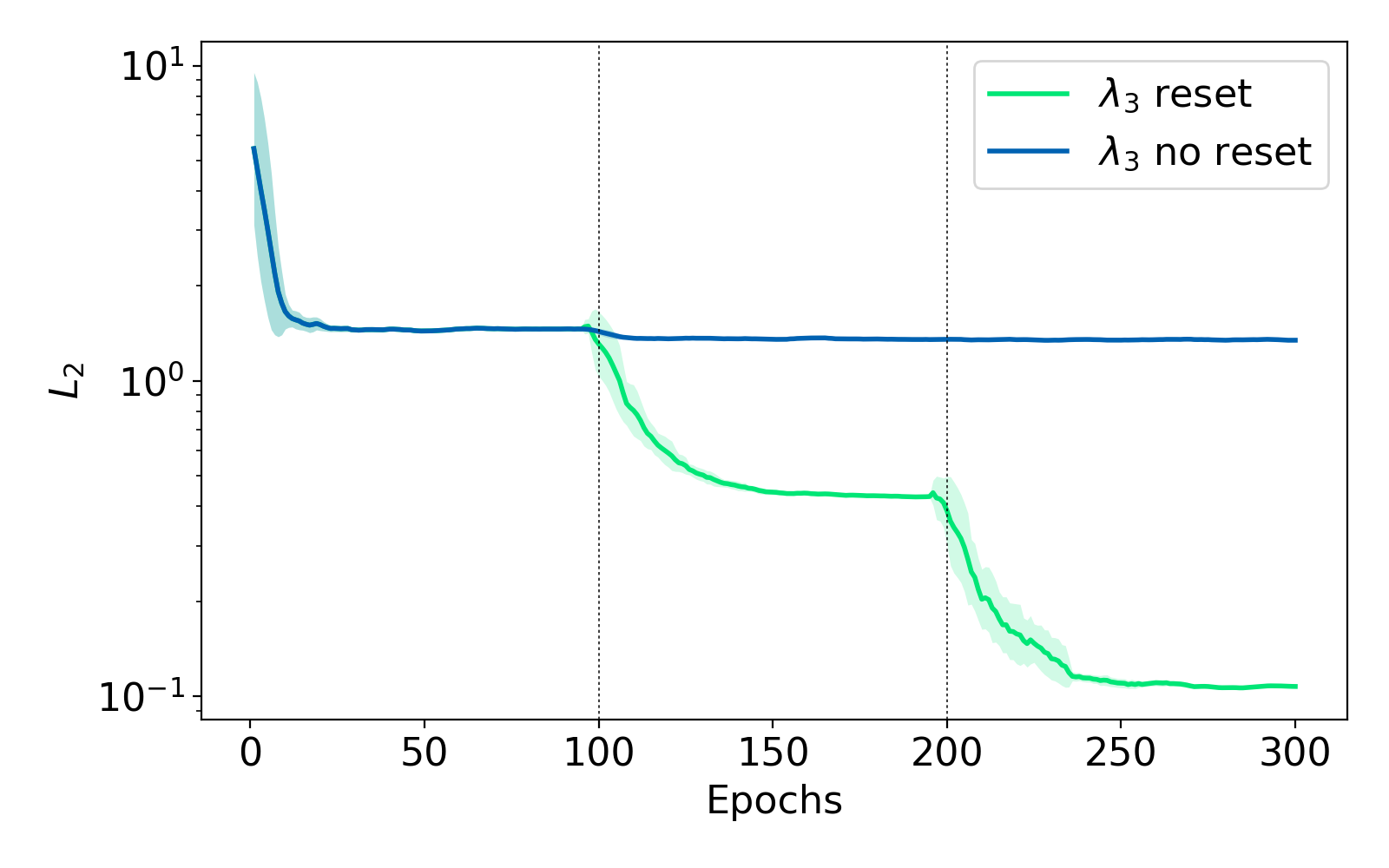}
    \end{subfigure}\hfill
    \begin{subfigure}[t]{0.5\textwidth}
        \centering
        \includegraphics[width=\linewidth]{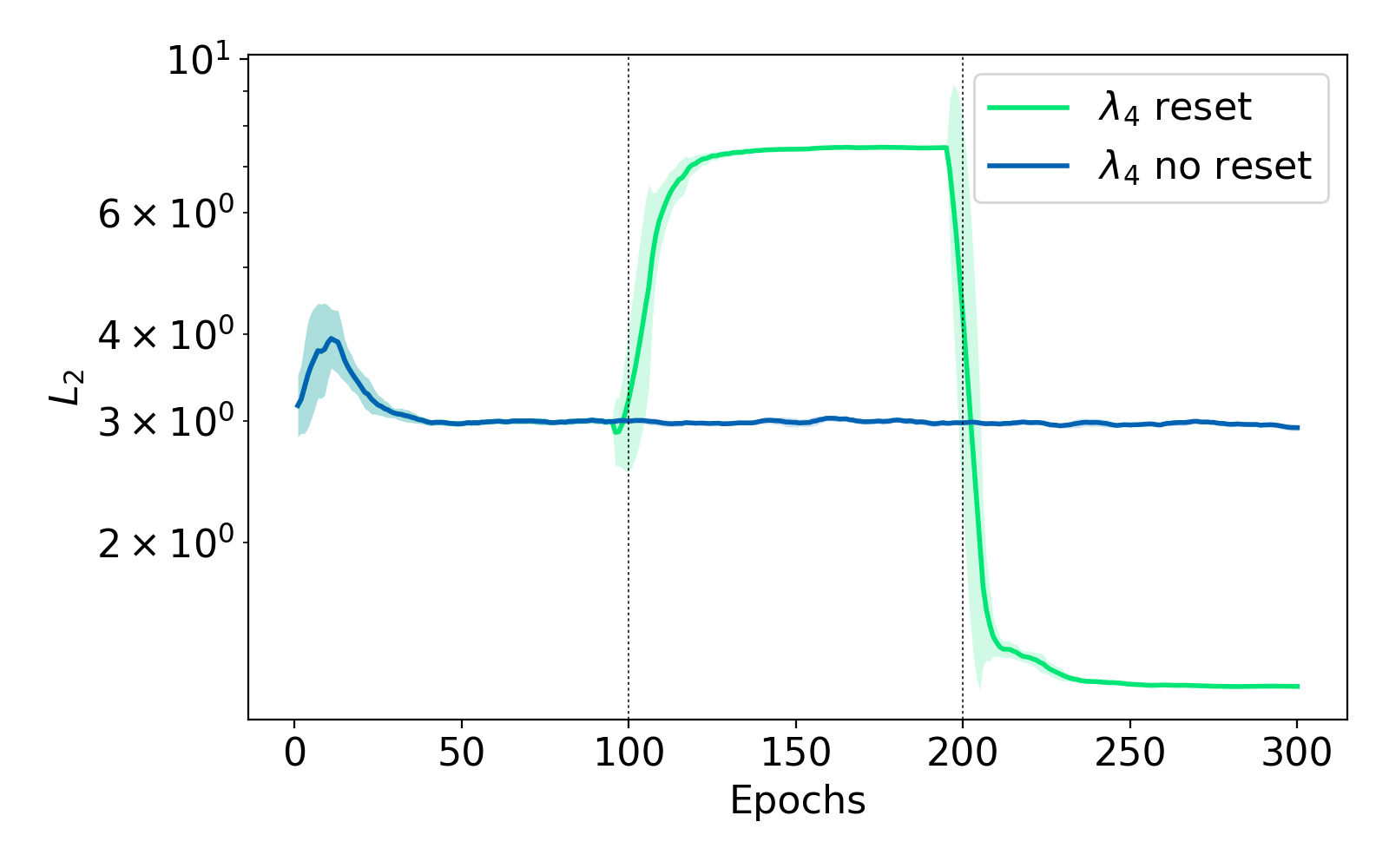}
    \end{subfigure}\hfill

    \caption{\textbf{Effect of optimizer reinitialization across different $\boldsymbol{\lambda}$ configurations for Burgers' equation.}
    Each panel compares the multistage training process with (green) and without (blue) Adam reinitialization.
    The four subfigures correspond to distinct choices of the loss weights $\lambda_{\mathrm{bd}}$, 
    $\lambda_{\mathrm{res}}$ used in the curriculum schedule (see Table~\ref{tab:ns_burger}).}

    \label{fig:lambda_ms}
\end{figure*}

\begin{figure*}[htbp]
    \centering
    \begin{subfigure}[t]{0.5\textwidth}
        \centering
        \includegraphics[width=\linewidth]{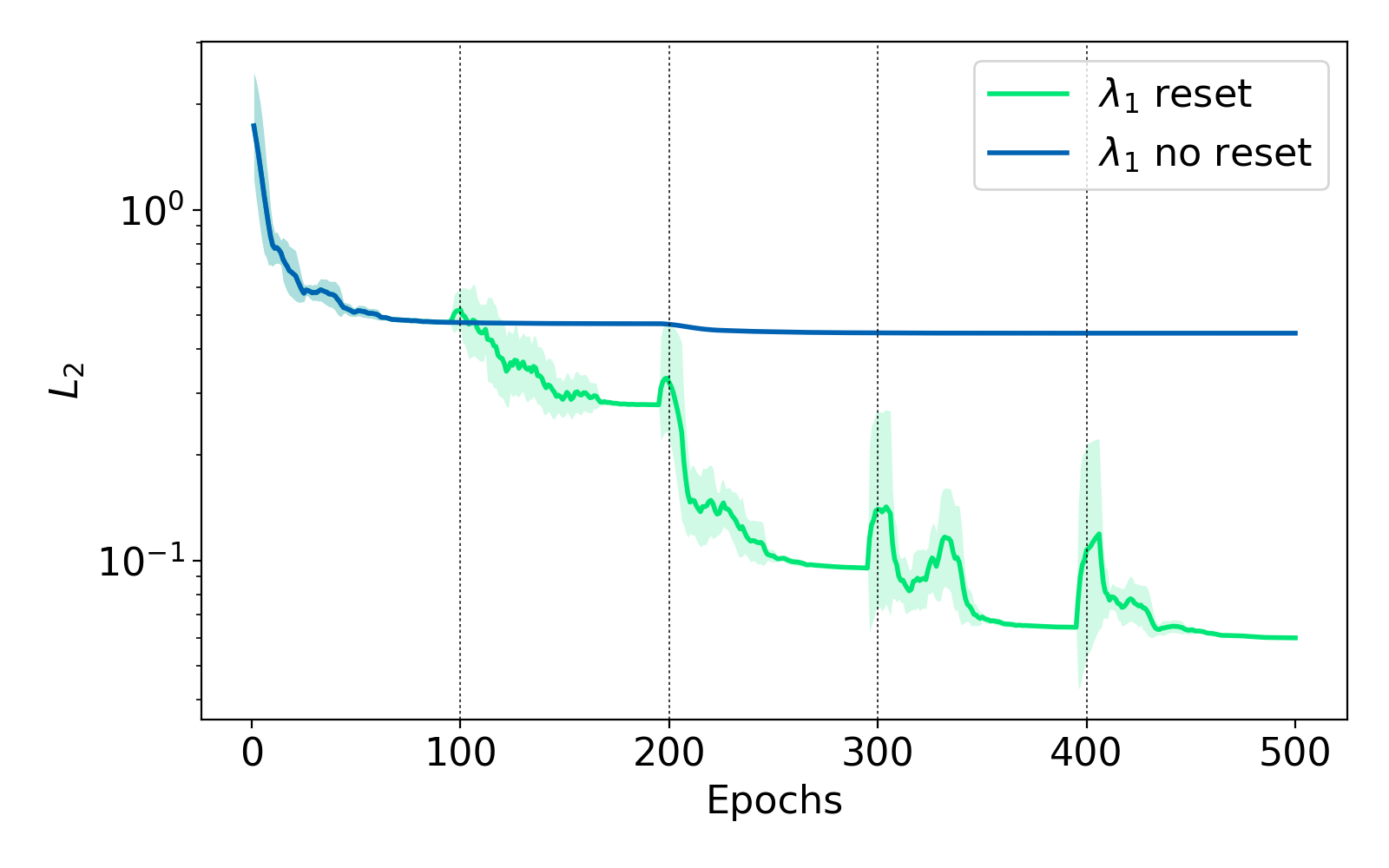}
    \end{subfigure}\hfill
    \begin{subfigure}[t]{0.5\textwidth}
        \centering
        \includegraphics[width=\linewidth]{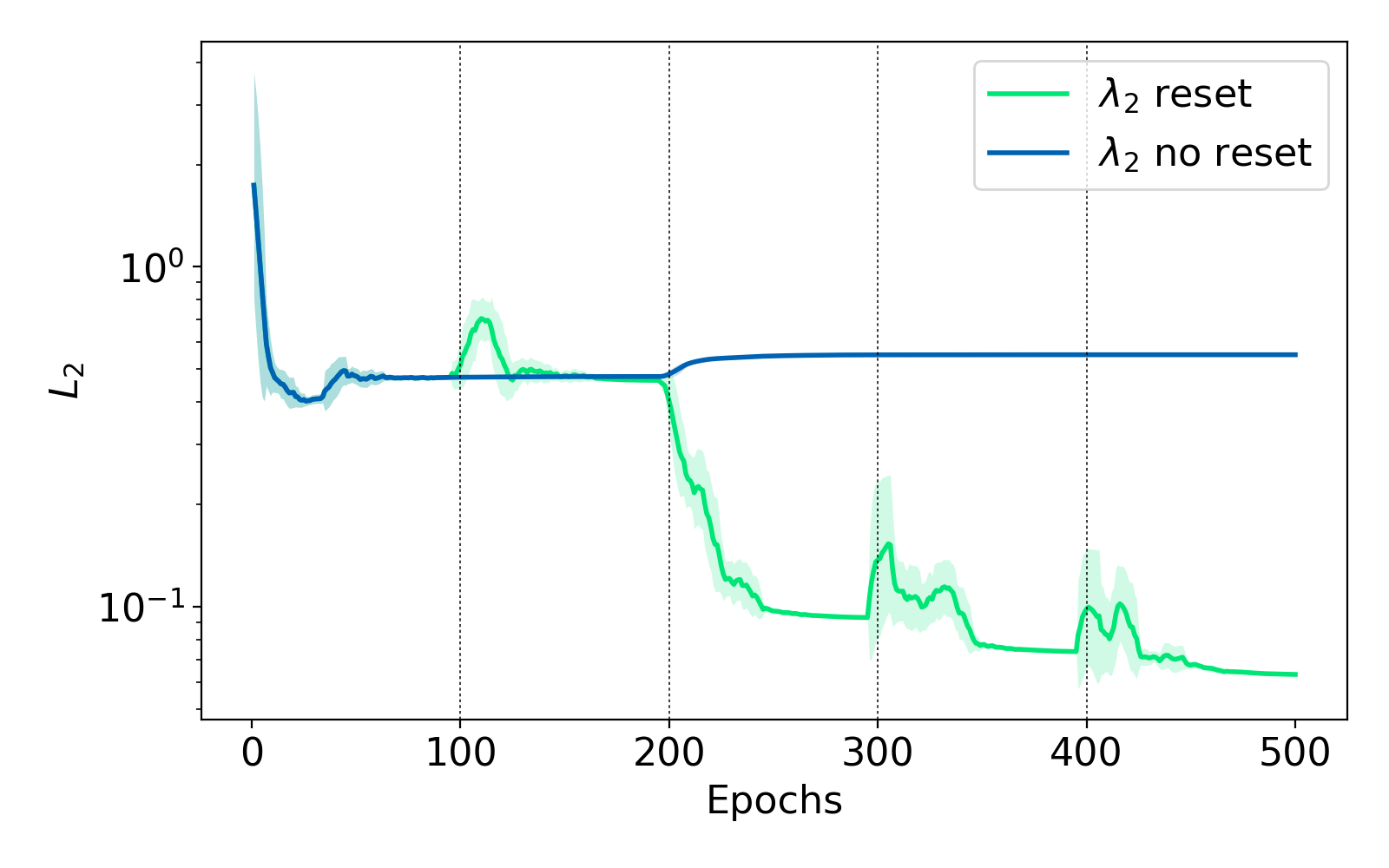}
    \end{subfigure}\hfill
    \begin{subfigure}[t]{0.5\textwidth}
        \centering
        \includegraphics[width=\linewidth]{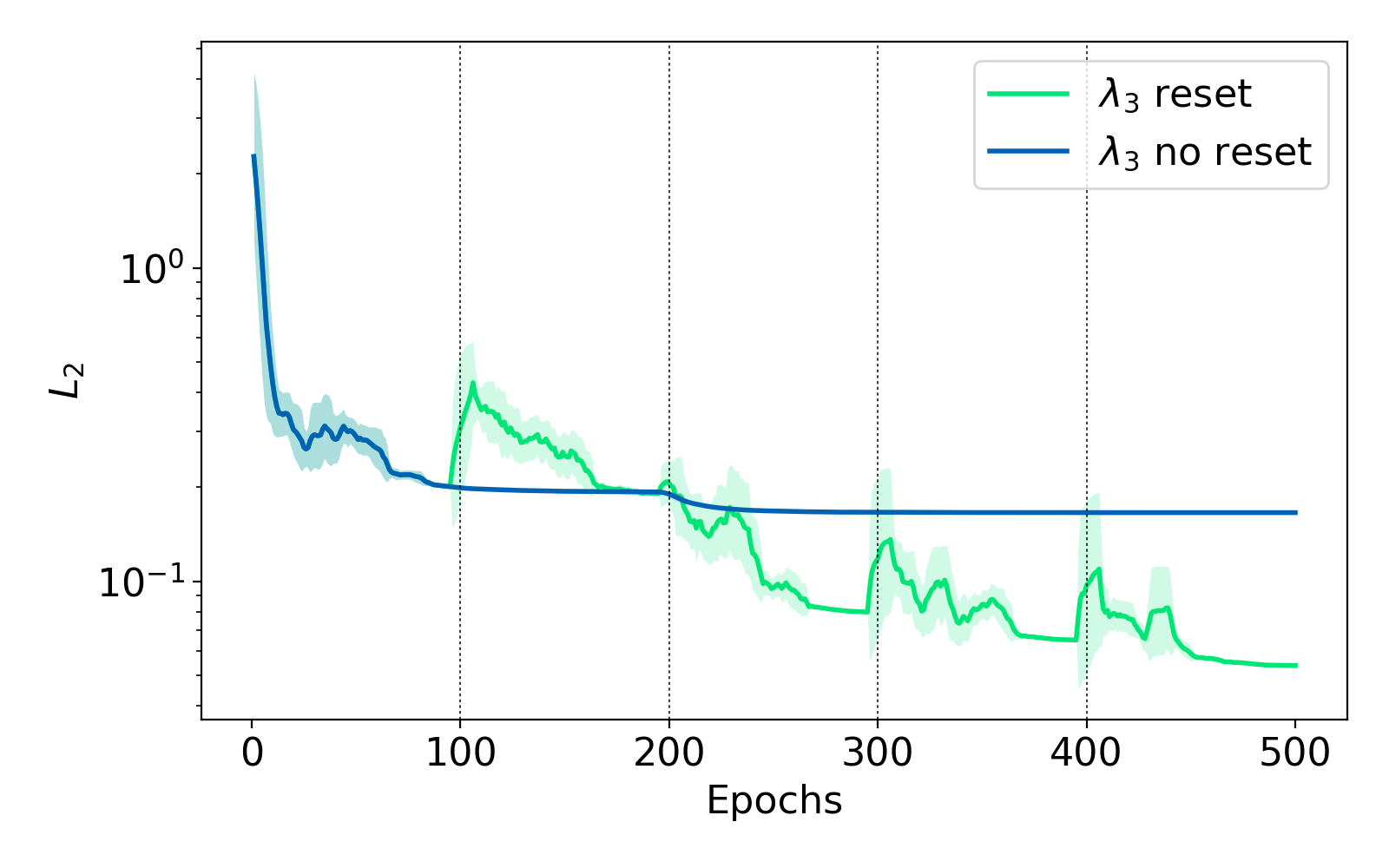}
    \end{subfigure}\hfill
    \begin{subfigure}[t]{0.5\textwidth}
        \centering
        \includegraphics[width=\linewidth]{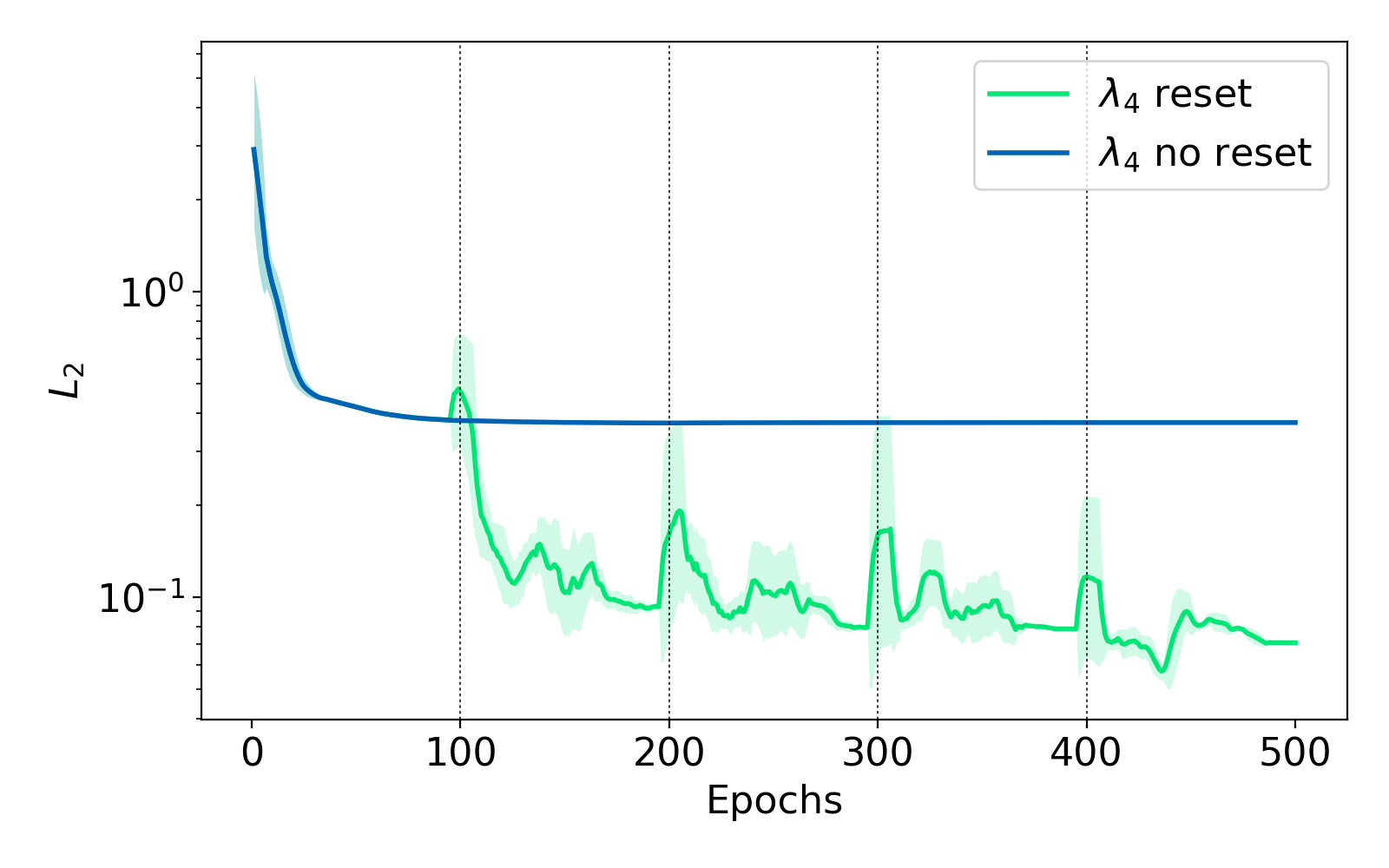}
    \end{subfigure}\hfill

    \caption{\textbf{Effect of optimizer reinitialization across different $\boldsymbol{\lambda}$ configurations for Navier Stokes equation.}
    Each panel compares the multistage training process with (green) and without (blue) Adam reinitialization.
    The four subfigures correspond to distinct choices of the loss weights $\lambda_{\mathrm{bd}}$, 
    $\lambda_{\mathrm{res}}$ used in the curriculum schedule (see Table~\ref{tab:ns_burger}).}

    \label{fig:lambda_ms_ns}
\end{figure*}

The baseline configuration was selected empirically as the one providing the most stable and efficient convergence. To evaluate the sensitivity of the method to these hyperparameters, we compared it with three alternative choices.
All values for Burgers and Navier Stokes are reported in Table~\ref{tab:ns_burger}, while the corresponding training curves appear in Figures~\ref{fig:lambda_ms} and~\ref{fig:lambda_ms_ns}. The results indicate that multistage training with optimizer reinitialization consistently produces more stable behavior and reliably converges across all tested configurations, whereas the version without reinitialization exhibits stagnation, overfitting and some irregularities as expected. In Burger case, we can see that excessively unbalanced weights can lead to unstable dynamics, where unsuitable residual scaling prevents the model from reaching convergence.

\begin{table*}[htbp]
\centering
\caption{\textbf{Unified multistage training schedules for Burgers and Navier--Stokes.}
The table reports the boundary and residual loss weights $(\lambda_{\mathrm{bd}},\lambda_{\mathrm{res}})$
for each schedule and phase. Burgers uses 3 phases; Navier--Stokes uses 5.}
\vspace{6pt}

\begin{tabular}{|l|c|cc|cc|}
\hline
\multirow{2}{*}{\textbf{Schedule}} 
& \multirow{2}{*}{\textbf{Phase}} 
& \multicolumn{2}{c|}{\textbf{Burgers}} 
& \multicolumn{2}{c|}{\textbf{Navier--Stokes}} \\

& 
& $\lambda_{\mathrm{bd}}$ & $\lambda_{\mathrm{res}}$ 
& $\lambda_{\mathrm{bd}}$ & $\lambda_{\mathrm{res}}$ \\
\hline

\multirow{5}{*}{Baseline: $\lambda_1$}
& 1 & 0.8  & 0.5  & 1.0  & 0.0  \\
& 2 & 0.5  & 1.0  & 1.0  & 0.0  \\
& 3 & 0.5  & 1.5  & 0.8  & 0.2  \\
& 4 & --  &  --     & 0.5  & 0.7  \\
& 5 &  --     &   --    & 0.2  & 1.0  \\
\hline

\multirow{5}{*}{Low-$\lambda$: $\lambda_2$}
& 1 & 1.0   & 0.0 & 0.5  & 0.0  \\
& 2 & 0.5  & 0.5  & 0.5  & 0.0  \\
& 3 & 0.25  & 0.75 & 0.4  & 0.1  \\
& 4 &  --      &  --     & 0.25 & 0.35 \\
& 5 &  --      &   --    & 0.1  & 0.5  \\
\hline

\multirow{5}{*}{High-$\lambda$: $\lambda_3$}
& 1 & 1.0  & 0.0 & 1.5  & 0.0  \\
& 2 & 1.0 & 0.5  & 1.5  & 0.0  \\
& 3 & 1.0 & 1.0 & 1.2  & 0.3  \\
& 4 &  --     &  --     & 0.75 & 1.05 \\
& 5 & --      &   --    & 0.3  & 1.5  \\
\hline

\multirow{5}{*}{Shape-changed: $\lambda_4$}
& 1 & 1.0  & 0.2  & 1.0   & 0.6   \\
& 2 & 0.7  & 0.8  & 1.0  & 0.7   \\
& 3 & 0.4  & 1.2  & 0.8   & 0.9   \\
& 4 &  --     &  --     & 0.5  & 1.0  \\
& 5 &   --    &  --     & 0.2  & 1.2  \\
\hline

\end{tabular}
\label{tab:ns_burger}
\end{table*}

\subsection{Dataset Generation} \label{sup:dataset}
All datasets employed in this study were synthetically generated using pseudo-spectral solvers,
following the numerical setup introduced in~\citep{Li2020FourierNO}.
This approach allows full control over viscosity, boundary conditions, and numerical accuracy.
Each dataset provides high-resolution spatiotemporal fields used for both supervised
and physics-informed training.

\paragraph{Burgers’ equation.}
The one-dimensional viscous Burgers dataset was generated by solving
\[
\partial_tu + \tfrac{1}{2}\partial_x(u^2) = \nu\,\partial_{xx}u,
\quad x \in [0,1), \; t \in (0,T],
\]
under periodic boundary conditions.
The initial velocity field $u(x,0)$ was sampled from a zero-mean Gaussian Random Field (GRF)
with spectral covariance
$C(k) \propto ((2\pi k)^2 + \tau^2)^{-\gamma/2}$,
implemented through real-valued Fourier coefficients with 2/3 dealiasing.
Temporal integration employed the pseudo-spectral ETDRK4 scheme (Cox–Matthews formulation)
with viscosity $\nu = 1\times10^{-1}$, time step $\Delta t = 10^{-4}$,
and final time $T = 0.1$.
Each trajectory contains $N_x = 8192$ spatial points and $N_t = 200$ time snapshots.
The dataset consists of $N_s = 100$ independent GRF realizations, stored in MATLAB format
with fields $a$(initial condition), $u$ (temporal evolution), $x$, and \texttt{t}.

\paragraph{Navier--Stokes equations (vorticity formulation).}
Two-dimensional incompressible Navier–Stokes dynamics were simulated in the vorticity formulation,
\[
\partial_t\omega + \mathbf{u}\cdot\nabla\omega = \nu\,\Delta\omega + f(x,y),
\quad \nabla\cdot\mathbf{u}=0,
\]
on the periodic domain $[0,1]^2$.
The initial vorticity $\omega(x,y,0)$ was sampled from a Gaussian Random Field
with spectral parameters $(\alpha=2.5,\,\tau=7)$.
Two viscosity regimes were simulated, $\nu = 10^{-3}$ and $\nu = 10^{-4}$,
to analyze the impact of increasing Reynolds number.
The external forcing term was set to
$f(x,y) = -4\pi^2(\sin(4\pi x) + \sin(4\pi y))$,
driving a vortex-dominated flow.
Temporal integration used a semi-implicit Crank–Nicolson scheme in Fourier space
with 2/3 dealiasing and time step $\Delta t = 10^{-4}$,
evolved up to $T = 15$ with $N_t = 60$ saved frames.
Each snapshot includes vorticity $\omega$, velocity components $(u_x,u_y)$,
and time instants, on a $256\times256$ grid.

\paragraph{Kolmogorov flow}
A second Navier–Stokes dataset was generated with a Kolmogorov-type forcing
$f(x,y) = -2\cos(4\pi y)$,
at viscosity $\nu = 2\times10^{-3}$.
All other numerical parameters matched those of the vorticity dataset.
This configuration produces a laminar-to-chaotic transition under periodic boundaries,
serving as an additional benchmark for assessing generalization under distinct forcing regimes.

\subsection{Burgers' Equation}\label{sup:burger}

\begin{figure*}[htbp]
    \centering
    \begin{subfigure}[t]{0.32\textwidth}
        \includegraphics[width=\linewidth]{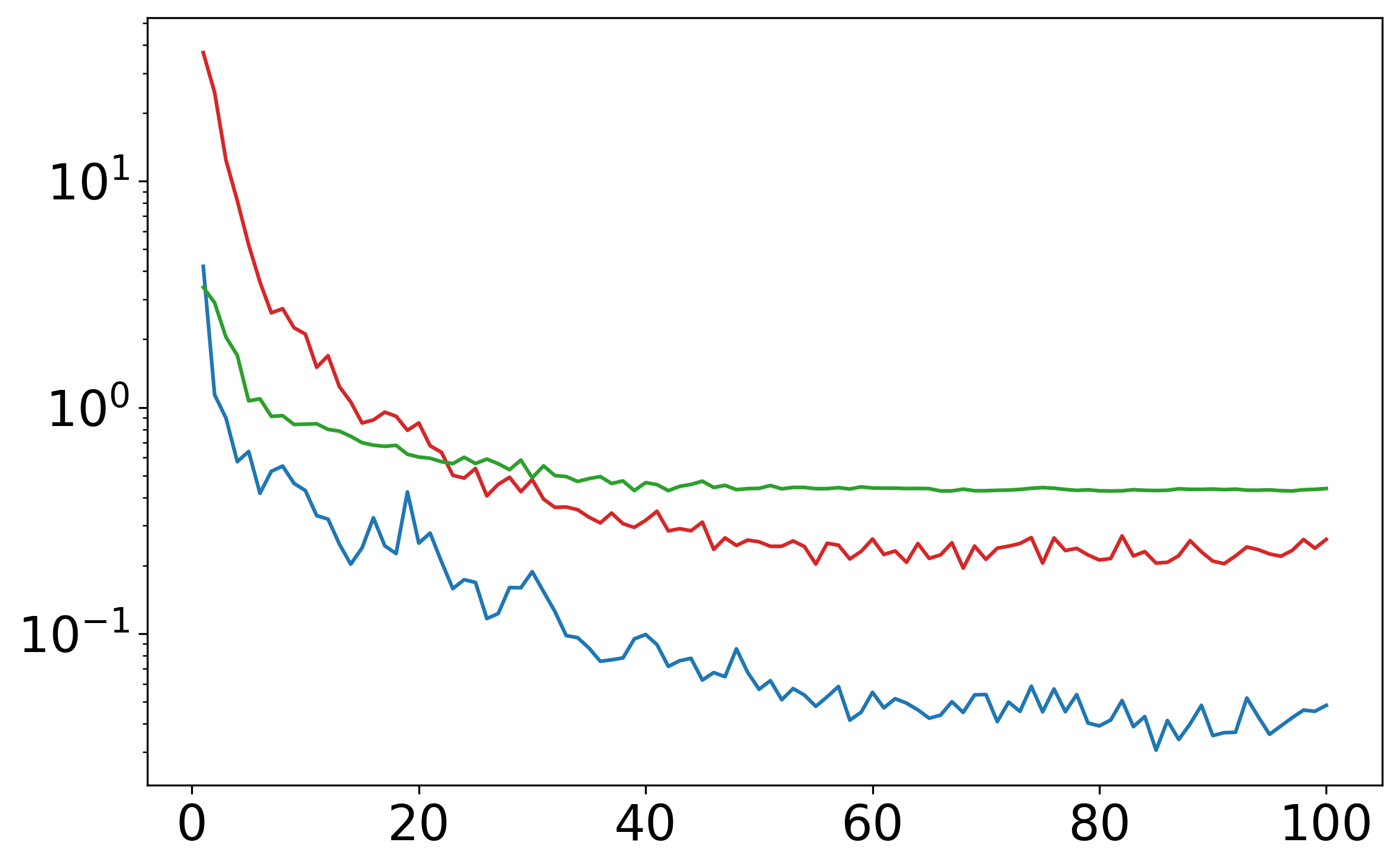}
        \captionsetup{justification=raggedright,singlelinecheck=false}
        \caption{Stage~1 final losses: $\mathcal{L}_{\mathrm{bd}}{=}4.82{\times}10^{-2}$, $\mathcal{L}_{\mathrm{res}}{=}2.61{\times}10^{-1}$, $\mathcal{L}_{\mathrm{test}}{=}4.38{\times}10^{-1}$.}
    \end{subfigure}
    \hfill
    \begin{subfigure}[t]{0.32\textwidth}
        \centering
        \includegraphics[width=\linewidth]{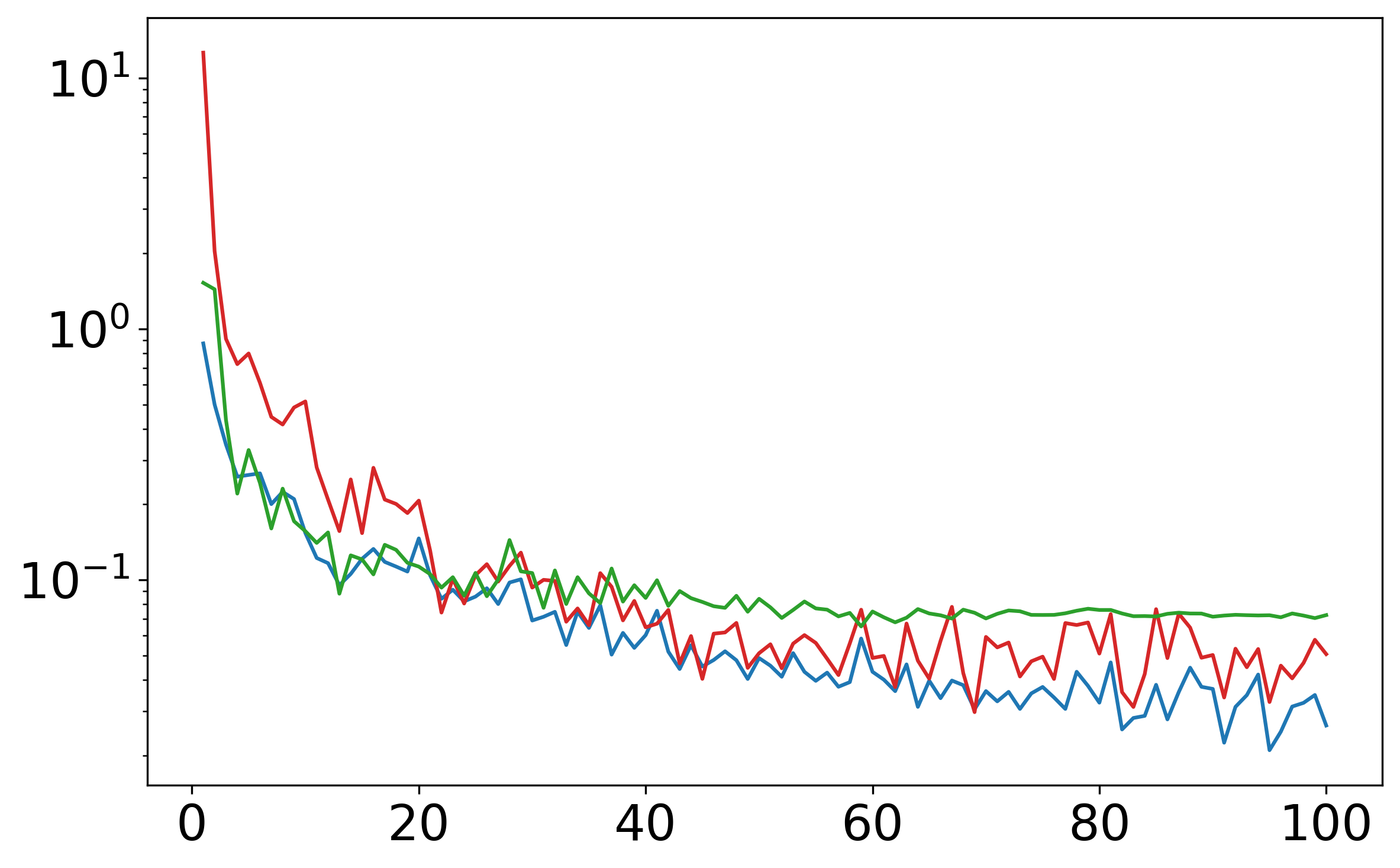}
        \captionsetup{justification=raggedright,singlelinecheck=false}
        \caption{Stage~2 final losses: $\mathcal{L}_{\mathrm{bd}}{=}2.63{\times}10^{-2}$, $\mathcal{L}_{\mathrm{res}}{=}5.06{\times}10^{-2}$, $\mathcal{L}_{\mathrm{test}}{=}7.24{\times}10^{-2}$.}
    \end{subfigure}
    \hfill
    \begin{subfigure}[t]{0.32\textwidth}
        \centering
        \includegraphics[width=\linewidth]{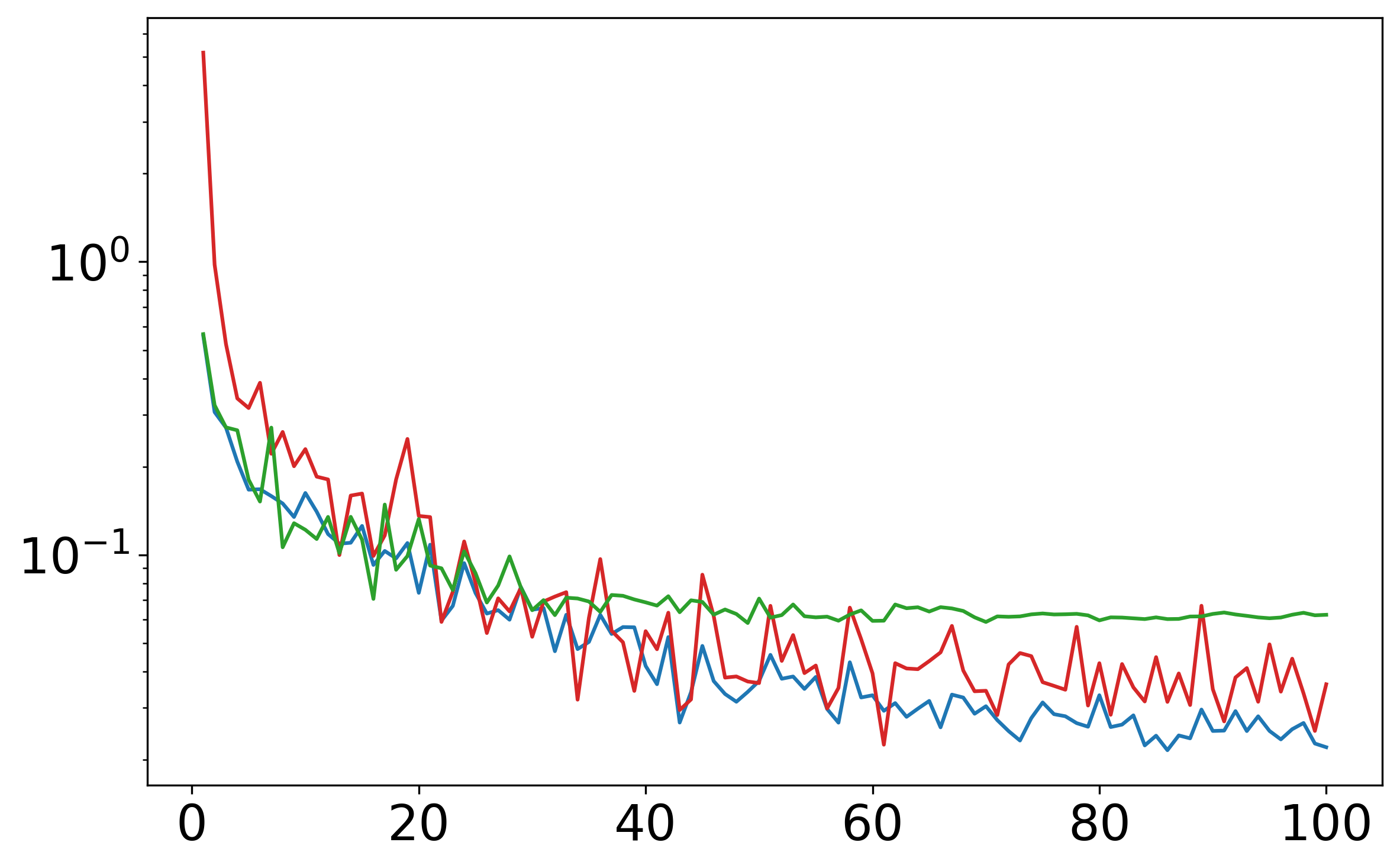}
        \captionsetup{justification=raggedright,singlelinecheck=false}
        \caption{Stage~3 final losses: $\mathcal{L}_{\mathrm{bd}}{=}2.20{\times}10^{-2}$, $\mathcal{L}_{\mathrm{res}}{=}3.61{\times}10^{-2}$, $\mathcal{L}_{\mathrm{test}}{=}6.24{\times}10^{-2}$.}
    \end{subfigure}

    \caption{\textbf{Stage-wise loss evolution during multi-stage training of the 1D Burgers equation.} 
    Red: residual loss; blue: boundary loss; green: validation loss over the full domain. 
    Each transition corresponds to a curriculum stage with new weighting $(\lambda_{\mathrm{bd}}, \lambda_{\mathrm{res}})$, 
    leading to progressive convergence and balanced enforcement of boundary and physics constraints.}
    \label{burgloss}
\end{figure*}

Figure~\ref{burgloss} illustrates the evolution of the individual loss components during the multi-stage training of the one-dimensional Burgers equation. 
Three successive stages were employed, with progressively increasing residual weighting $(\lambda_{\mathrm{res}})$ and decreasing boundary weighting $(\lambda_{\mathrm{bd}})$, as indicated in the figure. 
At the end of each phase, all losses consistently decrease, confirming the effectiveness of the staged optimization in enforcing both boundary and residual constraints.

\begin{figure*}[htbp]
    \centering
    \includegraphics[width=\linewidth]{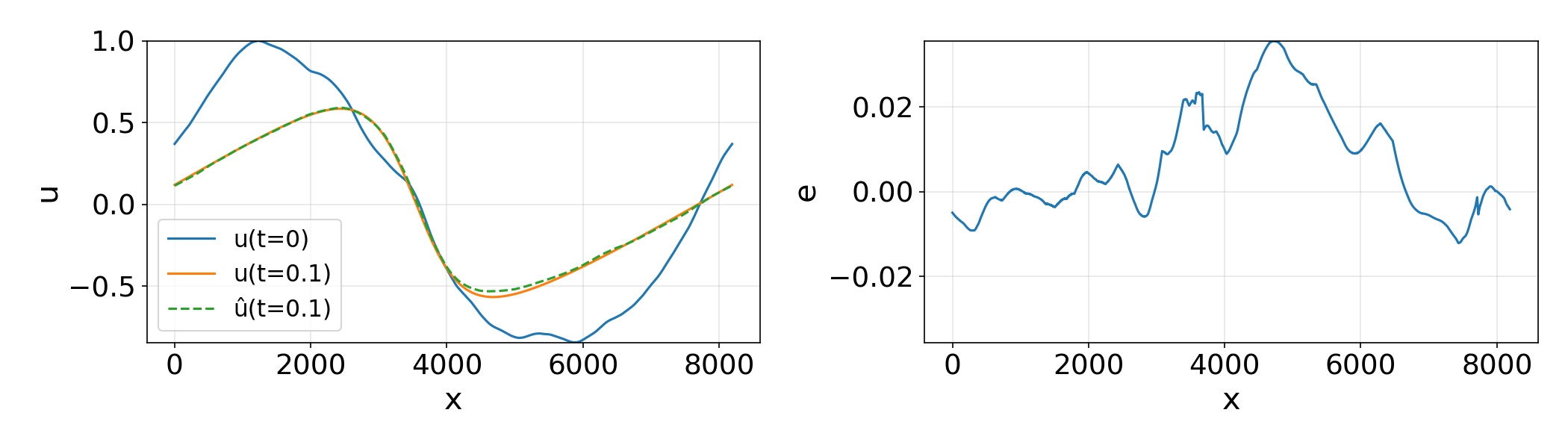}
    \caption{\textbf{Prediction on the Burgers' equation.} 
    \textbf{Left:} initial condition $u(t_0)$ (blue), reference solution at $t{=}0.1$ (orange), and PhIS--FNO prediction (green dashed). 
    \textbf{Right:} pointwise error between predicted and reference solutions, with $\|e\|_2 \approx 1.41{\times}10^{-2}$. 
    Only 5\% of boundary points were supervised, while interior dynamics were learned through the physics-informed residual loss.}
    \label{Burger}
\end{figure*}

\medskip
\noindent
Figure~\ref{Burger} reports the predicted solution obtained after the third stage. 
Training was performed with boundary supervision on only $5\%$ of the domain points (left and right edges), while the interior dynamics were learned exclusively from the PDE residual. 
The PhIS--FNO accurately recovers the reference solution, achieving an $L_2$ error of approximately $1.4{\times}10^{-2}$ and smooth physical consistency across the spatial domain.

\subsection{Navier Stokes 2D: Multi-stage training}\label{sup:navierstokes}
\begin{figure*}[htbp]
    \centering
    \textbf{(a) Navier--Stokes 2D, $\nu = 10^{-3}$}\par\vspace{4pt}
    \begin{subfigure}[t]{0.30\textwidth}
        \centering
        \includegraphics[width=\linewidth]{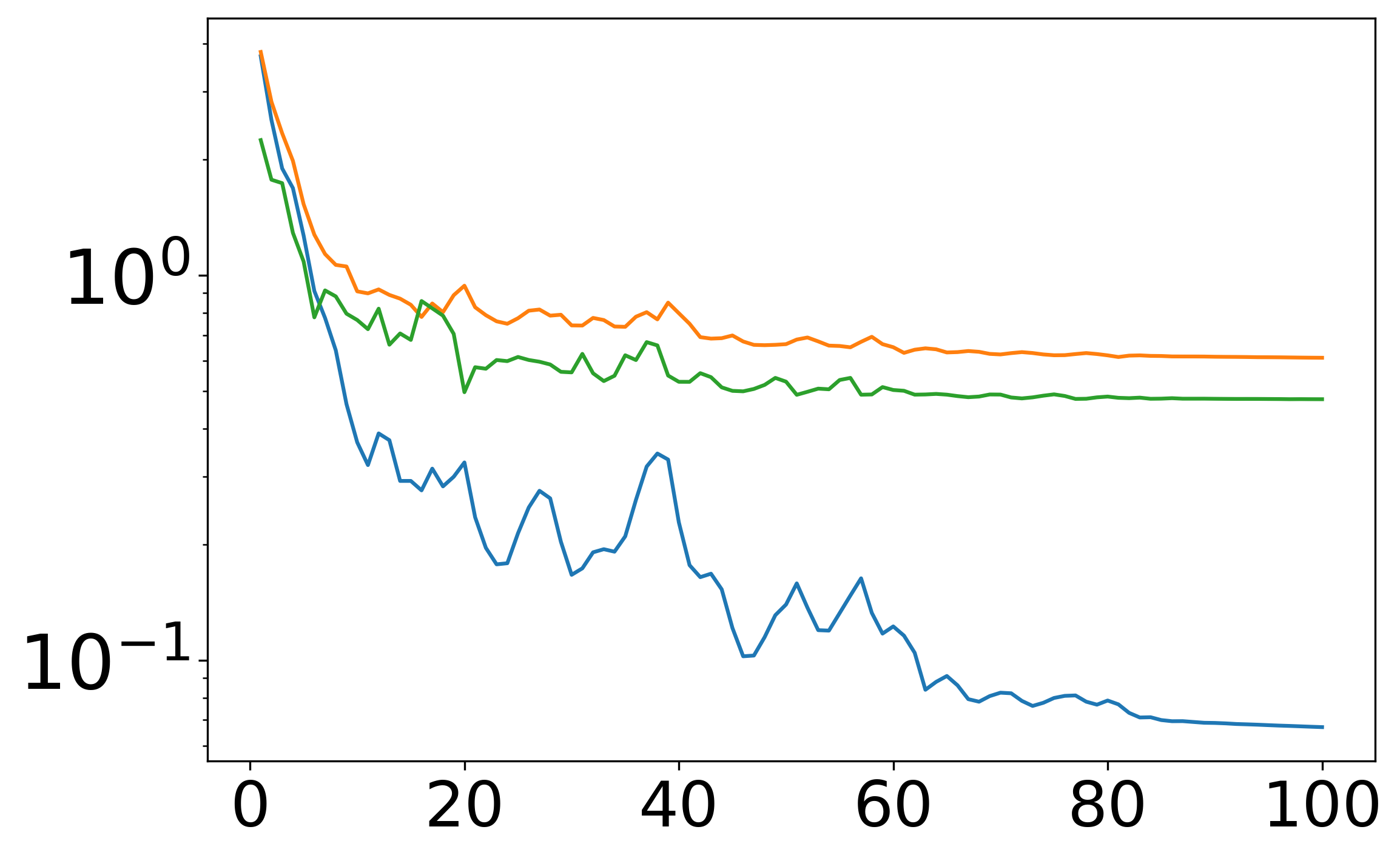}
        \caption{Stage 1}
    \end{subfigure}\hfill
    \begin{subfigure}[t]{0.30\textwidth}
        \centering
        \includegraphics[width=\linewidth]{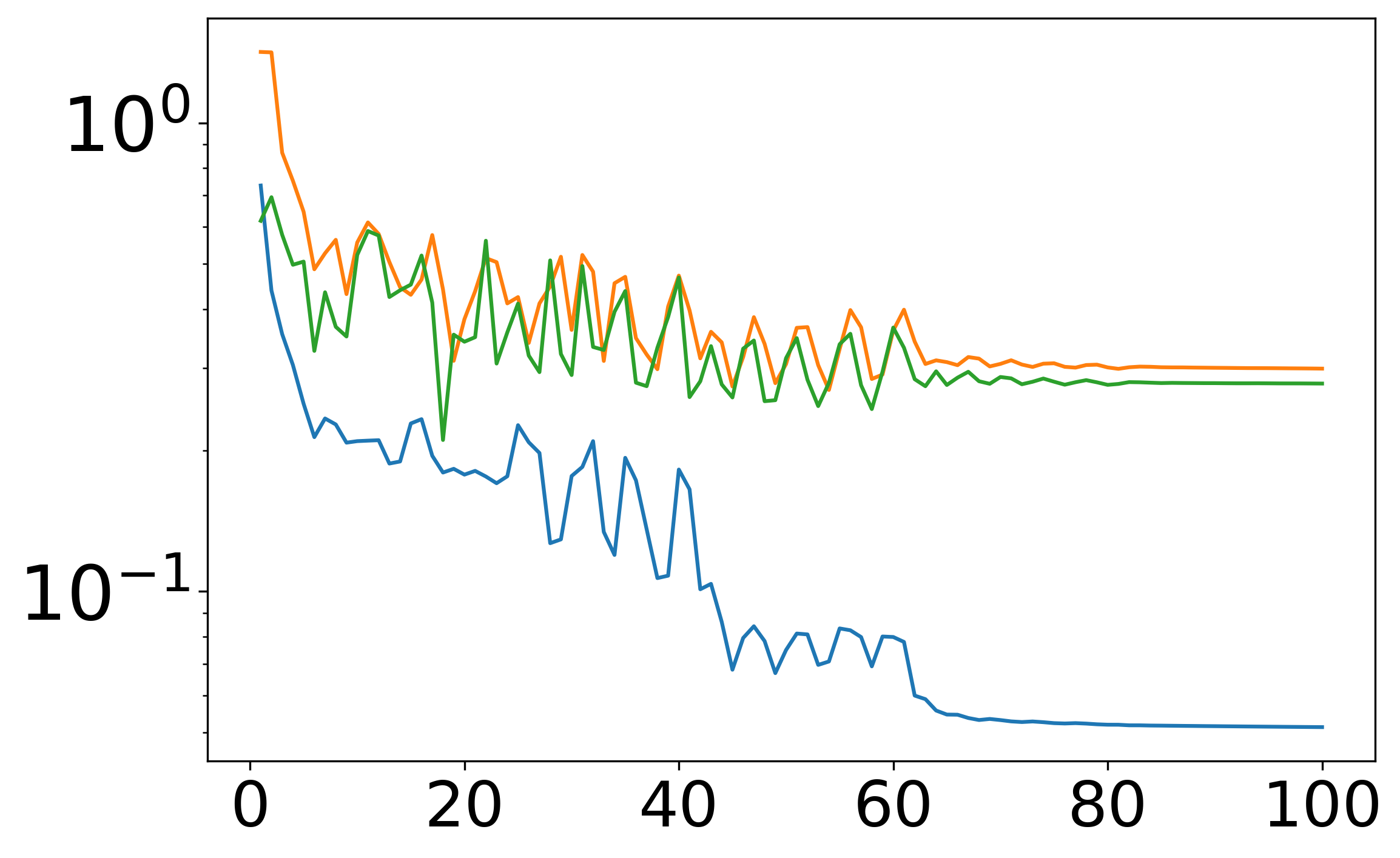}
        \caption{Stage 2}
    \end{subfigure}\hfill
    \begin{subfigure}[t]{0.30\textwidth}
        \centering
        \includegraphics[width=\linewidth]{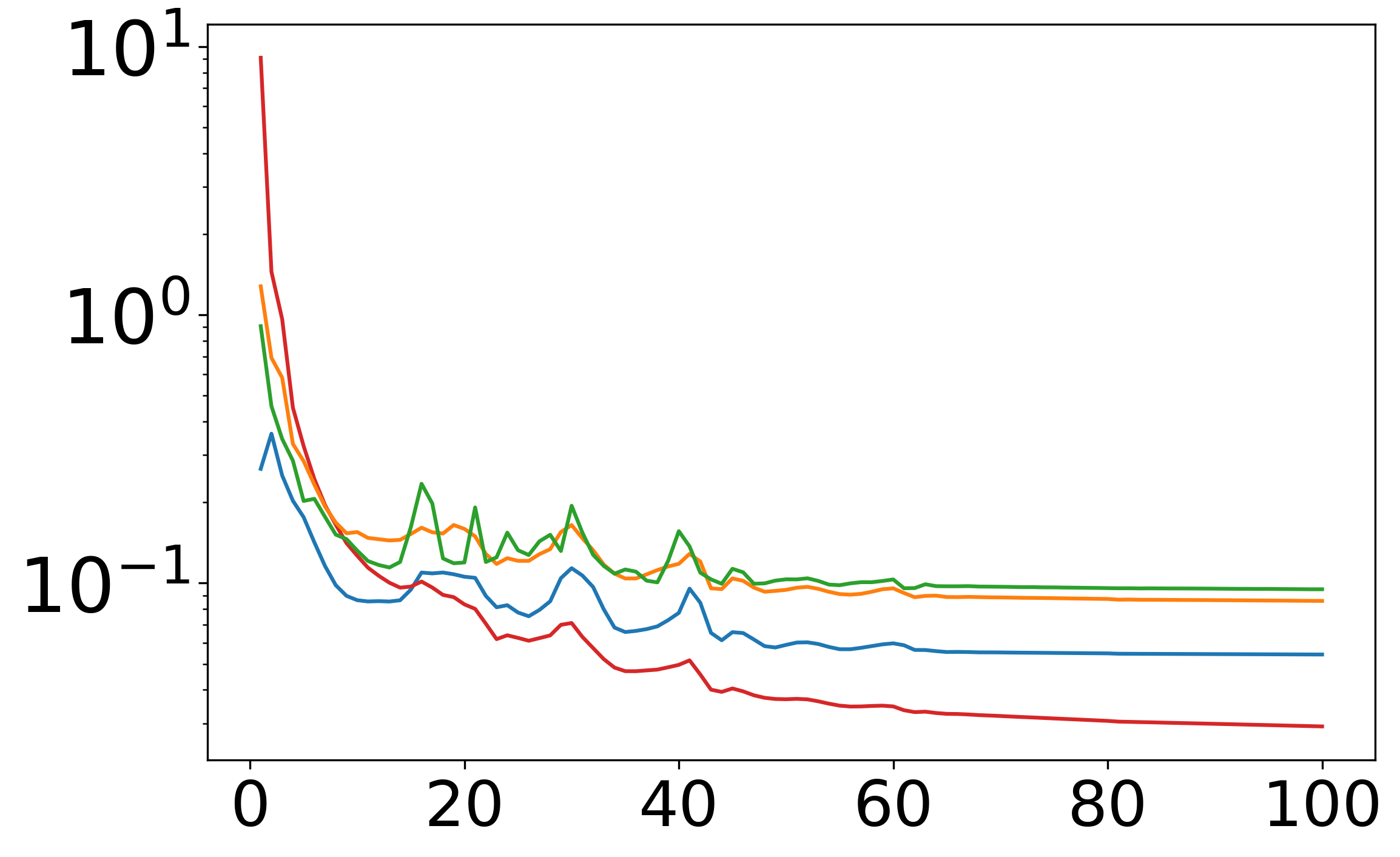}
        \caption{Stage 3}
    \end{subfigure}

    \vspace{4pt}

    \begin{subfigure}[t]{0.30\textwidth}
        \centering
        \includegraphics[width=\linewidth]{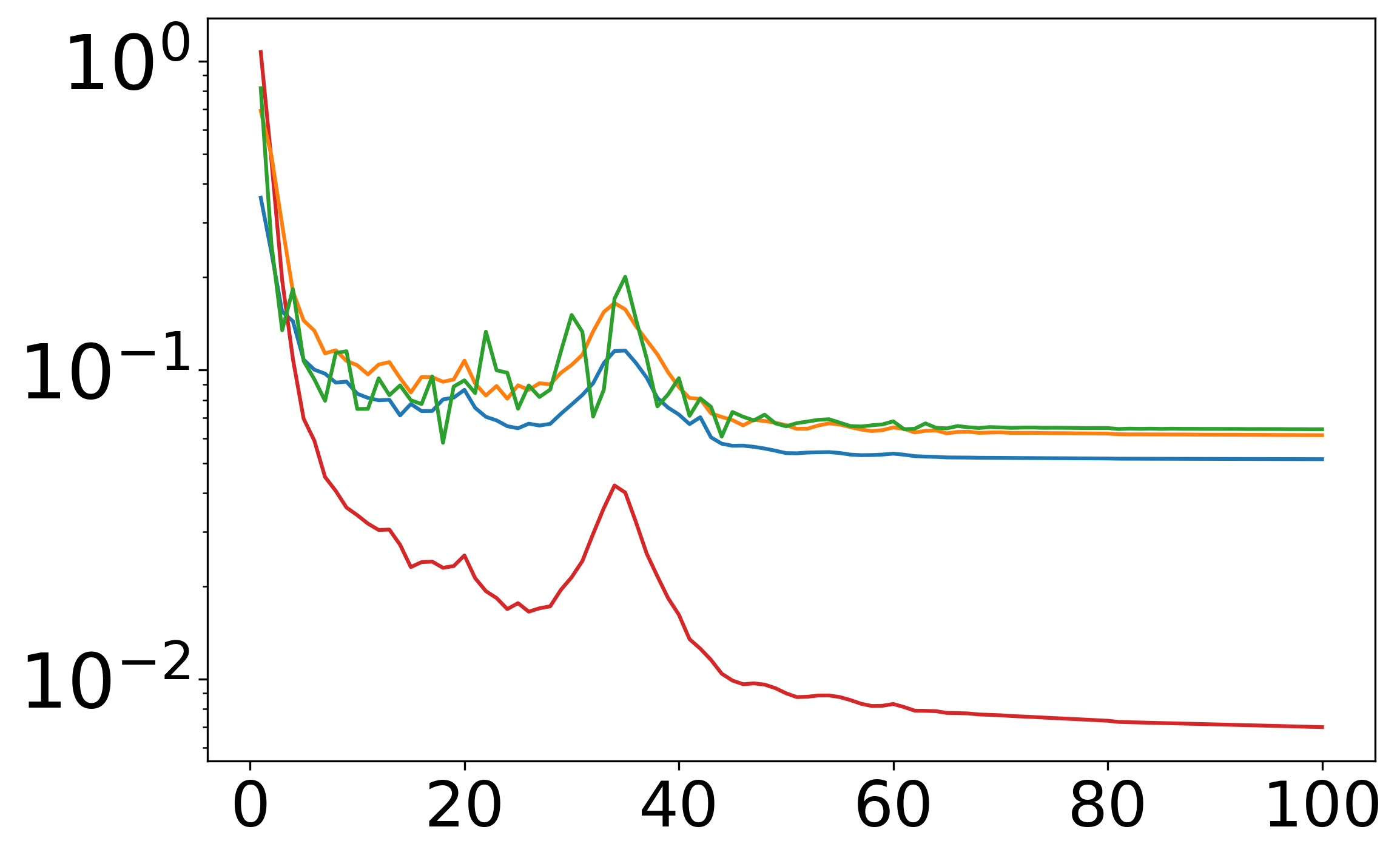}
        \caption{Stage 4}
    \end{subfigure}
    \begin{subfigure}[t]{0.30\textwidth}
        \centering
        \includegraphics[width=\linewidth]{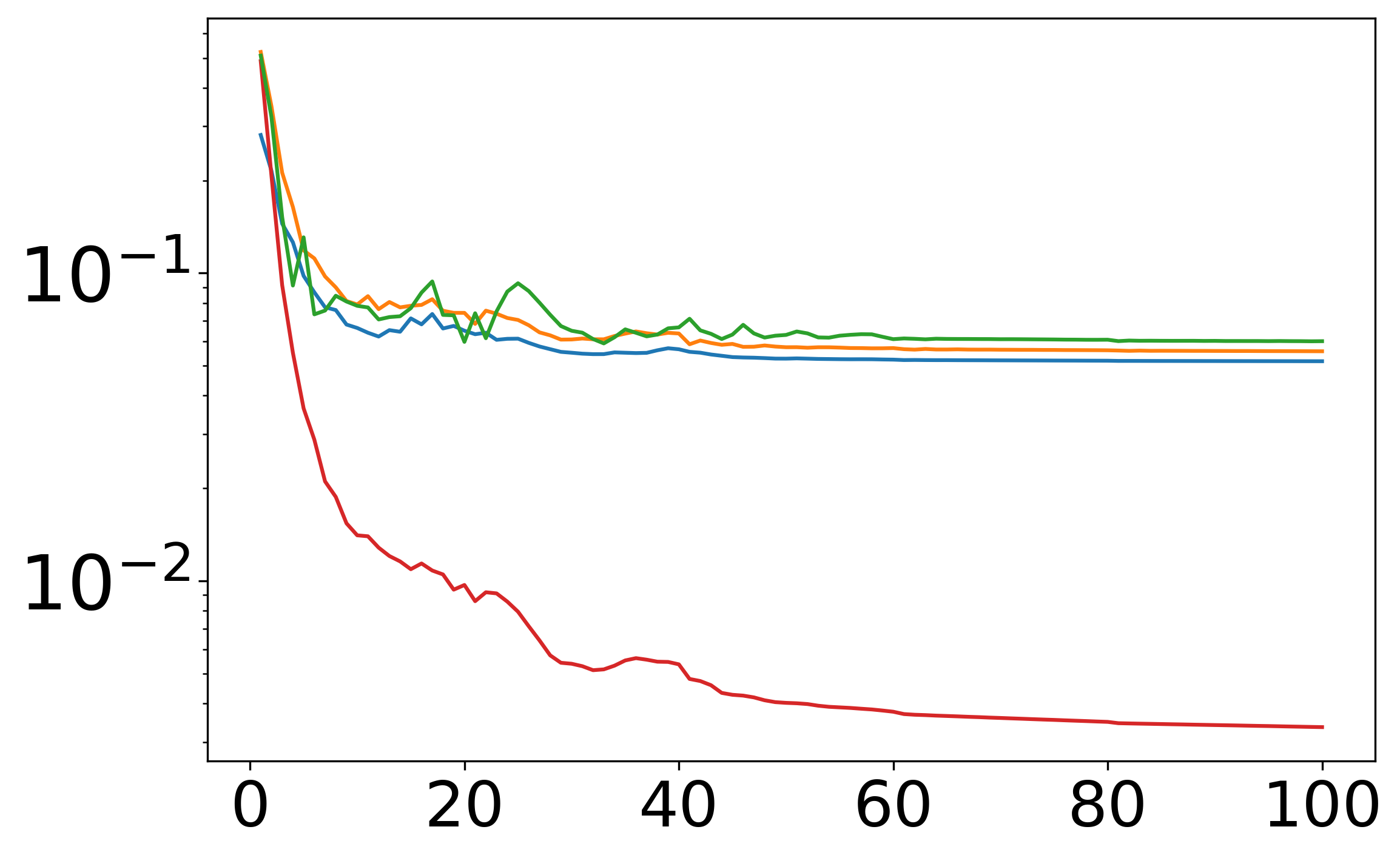}
        \caption{Stage 5}
    \end{subfigure}

    \vspace{12pt}

    \textbf{(b) Navier--Stokes 2D, $\nu = 10^{-4}$}\par\vspace{4pt}
    \begin{subfigure}[t]{0.30\textwidth}
        \centering
        \includegraphics[width=\linewidth]{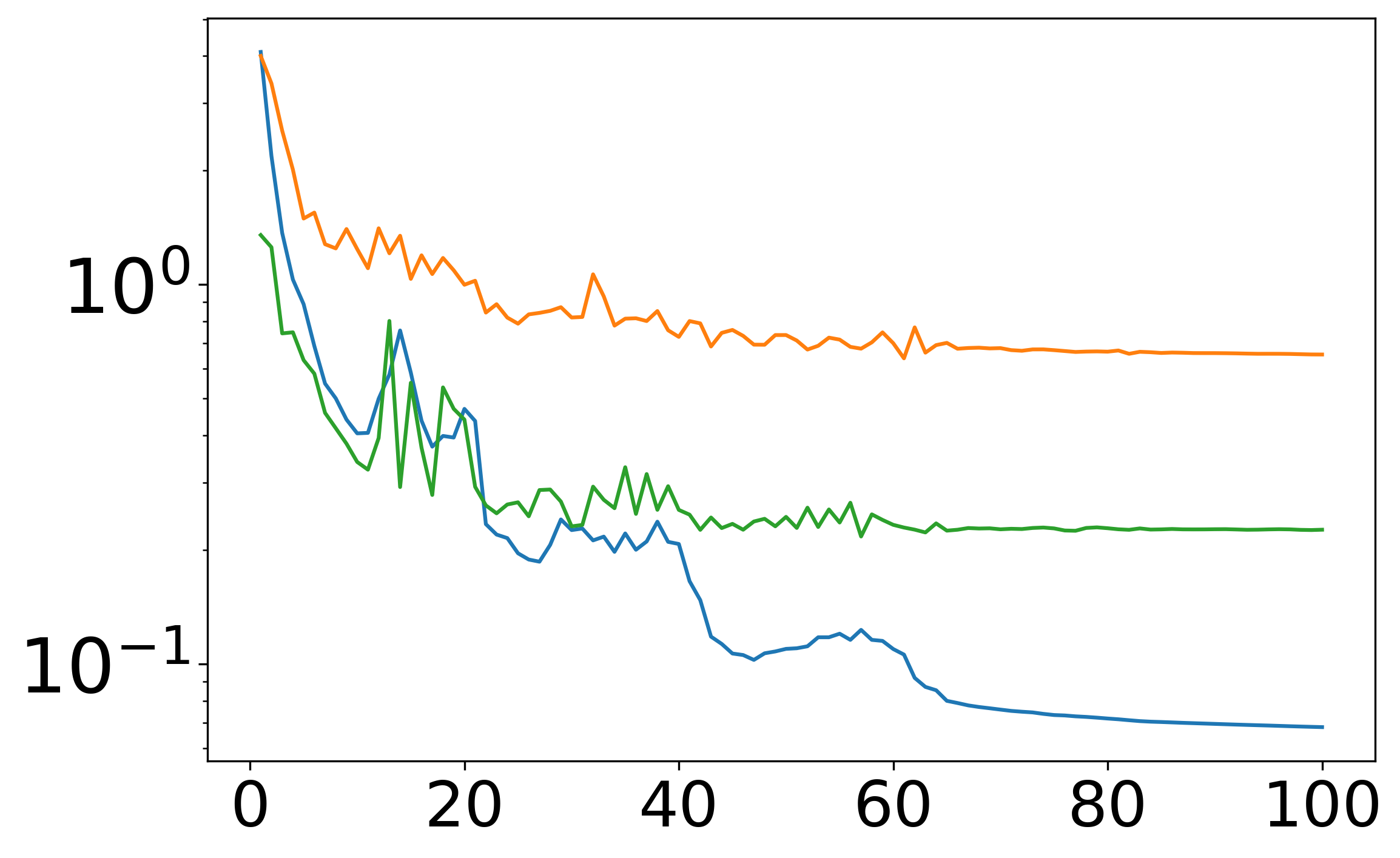}
        \caption{Stage 1}
    \end{subfigure}\hfill
    \begin{subfigure}[t]{0.30\textwidth}
        \centering
        \includegraphics[width=\linewidth]{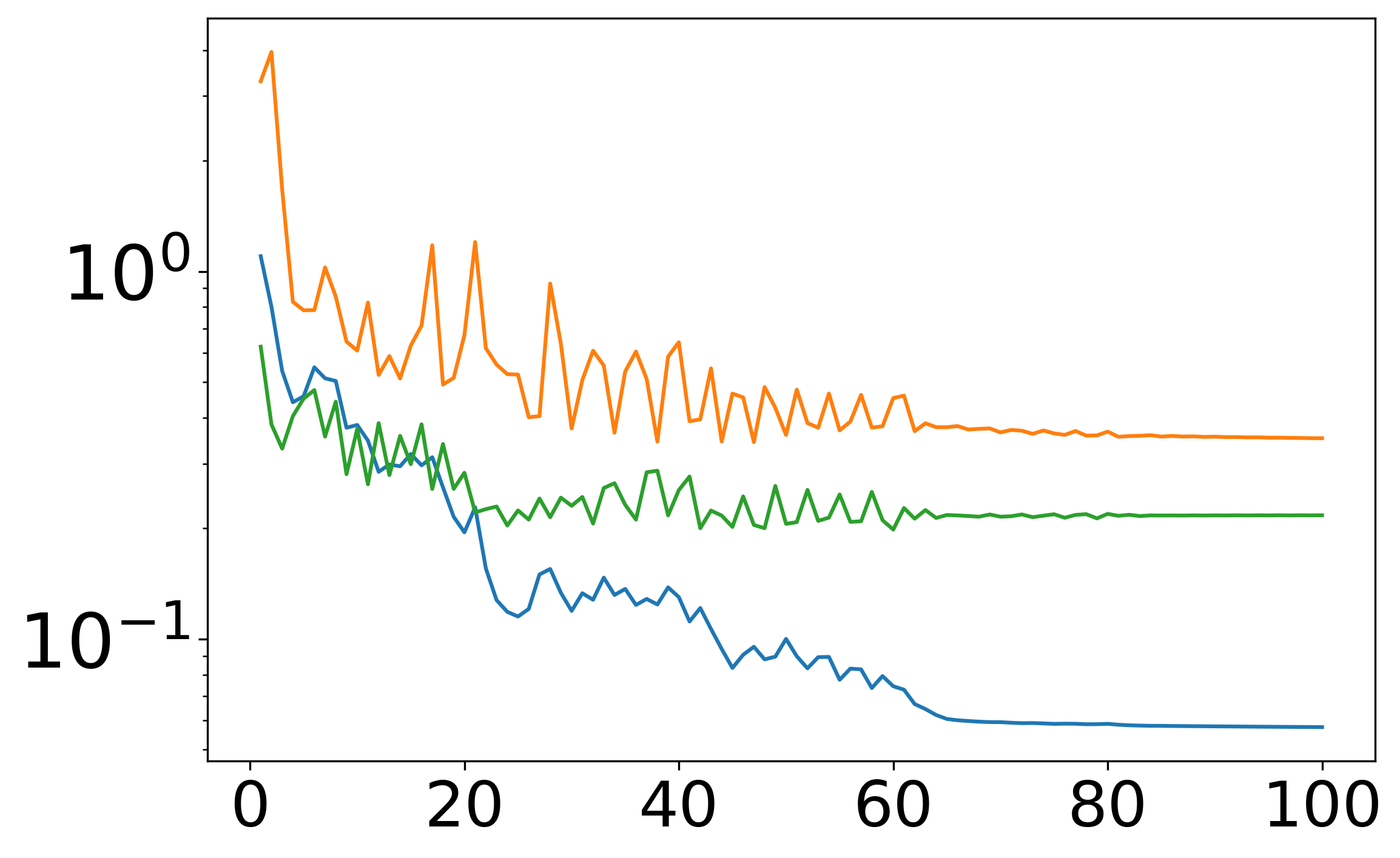}
        \caption{Stage 2}
    \end{subfigure}\hfill
    \begin{subfigure}[t]{0.30\textwidth}
        \centering
        \includegraphics[width=\linewidth]{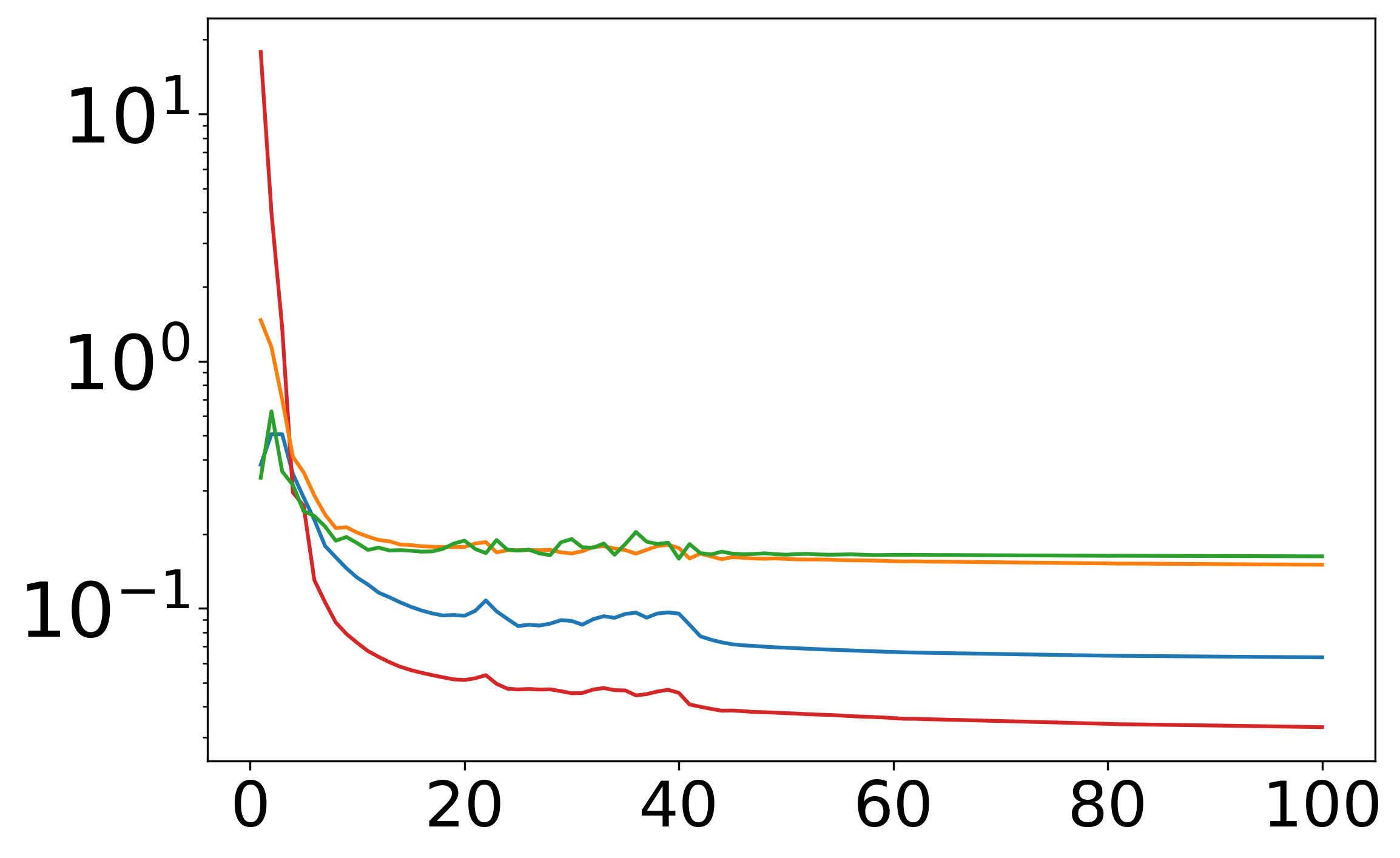}
        \caption{Stage 3}
    \end{subfigure}

    \vspace{8pt}

    \begin{subfigure}[t]{0.30\textwidth}
        \centering
        \includegraphics[width=\linewidth]{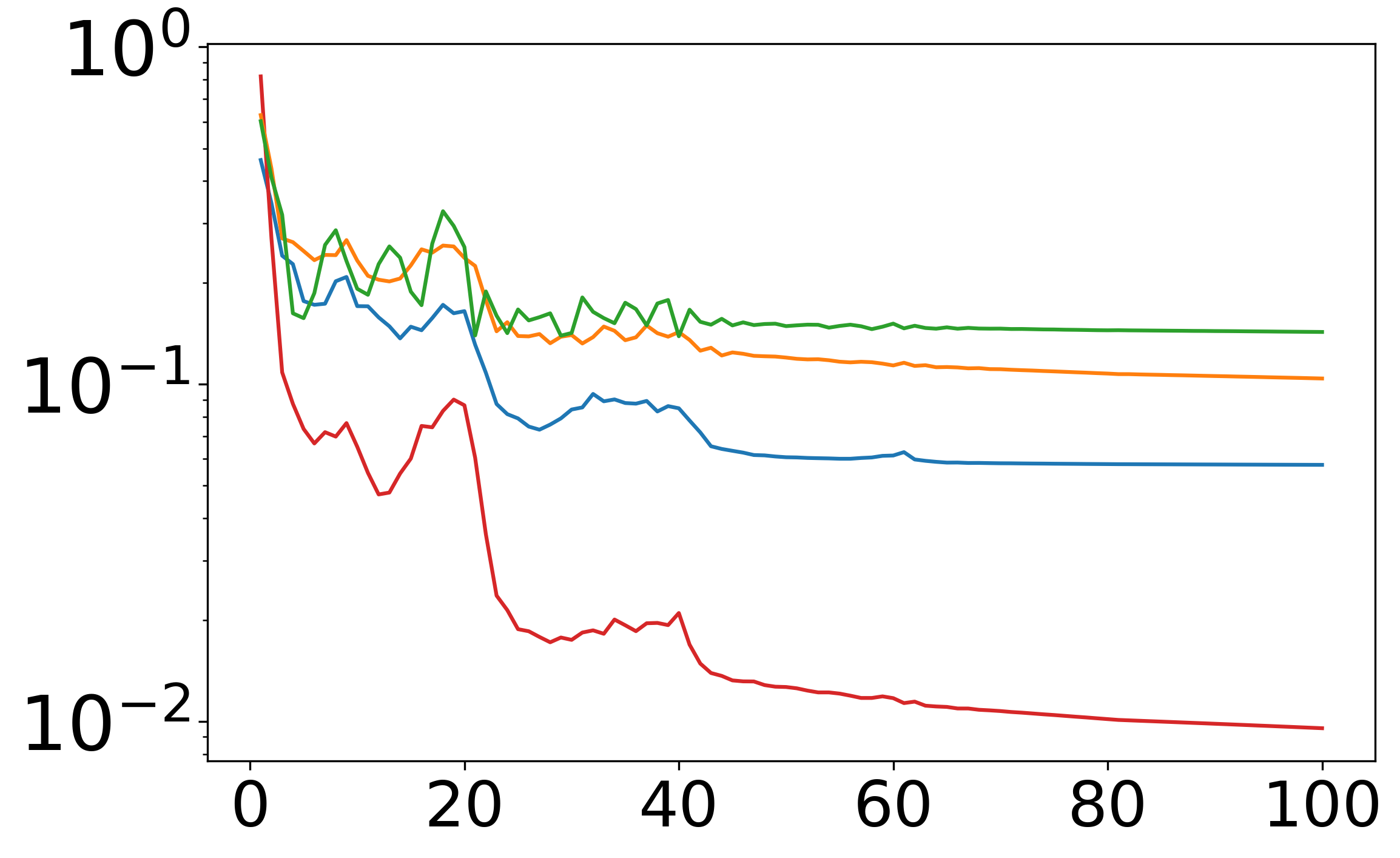}
        \caption{Stage 4}
    \end{subfigure}
    \begin{subfigure}[t]{0.30\textwidth}
        \centering
        \includegraphics[width=\linewidth]{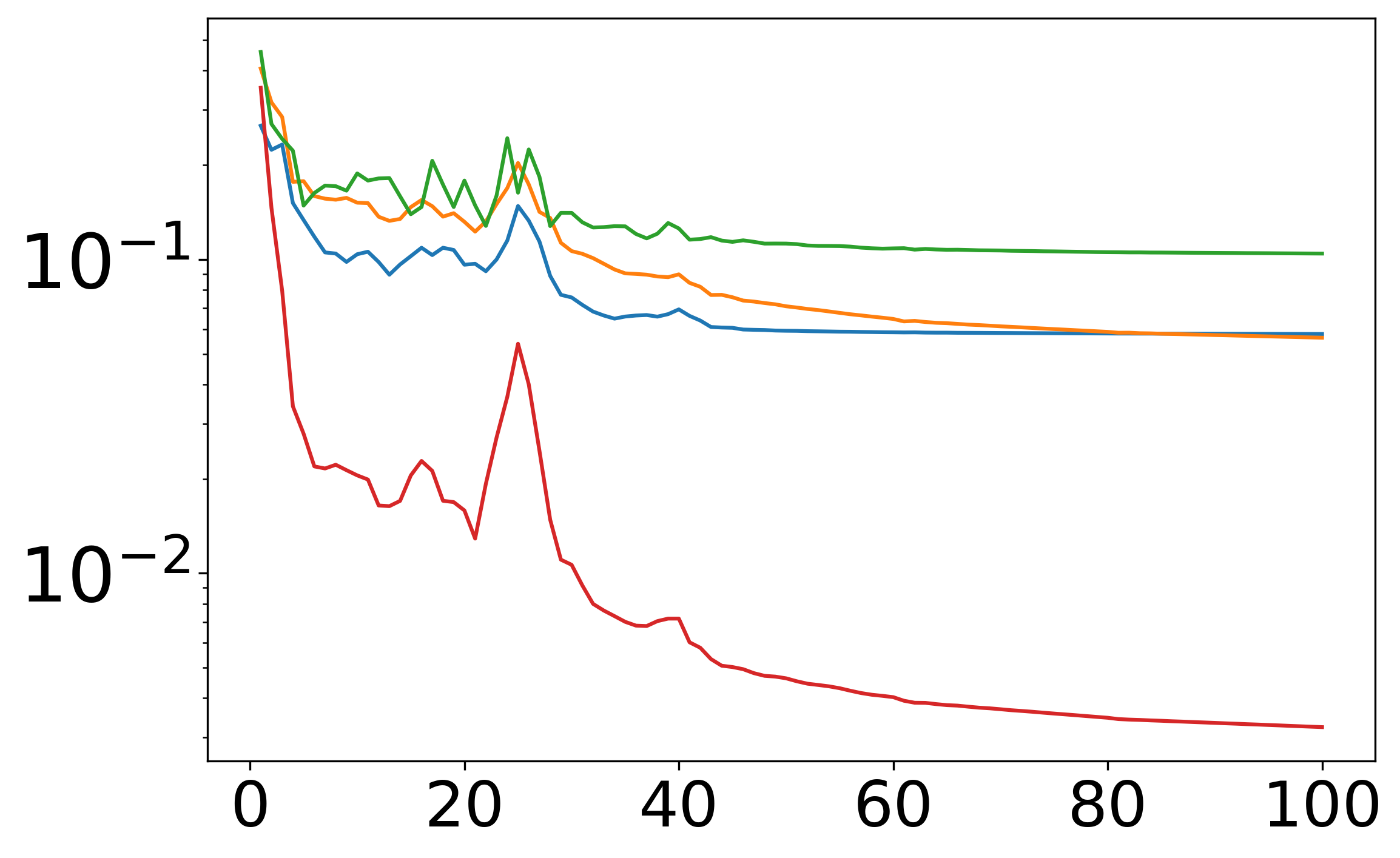}
        \caption{Stage 5}
    \end{subfigure}

\caption{\textbf{Stage-wise loss evolution for 2D Navier--Stokes at two viscosity regimes.}
Top: $\nu = 10^{-3}$; bottom: $\nu = 10^{-4}$.
Each panel shows the evolution, in logarithmic scale, of the losses used during training:
boundary loss $\mathcal{L}_{\mathrm{bd}}$ (blue),
physics residual loss $\mathcal{L}_{\mathrm{res}}$ (red),
total training loss $\mathcal{L}_{\mathrm{train}}$ (orange),
and total test loss $\mathcal{L}_{\mathrm{test}}$ (green).
As the curriculum shifts the supervision from boundary to residual enforcement,
all loss terms decrease monotonically, indicating stable optimization across both viscosity regimes.}

    \label{fig:nsloss-both}
\end{figure*}

Figures~\ref{fig:nsloss-both} illustrate the stage-wise evolution of the boundary, residual, training, and validation losses 
during multi-stage training of PhIS--FNO on the two-dimensional Navier--Stokes equation 
at viscosities \(\nu = 10^{-3}\) and \(\nu = 10^{-4}\), respectively. 
Each panel corresponds to one curriculum phase and the complete set of stage-dependent weights and final loss values for both viscosity regimes 
is summarized in Table~\ref{tab:ns_stage_losses}.\\

\begin{table}[htbp]
\captionsetup{font=footnotesize, skip=4pt}
\centering
\renewcommand{\arraystretch}{1.15}
\setlength{\tabcolsep}{8pt}

\begin{minipage}{\linewidth}
\centering
\text{Navier--Stokes (\(\nu = 10^{-3}\))}\\[4pt]
\resizebox{0.9\linewidth}{!}{%
\begin{tabular}{cccccccc}
\toprule
\textbf{Stage} & $\lambda_{\mathrm{bd}}$ & $\lambda_{\mathrm{res}}$ 
& $\mathcal{L}_{\text{bd}}$ & $\mathcal{L}_{\text{res}}$
& $\mathcal{L}_{\text{train}}$ & $\mathcal{L}_{\text{test}}$ \\
\midrule
1 & 1.0 & 0.0 & $6.78{\times}10^{-2}$ & -- & $6.13{\times}10^{-1}$ & $4.78{\times}10^{-1}$ \\
2 & 1.0 & 0.0 & $5.14{\times}10^{-2}$ & -- & $2.99{\times}10^{-1}$ & $2.78{\times}10^{-1}$ \\
3 & 0.8 & 0.5 & $5.43{\times}10^{-2}$ & $2.95{\times}10^{-2}$ & $8.61{\times}10^{-2}$ & $9.51{\times}10^{-2}$ \\
4 & 0.5 & 1.0 & $5.15{\times}10^{-2}$ & $7.06{\times}10^{-3}$ & $6.17{\times}10^{-2}$ & $6.44{\times}10^{-2}$ \\
5 & 0.3 & 1.5 & $5.18{\times}10^{-2}$ & $3.38{\times}10^{-3}$ & $5.58{\times}10^{-2}$ & $6.01{\times}10^{-2}$ \\
\bottomrule
\end{tabular}%
}
\end{minipage}

\vspace{8pt}

\begin{minipage}{\linewidth}
\centering
\text{Navier--Stokes (\(\nu = 10^{-4}\))}\\[4pt]
\resizebox{0.9\linewidth}{!}{%
\begin{tabular}{cccccccc}
\toprule
\textbf{Stage} & $\lambda_{\mathrm{bd}}$ & $\lambda_{\mathrm{res}}$ 
& $\mathcal{L}_{\text{bd}}$ & $\mathcal{L}_{\text{res}}$
& $\mathcal{L}_{\text{train}}$ & $\mathcal{L}_{\text{test}}$ \\
\midrule
1 & 1.0 & 0.0 & $6.87{\times}10^{-2}$ & -- & $6.56{\times}10^{-1}$ & $2.26{\times}10^{-1}$ \\
2 & 1.0 & 0.0 & $5.77{\times}10^{-2}$ & -- & $3.53{\times}10^{-1}$ & $2.17{\times}10^{-1}$ \\
3 & 0.8 & 0.5 & $6.36{\times}10^{-2}$ & $3.33{\times}10^{-2}$ & $1.51{\times}10^{-1}$ & $1.63{\times}10^{-1}$ \\
4 & 0.5 & 1.0 & $5.77{\times}10^{-2}$ & $9.68{\times}10^{-3}$ & $1.05{\times}10^{-1}$ & $1.43{\times}10^{-1}$ \\
5 & 0.3 & 1.5 & $5.79{\times}10^{-2}$ & $3.28{\times}10^{-3}$ & $5.68{\times}10^{-2}$ & $1.05{\times}10^{-1}$ \\
\bottomrule
\end{tabular}%
}
\end{minipage}

\caption{Final values of the boundary, residual, training, and test losses (\(\mathcal{L}_{\text{bd}}\), \(\mathcal{L}_{\text{res}}\), \(\mathcal{L}_{\text{train}}\), \(\mathcal{L}_{\text{test}}\)) across all curriculum stages for the two-dimensional Navier--Stokes equation at viscosities \(\nu = 10^{-3}\) and \(\nu = 10^{-4}\). Each stage corresponds to a distinct pair of boundary and residual weights \((\lambda_{\mathrm{bd}}, \lambda_{\mathrm{res}})\).}
\label{tab:ns_stage_losses}
\end{table}

\begin{figure}[htbp]

    \centering
    \includegraphics[width=0.85\linewidth]{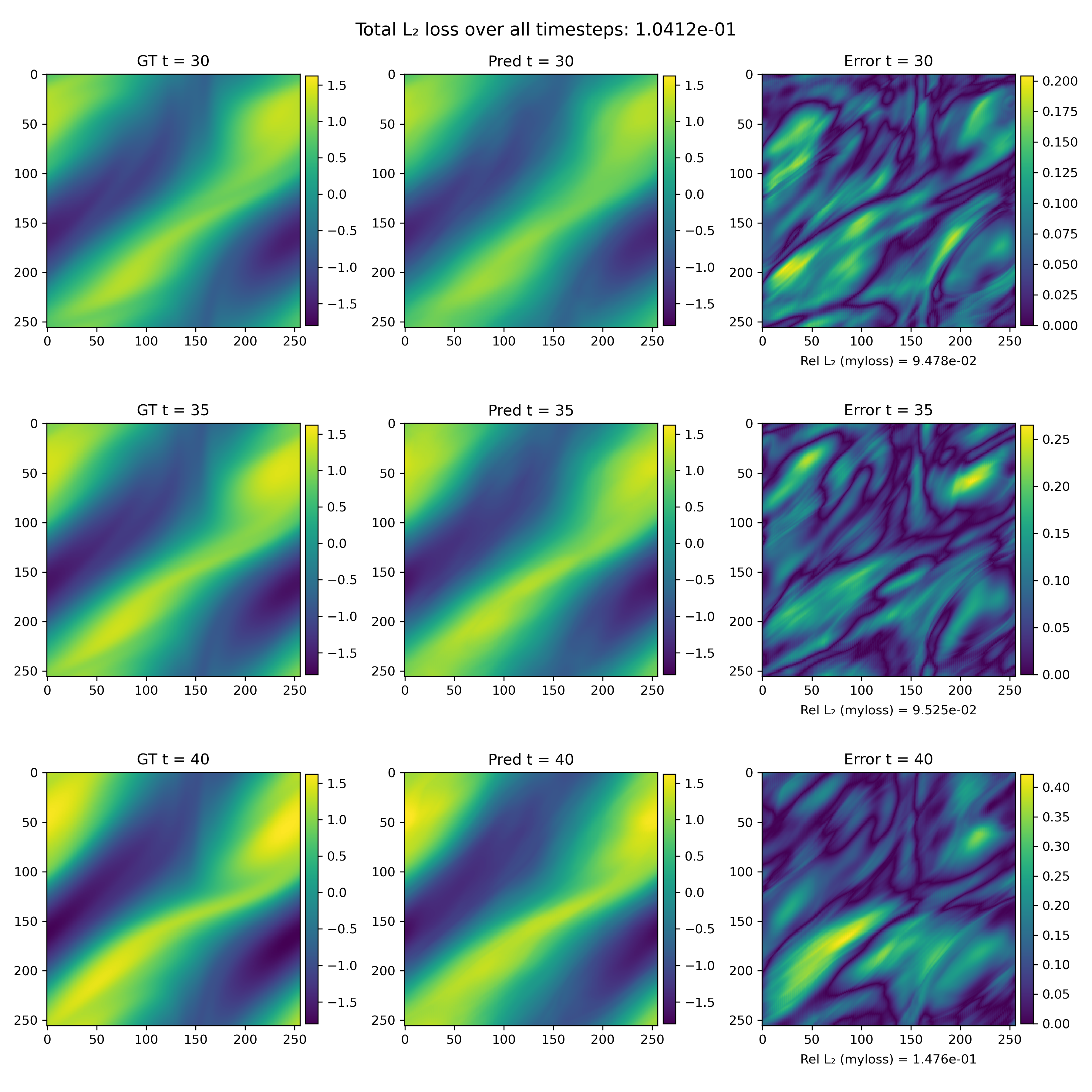}
    \caption{Evolution of vorticity field from test dataset across different timesteps for  $\nu=10^{-4}$. Left: ground truth, center: Prediction, rigth: Absolute error.}
    \label{nscomparison}
   
\end{figure}

Figure~\ref{nscomparison} illustrates qualitative predictions of PhIS-FNO on a representative test sample of the same dataset. Each panel shows the ground truth, the predicted vorticity field, and the corresponding absolute error across selected timesteps. The results confirm that the unsupervised model maintains high accuracy even at higher Reynolds numbers.

\subsection{Cylinder Wake Flow}\label{sup:cylinderwake}

\begin{figure}[htbp]
    \centering
    \includegraphics[width=0.85\linewidth]{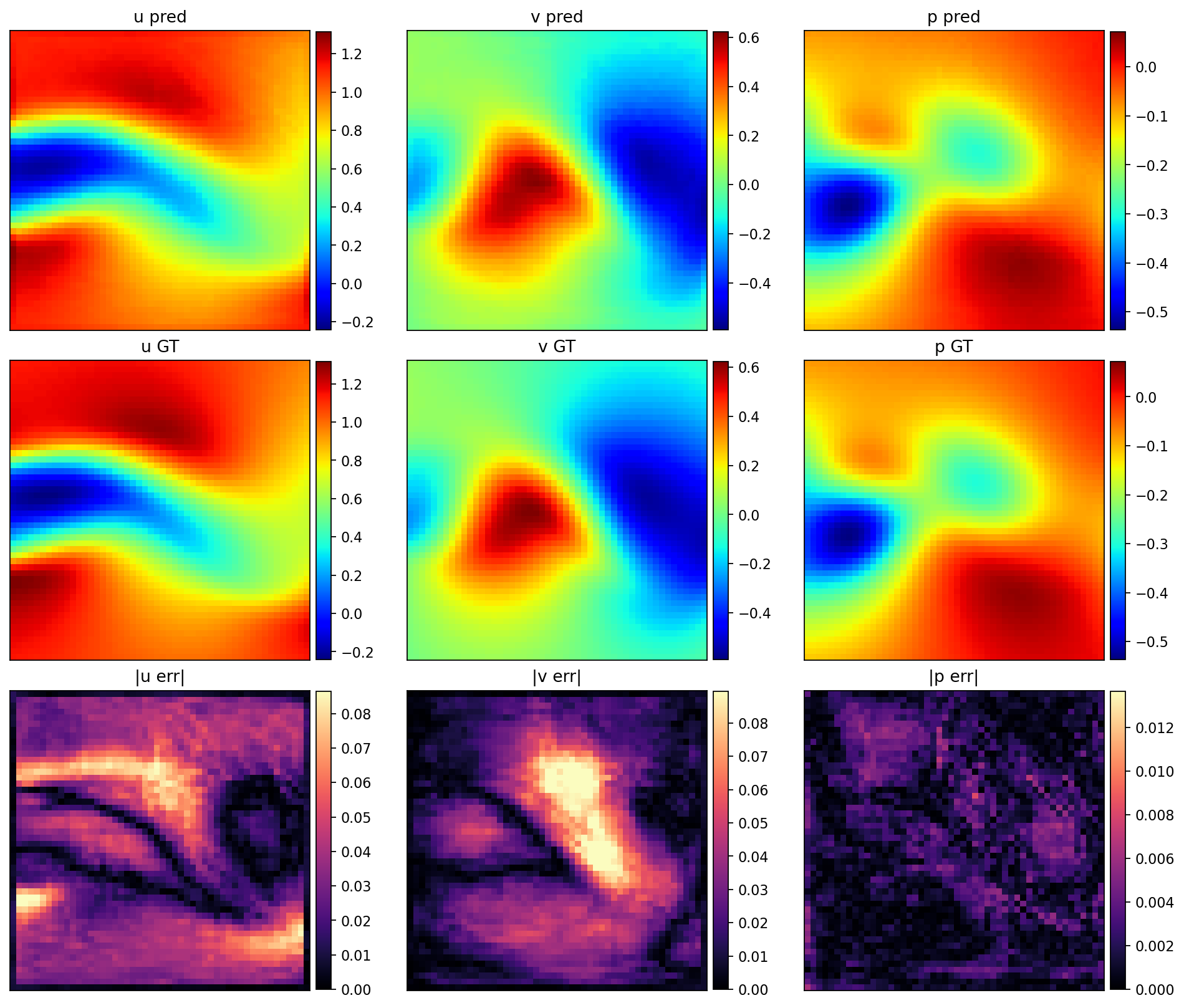}
    \caption{
    \textbf{Qualitative evaluation of the PhIS--FNO on the cylinder wake flow at \(T{=}10\).}
    The first and second rows display, respectively, the predicted and ground-truth fields for the streamwise velocity \(u\), 
    cross-stream velocity, and pressure. 
    The third row reports the pointwise absolute error maps 
    \(|u_{\text{err}}|\), \(|v_{\text{err}}|\), and \(|p_{\text{err}}|\).
    }
    \label{fig:cylinderwake}
\end{figure}

The cylinder wake flow represents a canonical benchmark for non-periodic, vortex-shedding dynamics. 
In this experiment, the PhIS--FNO was trained with full-domain supervision on the pressure field 
and boundary supervision on the velocity components, 
while the interior dynamics were learned in an unsupervised manner 
through progressive enforcement of the Navier Stokes residual within the multi-stage curriculum. 
Figure~\ref{fig:cylinderwake} illustrates the qualitative comparison between predicted and reference fields 
at the final simulation step (\(T = 10\)). The PhIS--FNO achieves excellent reconstruction accuracy across the entire domain, with the lowest errors observed near the cylinder surface and along the boundaries.  This indicates that the model effectively preserves physical consistency at the boundaries while maintaining smooth and stable convergence throughout the multi-stage unsupervised training.

\bibliographystyle{elsarticle-harv} 
\bibliography{bibliography.bib}

@article{Li2020FourierNO,
  title = {Fourier Neural Operator for Parametric Partial Differential Equations},
  author = {Zong-Yi Li and Nikola B. Kovachki and Kamyar Azizzadenesheli and Burigede Liu and Kaushik Bhattacharya and Andrew M. Stuart and Anima Anandkumar},
  journal = {ArXiv},
  year = {2020},
  volume = {abs/2010.08895},
  url = {https://api.semanticscholar.org/CorpusID:224705257}
}

@article{NeuralOperator,
  author = {Kovachki, Nikola and Li, Zongyi and Liu, Burigede and Azizzadenesheli, Kamyar and Bhattacharya, Kaushik and Stuart, Andrew and Anandkumar, Anima},
  title = {Neural operator: learning maps between function spaces with applications to PDEs},
  year = {2023},
  issue_date = {January 2023},
  publisher = {JMLR.org},
  volume = {24},
  number = {1},
  issn = {1532-4435},
  journal = {J. Mach. Learn. Res.},
  month = {jan},
  articleno = {89},
  numpages = {97}
}

@article{raissi2018deephiddenphysics,
  author = {Maziar Raissi},
  title = {Deep Hidden Physics Models: Deep Learning of Nonlinear Partial Differential Equations},
  journal = {Journal of Machine Learning Research},
  volume = {19},
  number = {1},
  pages = {932--955},
  year = {2018},
  url = {https://www.jmlr.org/papers/volume19/18-046/18-046.pdf}
}

@article{guo2025longterm,
  title = {Long-term simulation of physical and mechanical behaviors using curriculum-transfer-learning based physics-informed neural networks},
  author = {Guo, Y. and Fu, Z. and Min, J. and Lin, S. and Liu, X. and Rashed, Y. F. and Zhuang, X.},
  journal = {Neural Networks},
  volume = {191},
  pages = {107825},
  year = {2025},
  doi = {10.1016/j.neunet.2025.107825}
}

@article{zhu2019physicsconstrained,
  author = {Yinhao Zhu and Nicholas Zabaras and Phaedon-Stelios Koutsourelakis and Paris Perdikaris},
  title = {Physics-constrained deep learning for high-dimensional surrogate modeling and uncertainty quantification without labeled data},
  journal = {Journal of Computational Physics},
  volume = {394},
  pages = {56--81},
  year = {2019},
  doi = {10.1016/j.jcp.2019.05.024},
  url = {https://www.sciencedirect.com/science/article/pii/S0021999119303559}
}

@article{fuks2020limitations,
  author = {Olga Fuks and Hamdi A. Tchelepi},
  title = {Limitations of Physics Informed Machine Learning for Nonlinear Two-Phase Transport in Porous Media},
  journal = {Journal of Machine Learning for Modeling and Computing},
  volume = {1},
  number = {1},
  year = {2020},
  url = {https://www.dl.begellhouse.com/journals/52034eb04b657aea,4a6efccf0b16b3d5,1ad2d8c64d7920c0.html}
}

@inproceedings{rahaman2019spectralbiasneuralnetworks,
  author = {Nasim Rahaman and Aristide Baratin and Devansh Arpit and Felix Draxler and Min Lin and Fred A. Hamprecht and Yoshua Bengio and Aaron Courville},
  title = {On the Spectral Bias of Neural Networks},
  booktitle = {Proceedings of the 36th International Conference on Machine Learning (ICML)},
  pages = {5301--5310},
  year = {2019},
  url = {https://proceedings.mlr.press/v97/rahaman19a.html}
}

@article{cao2019spectralbias,
  author = {Yuan Cao and Zhiying Fang and Yue Wu and Ding-Xuan Zhou and Quanquan Gu},
  title = {Towards Understanding the Spectral Bias of Deep Learning},
  journal = {arXiv preprint arXiv:1912.01198},
  year = {2019},
  url = {https://arxiv.org/abs/1912.01198}
}

@article{tancik2020fourierfeatures,
  author = {Matthew Tancik and Pratul P. Srinivasan and Ben Mildenhall and Sara Fridovich-Keil and Nithin Raghavan and Utkarsh Singhal and Ravi Ramamoorthi and Jonathan T. Barron and Ren Ng},
  title = {Fourier Features Let Networks Learn High Frequency Functions in Low Dimensional Domains},
  journal = {Advances in Neural Information Processing Systems (NeurIPS)},
  year = {2020},
  url = {https://arxiv.org/abs/2006.10739}
}

@inproceedings{ronneberger2015unet,
  title = {U-Net: Convolutional Networks for Biomedical Image Segmentation},
  author = {Olaf Ronneberger and Philipp Fischer and Thomas Brox},
  booktitle = {Medical Image Computing and Computer-Assisted Intervention (MICCAI)},
  pages = {234--241},
  year = {2015},
  organization = {Springer},
  series = {Lecture Notes in Computer Science},
  volume = {9351},
  doi = {10.1007/978-3-319-24574-4\_28}
}

@book{allgower2003introduction,
  title     = {Introduction to Numerical Continuation Methods},
  author    = {Eugene L. Allgower and Kurt Georg},
  publisher = {Society for Industrial and Applied Mathematics (SIAM)},
  address   = {Philadelphia, PA},
  year      = {2003},
  doi       = {10.1137/1.9780898719154},
  isbn      = {978-0-89871-546-9},
  url       = {https://doi.org/10.1137/1.9780898719154}
}

@misc{SplineWandel,
  title = {Spline-PINN: Approaching PDEs without Data using Fast, Physics-Informed Hermite-Spline CNNs},
  author = {Nils Wandel and Michael Weinmann and Michael Neidlin and Reinhard Klein},
  year = {2022},
  eprint = {2109.07143},
  archivePrefix = {arXiv},
  primaryClass = {cs.LG},
  url = {https://arxiv.org/abs/2109.07143}
}

@misc{wandel2021learningincompressiblefluiddynamics,
  title = {Learning Incompressible Fluid Dynamics from Scratch -- Towards Fast, Differentiable Fluid Models that Generalize},
  author = {Nils Wandel and Michael Weinmann and Reinhard Klein},
  year = {2021},
  eprint = {2006.08762},
  archivePrefix = {arXiv},
  primaryClass = {cs.LG},
  url = {https://arxiv.org/abs/2006.08762}
}

@misc{wandel2025metamizerversatileneuraloptimizer,
  title = {Metamizer: a versatile neural optimizer for fast and accurate physics simulations},
  author = {Nils Wandel and Stefan Schulz and Reinhard Klein},
  year = {2025},
  eprint = {2410.19746},
  archivePrefix = {arXiv},
  primaryClass = {physics.comp-ph},
  url = {https://arxiv.org/abs/2410.19746}
}

@misc{splinedeep,
  title = {SplineCNN: Fast Geometric Deep Learning with Continuous B-Spline Kernels},
  author = {Matthias Fey and Jan Eric Lenssen and Frank Weichert and Heinrich Müller},
  year = {2018},
  eprint = {1711.08920},
  archivePrefix = {arXiv},
  primaryClass = {cs.CV},
  url = {https://arxiv.org/abs/1711.08920}
}

@misc{curr1,
  title = {Physics-informed neural networks with curriculum training for poroelastic flow and deformation processes},
  author = {Yared W. Bekele},
  year = {2024},
  eprint = {2404.13909},
  archivePrefix = {arXiv},
  primaryClass = {cs.CE},
  url = {https://arxiv.org/abs/2404.13909}
}

@article{kingma2014adam,
  title     = {Adam: A Method for Stochastic Optimization},
  author    = {Kingma, Diederik P. and Ba, Jimmy},
  journal   = {arXiv preprint arXiv:1412.6980},
  year      = {2015},
  url       = {https://arxiv.org/abs/1412.6980}
}

@article{kharazmi2021hpinns,
  title     = {hp-VPINNs: Variational Physics-Informed Neural Networks with Domain Decomposition},
  author    = {Kharazmi, Ehsan and Zhang, Zhen and Karniadakis, George Em},
  journal   = {Computer Methods in Applied Mechanics and Engineering},
  volume    = {374},
  pages     = {113547},
  year      = {2021},
  publisher = {Elsevier},
  doi       = {10.1016/j.cma.2020.113547}
}

@article{Khadijeh2025,
  author = {Mahmoud Khadijeh and Veronica Cerqueglini and Cor Kasbergen and Sandra Erkens and Aikaterini Varveri},
  title = {Multistage physics informed neural network for solving coupled multiphysics problems in material degradation and fluid dynamics},
  journal = {Engineering with Computers},
  year = {2025},
  month = {June},
  day = {29},
  doi = {10.1007/s00366-025-02174-4},
  url = {https://doi.org/10.1007/s00366-025-02174-4},
  issn = {1435-5663}
}

@article{MATTEY2022114474,
  title = {A novel sequential method to train physics informed neural networks for Allen Cahn and Cahn Hilliard equations},
  journal = {Computer Methods in Applied Mechanics and Engineering},
  volume = {390},
  pages = {114474},
  year = {2022},
  issn = {0045-7825},
  doi = {https://doi.org/10.1016/j.cma.2021.114474},
  url = {https://www.sciencedirect.com/science/article/pii/S0045782521006939},
  author = {Revanth Mattey and Susanta Ghosh}
}

@inproceedings{duan2025copinn,
  title = {CoPINN: Cognitive Physics-Informed Neural Networks},
  author = {Siyuan Duan and Wenyuan Wu and Peng Hu and Zhenwen Ren and Dezhong Peng and Yuan Sun},
  booktitle = {Forty-second International Conference on Machine Learning},
  year = {2025},
  url = {https://openreview.net/forum?id=4vAa0A98xI}
}

@misc{li2023physicsinformedneuraloperatorlearning,
  title = {Physics-Informed Neural Operator for Learning Partial Differential Equations},
  author = {Zongyi Li and Hongkai Zheng and Nikola Kovachki and David Jin and Haoxuan Chen and Burigede Liu and Kamyar Azizzadenesheli and Anima Anandkumar},
  year = {2023},
  eprint = {2111.03794},
  archivePrefix = {arXiv},
  primaryClass = {cs.LG},
  url = {https://arxiv.org/abs/2111.03794}
}

@article{RAISSI2019686,
  title = {Physics-informed neural networks: A deep learning framework for solving forward and inverse problems involving nonlinear partial differential equations},
  journal = {Journal of Computational Physics},
  volume = {378},
  pages = {686-707},
  year = {2019},
  issn = {0021-9991},
  doi = {https://doi.org/10.1016/j.jcp.2018.10.045},
  url = {https://www.sciencedirect.com/science/article/pii/S0021999118307125},
  author = {M. Raissi and P. Perdikaris and G.E. Karniadakis}
}

@article{zhong2024pigano,
  title = {Physics-Informed Geometry-Aware Neural Operator},
  author = {Zhong, Weiheng and Meidani, Hadi},
  journal = {arXiv preprint arXiv:2408.01600},
  year = {2024},
  note = {Preprint}
}

@article{eshaghi2025vino,
  title = {Variational Physics-Informed Neural Operator (VINO) for solving partial differential equations},
  author = {Eshaghi, Mohammad Sadegh and Anitescu, Cosmin and Thombre, Manish and Wang, Yizheng and Zhuang, Xiaoying and Rabczuk, Timon},
  journal = {Computer Methods in Applied Mechanics and Engineering},
  volume = {437},
  pages = {117785},
  year = {2025},
  doi = {10.1016/j.cma.2025.117785},
  issn = {0045-7825},
  url = {https://www.sciencedirect.com/science/article/pii/S004578252500057X}
}

@article{li2021pino,
  title = {Physics-Informed Neural Operator for Learning Partial Differential Equations},
  author = {Li, Zongyi and Zheng, Hongkai and Kovachki, Nikola and Jin, David and Chen, Haoxuan and Liu, Burigede and Azizzadenesheli, Kamyar and Anandkumar, Anima},
  journal = {ACM Journal of Data Science},
  volume = {1},
  number = {3},
  pages = {1–27},
  year = {2021},
  doi = {10.1145/3648506}
}

@article{KIM2020109216,
  title = {Deep unsupervised learning of turbulence for inflow generation at various Reynolds numbers},
  journal = {Journal of Computational Physics},
  volume = {406},
  pages = {109216},
  year = {2020},
  issn = {0021-9991},
  doi = {https://doi.org/10.1016/j.jcp.2019.109216},
  url = {https://www.sciencedirect.com/science/article/pii/S0021999119309210},
  author = {Junhyuk Kim and Changhoon Lee}
}

@misc{mohan2020embeddinghardphysicalconstraints,
  title = {Embedding Hard Physical Constraints in Neural Network Coarse-Graining of 3D Turbulence},
  author = {Arvind T. Mohan and Nicholas Lubbers and Daniel Livescu and Michael Chertkov},
  year = {2020},
  eprint = {2002.00021},
  archivePrefix = {arXiv},
  primaryClass = {physics.comp-ph},
  url = {https://arxiv.org/abs/2002.00021}
}

@article{Thuerey_2020,
  title = {Deep Learning Methods for Reynolds-Averaged Navier–Stokes Simulations of Airfoil Flows},
  volume = {58},
  issn = {1533-385X},
  url = {http://dx.doi.org/10.2514/1.j058291},
  doi = {10.2514/1.j058291},
  number = {1},
  journal = {AIAA Journal},
  publisher = {American Institute of Aeronautics and Astronautics (AIAA)},
  author = {Thuerey, Nils and Weißenow, Konstantin and Prantl, Lukas and Hu, Xiangyu},
  year = {2020},
  month = {jan},
  pages = {25–36}
}

@misc{um2021solverinthelooplearningdifferentiablephysics,
  title = {Solver-in-the-Loop: Learning from Differentiable Physics to Interact with Iterative PDE-Solvers},
  author = {Kiwon Um and Robert Brand and Yun and Fei and Philipp Holl and Nils Thuerey},
  year = {2021},
  eprint = {2007.00016},
  archivePrefix = {arXiv},
  primaryClass = {physics.comp-ph},
  url = {https://arxiv.org/abs/2007.00016}
}

@misc{pfaff2021learningmeshbasedsimulationgraph,
  title = {Learning Mesh-Based Simulation with Graph Networks},
  author = {Tobias Pfaff and Meire Fortunato and Alvaro Sanchez-Gonzalez and Peter W. Battaglia},
  year = {2021},
  eprint = {2010.03409},
  archivePrefix = {arXiv},
  primaryClass = {cs.LG},
  url = {https://arxiv.org/abs/2010.03409}
}

@article{kim19a,
  author = {Kim, Byungsoo and C. Azevedo, Vinicius and Thuerey, Nils and Kim, Theodore and Gross, Markus and Solenthaler, Barbara},
  title = {Deep Fluids: A Generative Network for Parameterized Fluid Simulations},
  journal = {Computer Graphics Forum (Proc. Eurographics)},
  year = {2019},
  volume = {38},
  number = {2},
  pages = {59-70}
}

@article{doi:10.1126/science.aaw4741,
  author = {Maziar Raissi and Alireza Yazdani and George Em Karniadakis},
  title = {Hidden fluid mechanics: Learning velocity and pressure fields from flow visualizations},
  journal = {Science},
  volume = {367},
  number = {6481},
  pages = {1026-1030},
  year = {2020},
  doi = {10.1126/science.aaw4741},
  url = {https://www.science.org/doi/abs/10.1126/science.aaw4741},
  eprint = {https://www.science.org/doi/pdf/10.1126/science.aaw4741}
}

@misc{wang2020understandingmitigatinggradientpathologies,
  title = {Understanding and mitigating gradient pathologies in physics-informed neural networks},
  author = {Sifan Wang and Yujun Teng and Paris Perdikaris},
  year = {2020},
  eprint = {2001.04536},
  archivePrefix = {arXiv},
  primaryClass = {cs.LG},
  url = {https://arxiv.org/abs/2001.04536}
}

@article{Sun_2020,
  title = {Surrogate modeling for fluid flows based on physics-constrained deep learning without simulation data},
  volume = {361},
  issn = {0045-7825},
  url = {http://dx.doi.org/10.1016/j.cma.2019.112732},
  doi = {10.1016/j.cma.2019.112732},
  journal = {Computer Methods in Applied Mechanics and Engineering},
  publisher = {Elsevier BV},
  author = {Sun, Luning and Gao, Han and Pan, Shaowu and Wang, Xun},
  year = {2020},
  month = {apr},
  pages = {112732}
}

@misc{li2020multipolegraphneuraloperator,
  title = {Multipole Graph Neural Operator for Parametric Partial Differential Equations},
  author = {Zongyi Li and Nikola Kovachki and Kamyar Azizzadenesheli and Burigede Liu and Kaushik Bhattacharya and Andrew Stuart and Anima Anandkumar},
  year = {2020},
  eprint = {2006.09535},
  archivePrefix = {arXiv},
  primaryClass = {cs.LG},
  url = {https://arxiv.org/abs/2006.09535}
}

@misc{li2020neuraloperatorgraphkernel,
  title = {Neural Operator: Graph Kernel Network for Partial Differential Equations},
  author = {Zongyi Li and Nikola Kovachki and Kamyar Azizzadenesheli and Burigede Liu and Kaushik Bhattacharya and Andrew Stuart and Anima Anandkumar},
  year = {2020},
  eprint = {2003.03485},
  archivePrefix = {arXiv},
  primaryClass = {cs.LG},
  url = {https://arxiv.org/abs/2003.03485}
}

@article{Kovachki,
  author = {Nikola Kovachki and Zongyi Li and Burigede Liu and Kamyar Azizzadenesheli and Kaushik Bhattacharya and Andrew Stuart and Anima Anandkumar},
  title = {Neural Operator: Learning Maps Between Function Spaces With Applications to PDEs},
  journal = {Journal of Machine Learning Research},
  year = {2023},
  volume = {24},
  number = {89},
  pages = {1--97},
  url = {http://jmlr.org/papers/v24/21-1524.html}
}

@misc{graves2017automatedcurriculumlearningneural,
  title = {Automated Curriculum Learning for Neural Networks},
  author = {Alex Graves and Marc G. Bellemare and Jacob Menick and Remi Munos and Koray Kavukcuoglu},
  year = {2017},
  eprint = {1704.03003},
  archivePrefix = {arXiv},
  primaryClass = {cs.NE},
  url = {https://arxiv.org/abs/1704.03003}
}

@article{https://doi.org/10.1002/qj.3803,
  author = {Hersbach, Hans and Bell, Bill and Berrisford, Paul and Hirahara, Shoji and Horányi, András and Muñoz-Sabater, Joaquín and Nicolas, Julien and Peubey, Carole and Radu, Raluca and Schepers, Dinand and Simmons, Adrian and Soci, Cornel and Abdalla, Saleh and Abellan, Xavier and Balsamo, Gianpaolo and Bechtold, Peter and Biavati, Gionata and Bidlot, Jean and Bonavita, Massimo and De Chiara, Giovanna and Dahlgren, Per and Dee, Dick and Diamantakis, Michail and Dragani, Rossana and Flemming, Johannes and Forbes, Richard and Fuentes, Manuel and Geer, Alan and Haimberger, Leo and Healy, Sean and Hogan, Robin J. and Hólm, Elías and Janisková, Marta and Keeley, Sarah and Laloyaux, Patrick and Lopez, Philippe and Lupu, Cristina and Radnoti, Gabor and de Rosnay, Patricia and Rozum, Iryna and Vamborg, Freja and Villaume, Sebastien and Thépaut, Jean-Noël},
  title = {The ERA5 global reanalysis},
  journal = {Quarterly Journal of the Royal Meteorological Society},
  year = {2020},
  volume = {146},
  number = {730},
  pages = {1999-2049},
  doi = {https://doi.org/10.1002/qj.3803},
  url = {https://rmets.onlinelibrary.wiley.com/doi/abs/10.1002/qj.3803}
}

@misc{wang2020pinnsfailtrainneural,
      title= {When and why PINNs fail to train: A neural tangent kernel perspective}, 
      author= {Sifan Wang and Xinling Yu and Paris Perdikaris},
      year= {2020},
      eprint= {2007.14527},
      archivePrefix= {arXiv},
      primaryClass= {cs.LG},
      url= {https://arxiv.org/abs/2007.14527}, 
}

@unknown{deeponet,
  author = {Lu, Lu and Jin, Pengzhan and Karniadakis, George},
  year = {2019},
  month = {10},
  title = {DeepONet: Learning nonlinear operators for identifying differential equations based on the universal approximation theorem of operators}
}


\end{document}